\documentclass{article}

\usepackage[final]{neurips_2023}

\usepackage[utf8]{inputenc} %
\usepackage[T1]{fontenc}    %
\usepackage[colorlinks=true,linkcolor={sb_blue},citecolor={sb_blue},filecolor={sb_blue},urlcolor={sb_blue}]{hyperref}
\usepackage{url}            %
\usepackage{booktabs}       %
\usepackage{amsfonts}       %
\usepackage{nicefrac}       %
\usepackage{microtype}      %
\usepackage{xcolor}         %

\usepackage{mine}

\newcommand\blfootnote[1]{%
	\begingroup
	\renewcommand\thefootnote{}\footnote{#1}%
	\addtocounter{footnote}{-1}%
	\endgroup
} %

\title{For SALE: State-Action Representation Learning\\for Deep Reinforcement Learning}

\author{%
Scott Fujimoto \\
Mila, McGill University \\
\And
Wei-Di Chang \\
\makebox[\widthof{Mila, McGill University}]{McGill University}
\And
Edward J.\ Smith \\
\makebox[\widthof{Mila, McGill University}]{McGill University}
\AND 
Shixiang Shane Gu \\
\makebox[\widthof{Mila, McGill University}]{Google DeepMind}
\And 
Doina Precup \\
Mila, McGill University
\And 
David Meger \\
Mila, McGill University
}

\begin{document}

\maketitle

\begin{abstract}

	In the field of reinforcement learning (RL), representation learning is a proven tool for complex image-based tasks, but is often overlooked for environments with low-level states, such as physical control problems. 
	This paper introduces SALE, a novel approach for learning embeddings that model the nuanced interaction between state and action, enabling effective representation learning from low-level states. We extensively study the design space of these embeddings and highlight important design considerations. We integrate SALE and an adaptation of checkpoints for RL into TD3 to form the TD7 algorithm, which significantly outperforms existing continuous control algorithms. On OpenAI gym benchmark tasks, TD7 has an average performance gain of 276.7\% and 50.7\% over TD3 at 300k and 5M time steps, respectively, and works in both the online and offline settings. 
\end{abstract}

\section{Introduction}

Reinforcement learning (RL) is notoriously sample inefficient, particularly when compared to more straightforward paradigms in machine learning, such as supervised learning. \blfootnote{Corresponding author: \texttt{scott.fujimoto@mail.mcgill.ca}}
One possible explanation is the usage of the Bellman equation in most off-policy RL algorithms~\citep{DQN, DDPG}, which provides a weak learning signal due to an approximate and non-stationary learning target~\citep{fujimoto2022should}. 

A near-universal solution to sample inefficiency in deep learning is representation learning, whereby intermediate features are learned to capture the underlying structure and patterns of the data. These features can be found independently from the downstream task and considerations such as the learning horizon and dynamic programming. 
While feature learning of this type has found some success in the RL setting, it has been mainly limited to vision-based environments~\citep{jaderberg2017reinforcement, oord2018representation, anand2019unsupervised, laskin2020curl, stooke2021decoupling, yarats2022mastering}.

On the other hand, the application of representation learning to low-level states is much less common. At first glance, it may seem unnecessary to learn a representation over an already-compact state vector. However, we argue that the difficulty of a task is often defined by the complexity of the underlying dynamical system, rather than the size of the observation space. This means that regardless of the original observation space, there exists an opportunity to learn meaningful features by capturing the interaction between state and action. 

\textbf{SALE.} In this paper, we devise state-action learned embeddings~(SALE), a method that learns embeddings jointly over both state and action by modeling the dynamics of the environment in latent space. Extending prior work~\citep{ota2020can}, we introduce three important design considerations when learning a state-action representation online. Most importantly, we observe the surprising effect of extrapolation error~\citep{fujimoto2019off} when significantly expanding the action-dependent input and introduce a simple clipping technique to mitigate it. 

\textbf{Design study.} Learning to model environment dynamics in latent space is a common approach for feature learning which has been widely considered~\citep{watter2015embed, ha2018world, hafner2019learning, gelada2019deepmdp, schwarzer2020data}, with many possible variations in design. Consequently, the optimal design decision is often unclear without considering empirical performance. To this end, we perform an extensive empirical evaluation over the design space, with the aim of discovering which choices are the most significant contributors to final performance. 

\textbf{Checkpoints.} Next, we explore the usage of checkpoints in RL. Similar to representation learning, early stopping and checkpoints are standard techniques used to enhance the performance of deep learning models. A similar effect can be achieved in RL by fixing each policy for multiple training episodes, and then at test time, using the highest-performing policy observed during training. 

\textbf{TD7.} We combine TD3 with our state-action representation learning method SALE, the aforementioned checkpoints, prioritized experience replay~\citep{fujimoto2020equivalence}, and a behavior cloning term (used only for offline RL)~\citep{fujimoto2021minimalist} to form the TD7~(TD3\texttt{+}4 additions) algorithm. We benchmark the TD7 algorithm in both the online and offline RL setting. 
TD7 significantly outperforms existing %
methods without the additional complexity from competing methods such as large ensembles, additional updates per time step, or per-environment hyperparameters. 
Our key improvement, SALE, works in tandem with most RL methods and can be used to enhance existing approaches in both the online and offline setting. Our code is open-sourced\footnote{\url{https://github.com/sfujim/TD7}}.

\section{Related Work}

\textbf{Representation learning.} Representation learning has several related interpretations in RL. Historically, representation learning referred to \textit{abstraction}, mapping an MDP to a smaller one via bisimulation or other means~\citep{li2006towards, ferns2011bisimulation, zhang2020learning}. For higher-dimensional spaces, the notion of true abstraction has been replaced with \textit{compression}, where the intent is to embed the observation space (such as images) into a smaller manageable latent vector~\citep{watter2015embed, finn2016deep, gelada2019deepmdp}. Representation learning can also refer to \textit{feature learning}, where the objective is to learn features that capture relevant aspects of the environment or task, via auxiliary rewards or alternate training signals~\citep{sutton2011horde, jaderberg2017reinforcement, riedmiller2018learning, lin2019adaptive}. In recent years, representation learning in RL often refers to both compression and feature learning, and is commonly employed in image-based tasks~\citep{kostrikov2020image, yarats2021improving, liu2021return, cetin2022stabilizing} where the observation space is characterized by its high dimensionality and the presence of redundant information. 

Representation learning by predicting future states draws inspiration from a rich history~\citep{dayan1993improving, littman2001predictive}, spanning many approaches in both model-free RL~\citep{munk2016learning, van2016stable, zhang2018decoupling, gelada2019deepmdp, schwarzer2020data, fujimoto2021srdice, ota2020can, ota2021training} and model-based RL in latent space~\citep{watter2015embed, finn2016deep, karl2017deep, ha2018world, hansen2022temporal, hafner2019learning, hafner2023mastering}. Another related approach is representation learning over actions~\citep{tennenholtz2019natural, chandak2019learning, whitney2020dynamics}. Our key distinction from many previous approaches is the emphasis on learning joint representations of both state and action. 

Methods which do learn state-action representations, by auxiliary rewards to the value function~\citep{liu2021return}, or MDP homomorphisms~\citep{ravindran2004algebraic, van2020plannable, van2020mdp, rezaei2022continuous} emphasize abstraction more than feature learning. Our approach can be viewed as an extension of OFENet~\citep{ota2020can}, which also learns a state-action embedding. We build off of OFENet and other representation learning methods by highlighting crucial design considerations and addressing the difficulties that arise when using decoupled state-action embeddings. Our resulting improvements are reflected by significant performance gains in benchmark tasks.

\textbf{Stability in RL.} Stabilizing deep RL algorithms has been a longstanding challenge, indicated by numerous empirical studies that highlight practical concerns associated with deep RL methods~\citep{henderson2017deep, engstrom2019implementation}. 
Our use of checkpoints is most closely related to stabilizing policy performance via safe policy improvement~\citep{trpo, PPO, laroche2019safe}, as well as the combination of evolutionary algorithms~\citep{salimans2017evolution, mania2018simple} with RL~\citep{khadka2018evolution, pourchot2018cem}, where the checkpoint resembles the fittest individual and the mutation is defined exclusively by the underlying RL algorithm.

\section{Background}

In Reinforcement learning (RL) problems are framed as a Markov decision process (MDP). 
An MDP is a 5-tuple~($S$, $A$, $R$, $p$, $\y$) with state space~$S$, action space~$A$, reward function $R$, dynamics model~$p$, and discount factor~$\y$, where the objective is to find a policy~$\pi: S \rightarrow A$, a mapping from state~$s \in S$ to action~$a \in A$, which maximizes the return~$\sum_{t=1}^\infty \y^{t-1} r_{t}$, the discounted sum of rewards~$r$ obtained when following the policy. RL algorithms commonly use a value function~$Q^\pi(s,a) := \E \lb \sum_{t=1}^\infty \y^{t-1} r_{t} | s_0=s, a_0=a \rb$, which models the expected return, starting from an initial state~$s$ and action~$a$.

\section{State-Action Representation Learning} \label{sec:representation_learning}

\begin{figure}[t]
\pgfdeclarepatternformonly{swnestripes}{\pgfpoint{0cm}{0cm}}{\pgfpoint{1cm}{1cm}}{\pgfpoint{1cm}{1cm}}
{
    \foreach \i in {0.1, 0.3,...,0.9}
    {
    \pgfpathmoveto{\pgfpoint{\i cm}{0cm}}
    \pgfpathlineto{\pgfpoint{1cm}{1cm - \i cm}}
    \pgfpathlineto{\pgfpoint{1cm}{1cm - \i cm + 0.1cm}}
    \pgfpathlineto{\pgfpoint{\i cm - 0.1cm}{0cm}}
    \pgfpathclose%
    \pgfusepath{fill}
    \pgfpathmoveto{\pgfpoint{0cm}{\i cm}}
    \pgfpathlineto{\pgfpoint{1cm - \i cm}{1cm}}
    \pgfpathlineto{\pgfpoint{1cm - \i cm - 0.1cm}{1cm}}
    \pgfpathlineto{\pgfpoint{0cm}{\i cm + 0.1cm}}
    \pgfpathclose%
    \pgfusepath{fill}
    }
}

\pgfdeclarepatternformonly{swneStripes}{\pgfpoint{0cm}{0cm}}{\pgfpoint{1cm}{1cm}}{\pgfpoint{1cm}{1cm}}
{
    \foreach \i in {0.1, 0.3,...,0.9}
    {
     \pgfpathmoveto{\pgfpoint{\i cm}{0cm}}
     \pgfpathlineto{\pgfpoint{1cm}{1cm - \i cm}}
     \pgfpathlineto{\pgfpoint{1cm}{1cm - \i cm - 0.1cm}}
     \pgfpathlineto{\pgfpoint{\i cm + 0.1cm}{0cm}}
     \pgfpathclose%
     \pgfusepath{fill}
     \pgfpathmoveto{\pgfpoint{0cm}{\i cm}}
     \pgfpathlineto{\pgfpoint{1cm - \i cm}{1cm}}
     \pgfpathlineto{\pgfpoint{1cm - \i cm + 0.1cm}{1cm}}
     \pgfpathlineto{\pgfpoint{0cm}{\i cm - 0.1cm}}
     \pgfpathclose%
     \pgfusepath{fill}
    }
}

\pgfdeclarepatternformonly{senwstripes}{\pgfpoint{0cm}{0cm}}{\pgfpoint{1cm}{1cm}}{\pgfpoint{1cm}{1cm}}
{
    \foreach \i in {0.1, 0.3,...,0.9}
    {
     \pgfpathmoveto{\pgfpoint{0cm}{\i cm}}
     \pgfpathlineto{\pgfpoint{0cm}{\i cm + 0.1cm}}
     \pgfpathlineto{\pgfpoint{\i cm + 0.1cm}{0cm}}
     \pgfpathlineto{\pgfpoint{\i cm}{0cm}}
     \pgfpathclose%
     \pgfusepath{fill}
     \pgfpathmoveto{\pgfpoint{1cm}{\i cm}}
     \pgfpathlineto{\pgfpoint{\i cm}{1cm}}
     \pgfpathlineto{\pgfpoint{\i cm + 0.1cm}{1cm}}
     \pgfpathlineto{\pgfpoint{1cm}{\i cm + 0.1cm}}
     \pgfpathclose%
     \pgfusepath{fill}
    }
}

\centering

\begin{tikzpicture}
\small

\node[draw, rectangle, rounded corners, sb_blue, fill=sb_blue, fill opacity=0.25, very thick, minimum width=24pt, minimum height=24pt] at (0, 0) (linear) {};
\node[draw, rectangle, rounded corners, sb_blue, fill=sb_blue, fill opacity=0.25, very thick, minimum width=24pt, minimum height=24pt] at (6, 0) (critic) {};

\node[draw, rectangle, rounded corners, sb_orange, fill=sb_orange, fill opacity=0.25, very thick, minimum width=24pt, minimum height=24pt] at (0, 2) (zs) {};
\node[label={$g_t(z_t^s, a)$}, draw, rectangle, rounded corners, sb_orange, fill=sb_orange, fill opacity=0.25, very thick, minimum width=24pt, minimum height=24pt] at (4, 2) (zsa) {};

\fill[sb_orange, pattern=senwstripes, pattern color=sb_orange_02, very thick] ($(zs)+(-1.25,1)$) rectangle ($(zsa)+(1.5,-0.8)$);

\fill[sb_blue,pattern=swnestripes,pattern color=sb_blue_015, very thick] ($(linear)+(-1.25,1)$) rectangle ($(critic)+(1.15,-0.8)$);

\node[label={$\text{Linear}(s,a)$}, draw, rectangle, rounded corners, sb_blue, fill=sb_blue_025, very thick, minimum width=24pt, minimum height=24pt] at (0, 0) {};

\node[label={$f_t(s)$}, draw, rectangle, rounded corners, sb_orange, fill=sb_orange_025, very thick, minimum width=24pt, minimum height=24pt] at (0, 2) {};
\node[label={$g_t(z_t^s, a)$}, draw, rectangle, rounded corners, sb_orange, fill=sb_orange_025, very thick, minimum width=24pt, minimum height=24pt] at (4, 2) {};

\path[draw=sb_blue, very thick, -{Latex[length=5pt]}, line cap=round] ([yshift=-0.25cm]linear.east) -- node[midway, below] {$\text{AvgL1Norm}(\phi^{sa})$} ([yshift=-0.25cm]critic.west);

\path[draw=sb_orange, very thick, rounded corners, -{Latex[length=5pt]}, line cap=round] (zs.east) -- node[midway, above] {$\text{AvgL1Norm}(z_t^s)$} (3,2) --([yshift=0.25cm]3,2) -- ([yshift=0.25cm]zsa.west);

\path[draw=sb_orange, very thick, rounded corners, -{Latex[length=5pt]}, line cap=round] (zs.east) -- (2,2) -- (3,0) -- (critic.west);

\draw[sb_orange, very thick, rounded corners, -{Latex[length=5pt]}] (zsa.east) -- (4.65,2) -- node[near start, right, text=black] {$z_t^{sa}$} ([yshift=0.25cm]4.65,0) -- ([yshift=0.25cm]critic.west);

\path[draw=black, very thick, -{Latex[length=5pt]}, line cap=round] ([yshift=0.25cm]-1,0) -- node[midway, above] {$s$} ([yshift=0.25cm]linear.west);   
\path[draw=black, very thick, -{Latex[length=5pt]}, line cap=round] ([yshift=-0.25cm]-1,0) -- node[midway, above] {$a$} ([yshift=-0.25cm]linear.west);   
\path[draw=black, very thick, -{Latex[length=5pt]}, line cap=round] (-1,2) -- node[midway, above] {$s$} (zs.west);   
\path[draw=black, very thick, -{Latex[length=5pt]}, line cap=round] ([yshift=-0.25cm]3,2) -- node[midway, above] {$a$} ([yshift=-0.25cm]zsa.west);   
\node[draw, label={$Q_{t+1}(z_t^{sa}{,}z_t^s{,}\phi^{sa})$~~~}, rectangle, rounded corners, sb_blue, fill=sb_blue_025, very thick, minimum width=24pt, minimum height=24pt] at (6, 0) {};

\node[anchor=west] at (-1, -1.05) {\textcolor{sb_blue}{End-to-end}};

\node[anchor=west] at (-1, 3.2) {\textcolor{sb_orange}{Decoupled}};

\draw[dotted, very thick] (7.25,-0.8) to (7.25,3);

\node[anchor=mid] at (3, -1.5) {(a) Value function~$Q$ with SALE};

\end{tikzpicture}
\begin{tikzpicture}
\small

\node[draw, rectangle, rounded corners, sb_blue, fill=sb_blue, fill opacity=0.25, very thick, minimum width=24pt, minimum height=24pt] at (8.5, 0) (linearpi) {};
\node[draw, rectangle, rounded corners, sb_blue, fill=sb_blue, fill opacity=0.25, very thick, minimum width=24pt, minimum height=24pt] at (11.5, 0) (pi) {};

\node[draw, rectangle, rounded corners, sb_orange, fill=sb_orange, fill opacity=0.25, very thick, minimum width=24pt, minimum height=24pt] at (8.5, 2) (zspi) {};

\fill[sb_orange, pattern=senwstripes, pattern color=sb_orange_02, very thick] ($(zspi)+(-1.25,1)$) rectangle ($(zspi)+(1.25,-0.8)$);

\fill[sb_blue,pattern=swnestripes,pattern color=sb_blue_015, very thick] ($(linearpi)+(-1.25,1)$) rectangle ($(pi)+(0.9,-0.8)$);

\node[label={$\text{Linear}(s)$}, draw, rectangle, rounded corners, sb_blue, fill=sb_blue_025, very thick, minimum width=24pt, minimum height=24pt] at (8.5, 0) {};
\node[label={$\pi_{t+1}(z_t^s, \phi^s)$}, draw,rectangle, rounded corners, sb_blue, fill=sb_blue_025, very thick, minimum width=24pt, minimum height=24pt] at (11.5, 0) {};
\node[label={$f_t(s)$}, draw, rectangle, rounded corners, sb_orange, fill=sb_orange_025, very thick, minimum width=24pt, minimum height=24pt] at (8.5, 2) {};

\path[draw=sb_blue, very thick, -{Latex[length=5pt]}, line cap=round] ([yshift=-0.25cm]linearpi.east) -- ([yshift=-0.25cm]pi.west);
\node[anchor=mid] at (10, -0.6) {$\text{AvgL1Norm}(\phi^s)$};

\node[anchor=mid] at (11,1.4) {$\text{AvgL1Norm}(z_t^s)$};

\draw[sb_orange, very thick, rounded corners, -{Latex[length=5pt]}] (zspi.east) -- (9.25,2) -- ([yshift=0.25cm]10.25,0) -- ([yshift=0.25cm]pi.west);

\path[draw=black, very thick, -{Latex[length=5pt]}, line cap=round] (7.5,0) -- node[midway, above] {$s$} (linearpi.west);   
\path[draw=black, very thick, -{Latex[length=5pt]}, line cap=round] (7.5,2) -- node[midway, above] {$s$} (zspi.west);    

\node[anchor=west] at (7.5, -1.05) {\textcolor{sb_blue}{End-to-end}};

\node[anchor=west] at (7.5, 3.2) {\textcolor{sb_orange}{Decoupled}};

\node[anchor=mid] at (10, -1.5) {(b) Policy~$\pi$ with SALE};

\end{tikzpicture}

\caption[]{\textbf{Diagram of State-Action Learned Embeddings (SALE).} SALE uses encoders~$(f,g)$ to output embeddings~$(z^s, z^{sa})$ to enhance the input of the value function~$Q$ and policy~$\pi$. $\phi$ denotes the output of the corresponding linear layer. \cblock{sb_orange} The encoders~$(f,g)$ are jointly trained to predict the next state embedding (where $|\cdot|_\times$ denotes the stop-gradient operation), %
decoupled from the training of the value function and policy~(\autoref{eqn:encoder_loss}). \cblock{sb_blue} The end-to-end linear layers are trained with gradients from the corresponding network. AvgL1Norm is used to keep the scale of each of the inputs to the value function and policy constant. 
} \label{fig:representation}
\vspace{-8pt}
\end{figure}
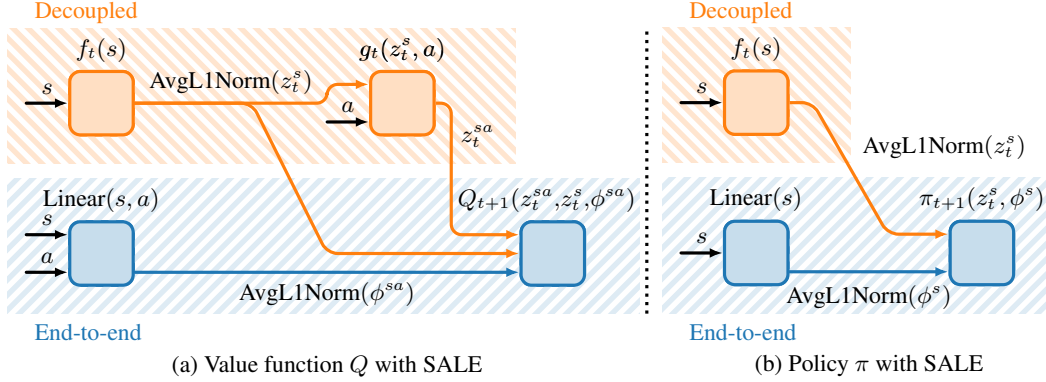

In this section, we introduce state-action learned embeddings~(SALE)~(\autoref{fig:representation}). We begin with the basic outline of SALE and then discuss 
three important considerations in how SALE is implemented. %
We then perform an extensive empirical evaluation on the design space to highlight the critical choices when learning embeddings from the dynamics of the environment. 

\subsection{State-Action Learned Embeddings} \label{subsec:SALE}

The objective of SALE is to discover learned embeddings $(z^{sa}, z^s)$ which capture relevant structure in the observation space, as well as the transition dynamics of the environment. To do so, SALE utilizes a pair of encoders $(f,g)$ where $f(s)$ encodes the state~$s$ into the state embedding $z^s$ and $g(z^s,a)$ jointly encodes both state~$s$ and action~$a$ into the state-action embedding $z^{sa}$:
\begin{align}
	z^s := f(s), \qquad z^{sa} := g(z^s,a).
\end{align}
The embeddings are split into state and state-action components so that the encoders can be trained with a dynamics prediction loss that solely relies on the next state~$s'$, 
independent of the next action or current policy. As a result, the encoders are jointly trained using the mean squared error~(MSE) between the state-action embedding~$z^{sa}$ and the embedding of the next state~$z^{s'}$: 
\begin{align} \label{eqn:encoder_loss}
	\Loss(f,g) := \Bigl( g(f(s),a) - |f(s')|_\times \Bigr)^2 = \lp z^{sa} - |z^{s'}|_\times \rp^2,
\end{align}
where $|\cdot|_\times$ denotes the stop-gradient operation. 
The embeddings are designed to model the underlying structure of the environment. However, they may not encompass all relevant information needed by the value function and policy, such as features related to the reward, current policy, or task horizon. Accordingly, we concatenate the embeddings with the original state and action, allowing the value and policy networks to learn relevant internal representations for their respective tasks:
\begin{align} \label{eqn:input}
	Q(s,a) \rightarrow Q(z^{sa}, z^s, s, a), \qquad \pi(s) \rightarrow \pi(z^s, s).
\end{align}
The encoders~$(f,g)$ are trained online and concurrently with the RL agent (updated at the same frequency as the value function and policy), but are decoupled (gradients from the value function and policy are not propagated to $(f,g)$). 
Although the embeddings are learned by considering the dynamics of the environment, their purpose is solely to improve the input to the value function and policy, and not to serve as a world model for planning or estimating rollouts.

There are three additional considerations in how SALE is implemented in practice. 

\textbf{Normalized embeddings.} The minimization of distances in embedding space can result in instability due to either monotonic growth or collapse to a redundant representation~\citep{gelada2019deepmdp}. To combat this risk, we introduce AvgL1Norm, a normalization layer that divides the input vector by its average absolute value in each dimension, thus keeping the relative scale of the embedding constant throughout learning. Let $x_i$ be the $i$-th dimension of an $N$-dimensional vector $x$, then
\begin{equation} \label{eqn:AvgL1Norm}
	\text{AvgL1Norm}(x) := \frac{x}{\frac{1}{N} \sum_i |x_i|}.
\end{equation}
AvgL1Norm is applied to the state embedding $z^s$. 
Similar to the normalized loss functions used by SPR~\citep{schwarzer2020data} and BYOL~\citep{grill2020bootstrap}, AvgL1Norm protects from monotonic growth, but also keeps the scale of the downstream input constant without relying on updating statistics (e.g. BatchNorm~\citep{ioffe2015batch}).
This is important for our approach as the embeddings are trained independently from the value function and policy. 
AvgL1Norm is not applied to the state-action embedding~$z^{sa}$, as it is trained to match the normalized next state embedding~$z^{s'}$.

We also apply AvgL1Norm to the state and action inputs (following a linear layer) to the value function~$Q$ and policy~$\pi$, to keep them at a similar scale to the learned embeddings. The input to the value function and policy then becomes:
\begin{equation} \label{eqn:norm_input}
	Q(z^{sa}, z^{s}, \text{AvgL1Norm}(\text{Linear}(s,a))), \qquad \pi(z^s, \text{AvgL1Norm}(\text{Linear}(s))).
\end{equation}
Unlike the embeddings $(z^s, z^{sa})$, these linear layers are learned end-to-end, and can consequently be viewed as an addition to the architecture of the value function or policy. 

\textbf{Fixed embeddings.}
Since an inconsistent input can cause instability, we freeze the embeddings used to train the current value and policy networks. This means at the iteration $t+1$, the input to the current networks~$(Q_{t+1}, \pi_{t+1})$ uses embeddings~$(z^{sa}_t, z^s_t)$ from the encoders~$(f_t,g_t)$ at the previous iteration $t$. The value function and policy are thus updated by: 
\begin{align}
	Q_{t+1}(z_{t}^{sa}, z_{t}^s, s, a) &\approx r + \y Q_{t} (z_{t-1}^{s'a'}, z_{t-1}^{s'}, s', a'), \qquad \text{ where } a' \sim \pi_{t}(z^{s'}_{t-1}, s'), \label{eqn:critic_loss} \\ 
	\pi_{t+1}(z_{t}^s, s) &\approx \argmax_{\pi} Q_{t+1}(z_{t}^{sa}, z_{t}^s, s, a), \qquad \text{ where } a \sim \pi(z_{t}^s, s).\label{eqn:actor_loss}
\end{align}
The current value function~$Q_{t+1}$ is also trained with respect to the previous value function~$Q_t$, known as a target network~\citep{DQN}. 
The current embeddings $z^s_{t+1}$ and $z^{sa}_{t+1}$ are trained with \autoref{eqn:encoder_loss}, using a target $z^{s'}_{t+1}$ (hence, without a target network). Every $n$ steps the iteration is incremented and all target networks are updated simultaneously: 
\begin{equation}\label{eqn:target_networks}
	Q_t \leftarrow Q_{t+1}, \qquad \pi_t \leftarrow \pi_{t+1}, \qquad (f_{t-1},g_{t-1}) \leftarrow (f_t,g_t), \qquad (f_t,g_t) \leftarrow (f_{t+1},g_{t+1}).
\end{equation}

\begin{figure}[t]
\centering
\hspace{12pt}
\begin{minipage}{0.5\textwidth}
\begin{tikzpicture}[trim axis right, trim axis left]
\begin{axis}[
    width=0.4\textwidth,
    title={\shortstack{Un-clipped\\performance by seed}},
    xlabel={Time steps (1M)},
    ylabel={Total Reward (1k)},
    xtick={0, 1, 2, 3, 4, 5},
    xticklabels={0, 1, 2, 3, 4, 5},
    ytick={-3, 0, 3, 6, 9, 12},
    yticklabels={-3, 0, 3, 6, 9, 12},
]
\addplot graphics [
ymin=-3.45, ymax=12.45,
xmin=-0.15, xmax=5.15,
]{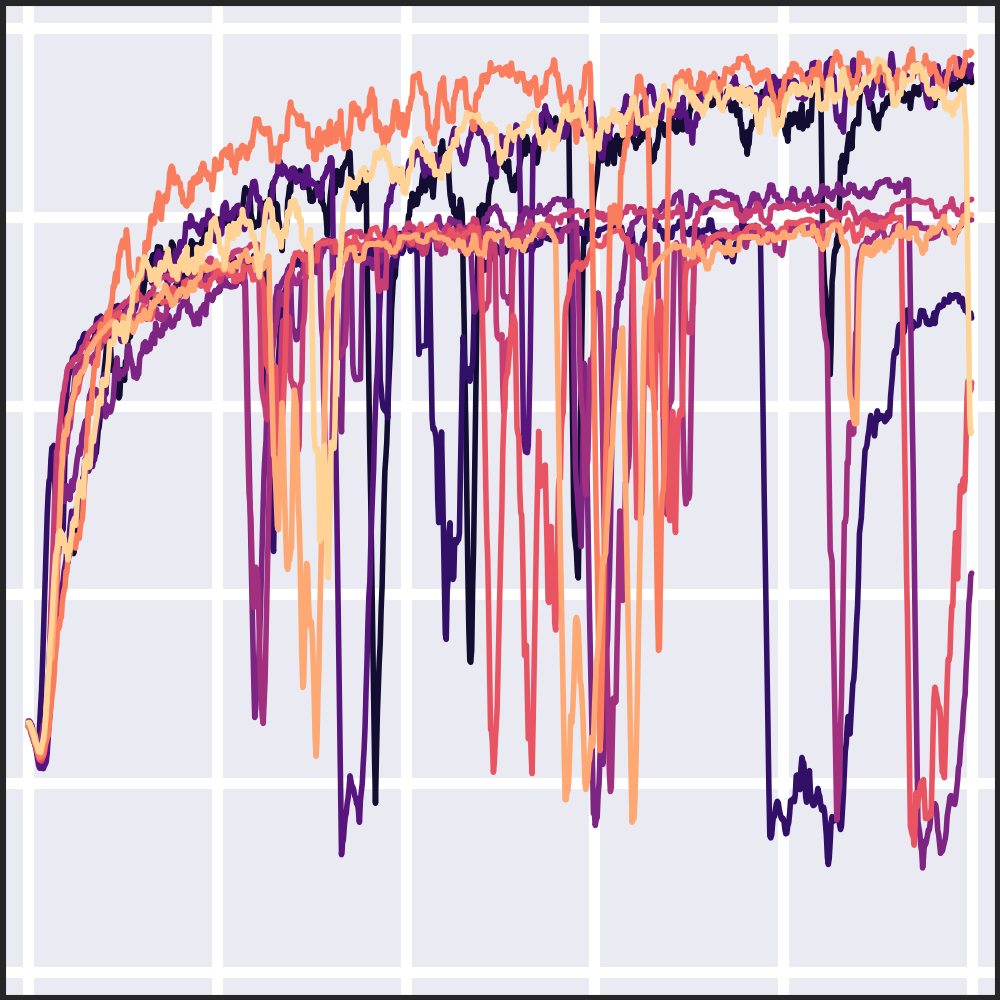};
\end{axis}
\end{tikzpicture}
\hspace{1pt}
\begin{tikzpicture}[trim axis right]
\begin{axis}[
    width=0.4\textwidth,
    title={\shortstack{Un-clipped\\value estimate by seed}},
    xlabel={Time steps (1M)},
    ylabel={Value Estimate (1k)},
    xtick={0, 1, 2, 3, 4, 5},
    xticklabels={0, 1, 2, 3, 4, 5},
    ytick={0, 1, 2, 3, 4},
    yticklabels={0, 1, 2, 3, 4},
]
\addplot graphics [
ymin=-0.06, ymax=2.06,
xmin=-0.15, xmax=5.15,
]{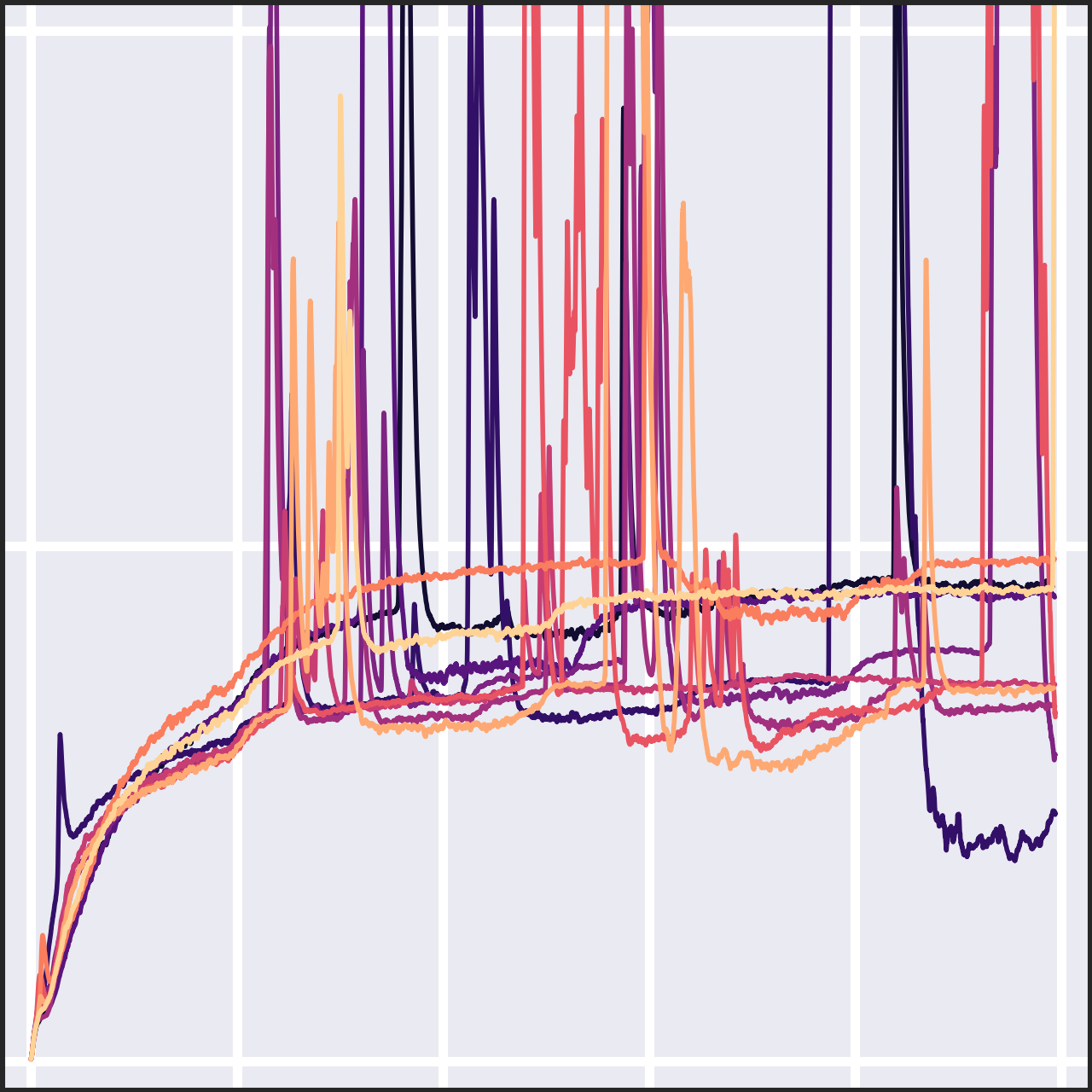};
\end{axis}
\end{tikzpicture}
\end{minipage}
\hspace{-12pt}
\begin{minipage}{0.2\textwidth}
\begin{tikzpicture}[trim axis right]
\begin{axis}[
    width=\textwidth,
    height=0.35\textwidth,
    title={ \vphantom{\shortstack{Unclipped\\by}}No $z^{sa}$\vphantom{$\phi^{sa}$}},
    xtick={0, 1, 2, 3, 4, 5},
    xticklabels={},
    ytick={1,2},
    yticklabels={1,2},
]
\addplot graphics [
ymin=-0.06, ymax=2.06,
xmin=-0.15, xmax=5.15,
]{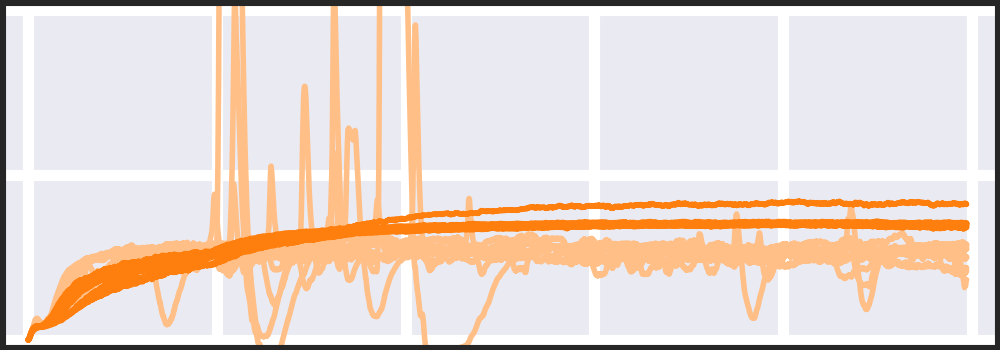};
\end{axis}
\end{tikzpicture}
\vspace{-19pt}

\begin{tikzpicture}[trim axis right]
\begin{axis}[
    width=\textwidth,
    height=0.35\textwidth,
    title={Small $\phi^{sa}$\vphantom{p}},
    xlabel={Time steps (1M)},
    xtick={0, 1, 2, 3, 4, 5},
    xticklabels={0, 1, 2, 3, 4, 5},
    ytick={1,2},
    yticklabels={1,2},
]
\addplot graphics [
ymin=-0.06, ymax=2.06,
xmin=-0.15, xmax=5.15,
]{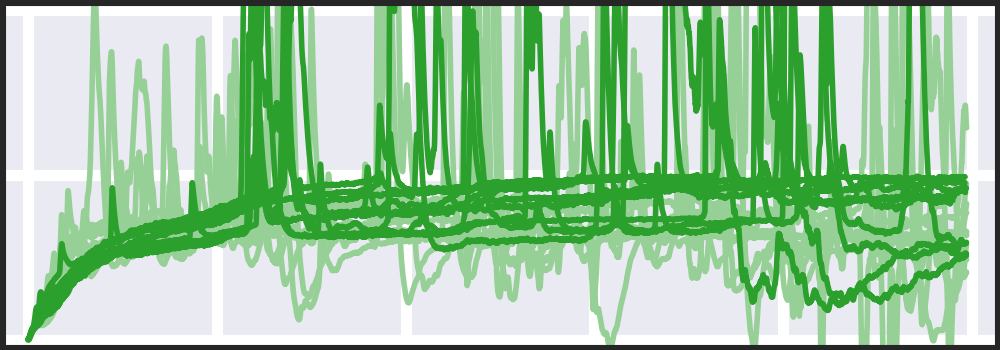};
\end{axis}
\end{tikzpicture}
\end{minipage}
\hspace{5pt}
\begin{minipage}{0.2\textwidth}
\begin{tikzpicture}[trim axis right]
\begin{axis}[
    width=\textwidth,
    height=0.35\textwidth,
    title={\vphantom{\shortstack{Unclipped\\by}}No $z^{sa}$ and small $\phi^{sa}$},
    xtick={0, 1, 2, 3, 4, 5},
    xticklabels={},
    ytick={1,2},
    yticklabels={1,2},
]
\addplot graphics [
ymin=-0.06, ymax=2.06,
xmin=-0.15, xmax=5.15,
]{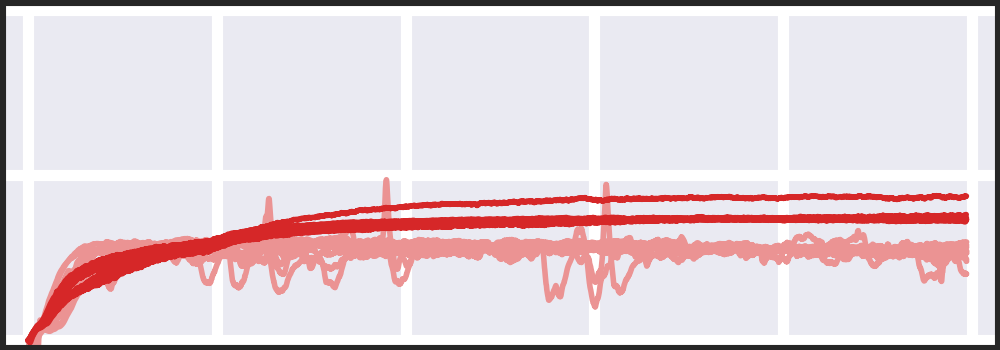};
\end{axis}
\end{tikzpicture}
\vspace{-19pt}

\begin{tikzpicture}[trim axis right]
\begin{axis}[
    width=\textwidth,
    height=0.35\textwidth,
    title={Clipped},
    xlabel={Time steps (1M)},
    xtick={0, 1, 2, 3, 4, 5},
    xticklabels={0, 1, 2, 3, 4, 5},
    ytick={1,2},
    yticklabels={1,2},
]
\addplot graphics [
ymin=-0.06, ymax=2.06,
xmin=-0.15, xmax=5.15,
]{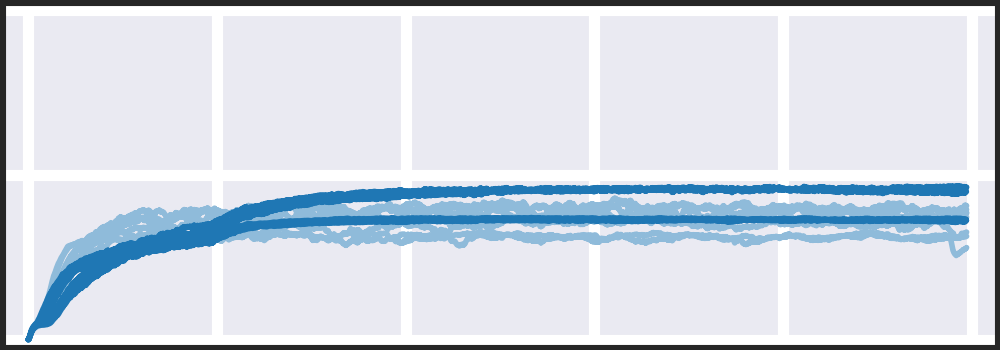};
\end{axis}
\end{tikzpicture}
\end{minipage}

\hspace{44pt}
\fcolorbox{gray}{gray!10}{
\small
\begin{tabular}{@{\hspace{2pt}}l@{}l@{}}
    \cblockfull{m_1}~\cblockfull{m_2}~\cblockfull{m_3}~\cblockfull{m_4}~\cblockfull{m_5}~\cblockfull{m_6}~\cblockfull{m_7}~\cblockfull{m_8}~\cblockfull{m_9}~\cblockfull{m_10}~ & Seeds (1-10) \\
    & Buffer: 1M
    \end{tabular}
}
\hfill
\fcolorbox{gray}{gray!10}{
\small
\begin{tabular}{@{\hspace{2pt}}ll@{}}
    Buffer: \cblockfull{sb_orange}~1M~~\cblockfull{sb_orange!50}~100k %
    & Buffer: \cblockfull{sb_red}~1M~~\cblockfull{sb_red!50}~100k \\%~No $z^{sa}$ and small $\phi$\\
    Buffer: \cblockfull{sb_green}~1M~~\cblockfull{sb_green!50}~100k %
    & Buffer: \cblockfull{sb_blue}~1M~~\cblockfull{sb_blue!50}~100k %
\end{tabular}
}
\caption[]{%
\textbf{Extrapolation error can occur in online RL when using state-action representation learning.} All figures use the Ant environment. $\phi^{sa}$ corresponds to the output of the linear layer $(\text{Linear}(s,a)=\phi^{sa})$ (\autoref{eqn:norm_input}). Both embeddings and $\phi^{sa}$ have a default dimension size of $256$. Small $\phi^{sa}$ means that $\text{Dim}(\phi^{sa})$ is set to $16$. No $z^{sa}$ means the value function input is $Q(z^s,s,a)$. 
\cblockfull{m_2}~\cblockfull{m_5}~\cblockfull{m_9}~The default performance and value estimate of 10 individual seeds without value clipping. While the performance trends upwards there are large dips in reward, which correspond with jumps in the estimated value. \cblockfull{sb_orange}~\cblockfull{sb_orange!50}~/~\cblockfull{sb_green}~\cblockfull{sb_green!50}~/~\cblockfull{sb_red}~\cblockfull{sb_red!50} Varying the input dimension can improve or harm stability of the value estimate. The severity is impacted by the replay buffer size (1M or 100k). 
The state embedding~$z^{s}$ is left unchanged in all settings, showing that the state-action embedding~$z^{sa}$ and the linear layer over the state-action input~$\phi^{sa}$ are the primary contributors to the extrapolation error. This shows the potential negative impact from increasing the dimension size of an input which relies on a potentially unseen action.
\cblockfull{sb_blue}~\cblockfull{sb_blue!50} Clipping stabilizes the value estimate, without modifying the input dimension size~(\autoref{eqn:clipped}). 
}
    \label{fig:extrapolation}
\vspace{-8pt}
\end{figure}

\textbf{Clipped Values.} Extrapolation error is the tendency for deep value functions to extrapolate to unrealistic values on state-actions pairs which are rarely seen in the dataset~\citep{fujimoto2019off}. Extrapolation error has a significant impact in offline RL, where the RL agent learns from a given dataset rather than collecting its own experience, as the lack of feedback on overestimated values can result in divergence. 

Surprisingly, we observe a similar phenomenon in online RL, when increasing the number of dimensions in the state-action input to the value function, as illustrated in \autoref{fig:extrapolation}. Our hypothesis is that the state-action embedding $z^{sa}$ expands the action input and makes the value function more likely to over-extrapolate on unknown actions. We show in \autoref{fig:extrapolation} that the dimension size of $z^{sa}$ as well as the state-action input plays an important role in the stability of value estimates. 

Fortunately, extrapolation error can be combated in a straightforward manner in online RL, where poor estimates are corrected by feedback from interacting with the environment. Consequently, we only need to stabilize the value estimate until the correction occurs. 
This can be achieved in SALE by tracking the range of values in the dataset $D$ (estimated over sampled mini-batches during training), and then bounding the target used in \autoref{eqn:critic_loss} by the range:
\begin{align}\label{eqn:clipped}
	Q_{t+1}(s,a) \approx r + \gamma \text{ clip} \lp Q_t(s',a'), \min_{(s,a) \in D} Q_t(s,a), \max_{(s,a) \in D} Q_t(s,a) \rp.
\end{align}
Additional discussion of extrapolation error, experimental details, and ablation of the proposed value clipping in SALE can be found in \autoref{appendix:sec:extrapolation} \& \ref{appendix:sec:design}. %

\subsection{Evaluating Design Choices} \label{subsec:design}

The effectiveness of learning embeddings by modeling the dynamics of the environment is a natural consequence of the relationship between the value function and future states. However, there are many design considerations for which all alternatives are potentially valid and the approach adopted differs among related methods in the literature.
In this section, we perform an extensive study over the design space to (1) show SALE uses the correct and highest performing set of choices, and (2) better understand which choices are the biggest contributors to performance when using SALE. %

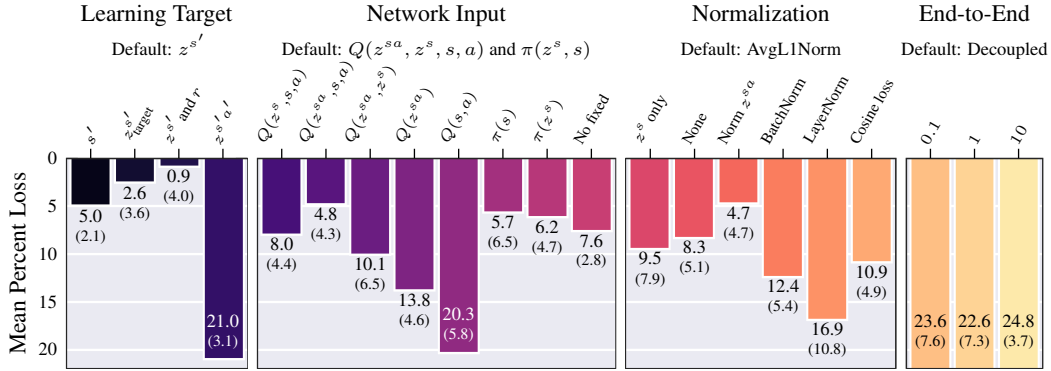
\begin{figure}[t]
\centering
\begin{tikzpicture}[trim axis right]
\begin{axis}[
    height=0.2\textwidth,
    width=0.172\textwidth,
    title style={yshift=36pt},
    title={\shortstack{\vphantom{p}Learning Target\\{\scriptsize \vphantom{$(z^{s'}$} Default: $z^{s'}$}}},
    ylabel={Mean Percent Loss},
    ytick={-20, -15, -10, -5, 0},
    yticklabels={20, 15, 10, 5, 0},
    xtick={0, 1, 2, 3},
    xticklabels={$s'$, $z^{s'}_\text{target}$, {\tiny $z^{s'}$and $r$}, $z^{s'a'}$},
    xticklabel style={rotate=60, font=\tiny},
    xticklabel pos=right,
    clip=false,
]
\addplot graphics [
xmin=-0.55, xmax=3.55,
ymin=-22, ymax=0,
]{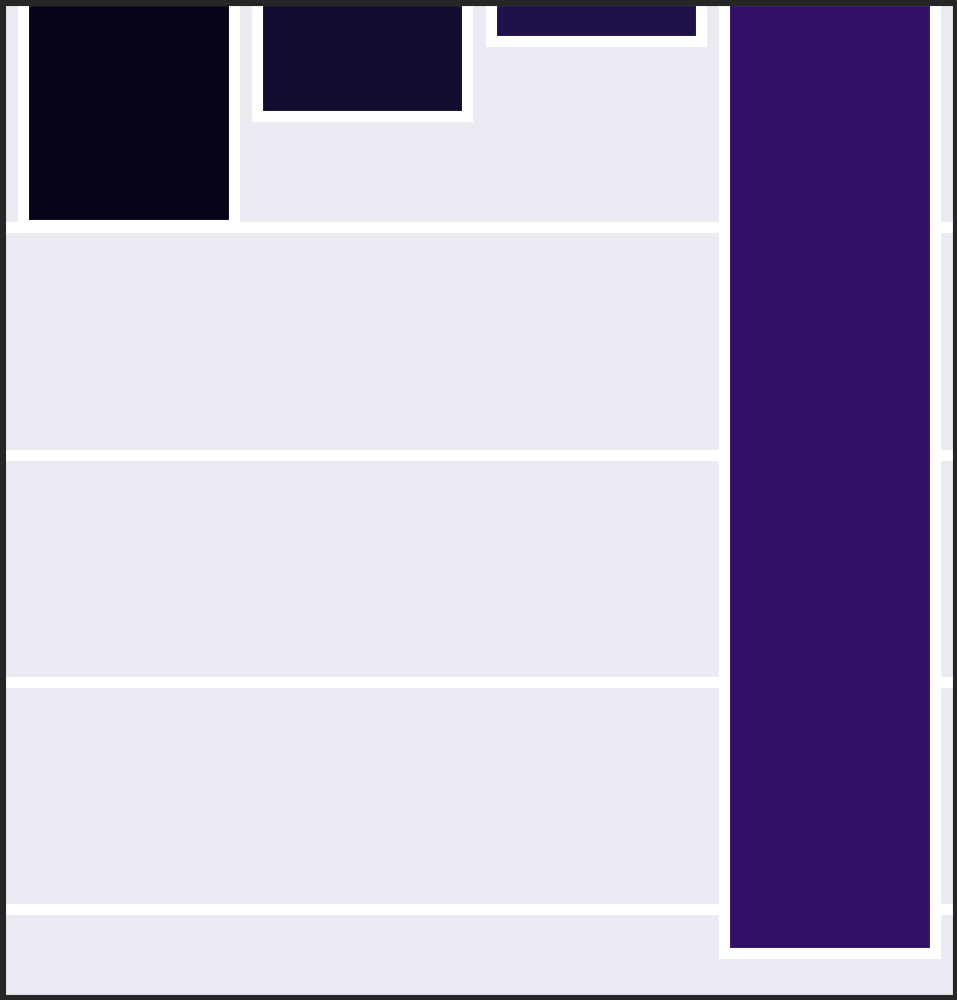};
\node[anchor=mid] at (0, -8.0) {\shortstack{\scriptsize 5.0\\{\tiny (2.1)}}};
\node[anchor=mid] at (1, -5.6) {\shortstack{\scriptsize 2.6\\{\tiny (3.6)}}};
\node[anchor=mid] at (2, -3.9) {\shortstack{\scriptsize 0.9\\{\tiny (4.0)}}};
\node[anchor=mid] at (3, -19.0) {\color{white} \shortstack{\scriptsize 21.0\\{\tiny (3.1)}}};
\draw[line width=0.5pt] (0,0) -- (0,1.0);
\draw[line width=0.5pt] (1,0) -- (1,1.0);
\draw[line width=0.5pt] (2,0) -- (2,1.0);
\draw[line width=0.5pt] (3,0) -- (3,1.0);
\end{axis}
\end{tikzpicture}
\hspace{-8pt}
\begin{tikzpicture}[trim axis right]
\begin{axis}[
    height=0.2\textwidth,
    width=0.341\textwidth,
    title style={yshift=36pt},
    title={\shortstack{Network Input\\{\scriptsize \vphantom{$(z^{s'}$}Default: $Q(z^{sa}, z^s, s, a)$ and $\pi(z^s, s)$}}},
    ytick={-15, -10, -5},
    yticklabels={},
    xtick={0, 1, 2, 3, 4, 5, 6, 7, 8},
    xticklabels={
    {\tiny $Q(z^s{,}s{,}a)$}, %
    {\tiny $Q(z^{sa}{,}s{,}a)$}, %
    {\tiny $Q(z^{sa}{,}z^s)$},
    {\tiny $Q(z^{sa})$},%
    {\tiny $Q(s{,}a)$},%
    $\pi(s)$, 
    $\pi(z^s)$,
    {\tiny No fixed}, 
    },
    xticklabel style={rotate=60, font=\tiny},
    xticklabel pos=right,
    clip=false,
]
\addplot graphics [
xmin=-0.55, xmax=7.55,
ymin=-22, ymax=0,
]{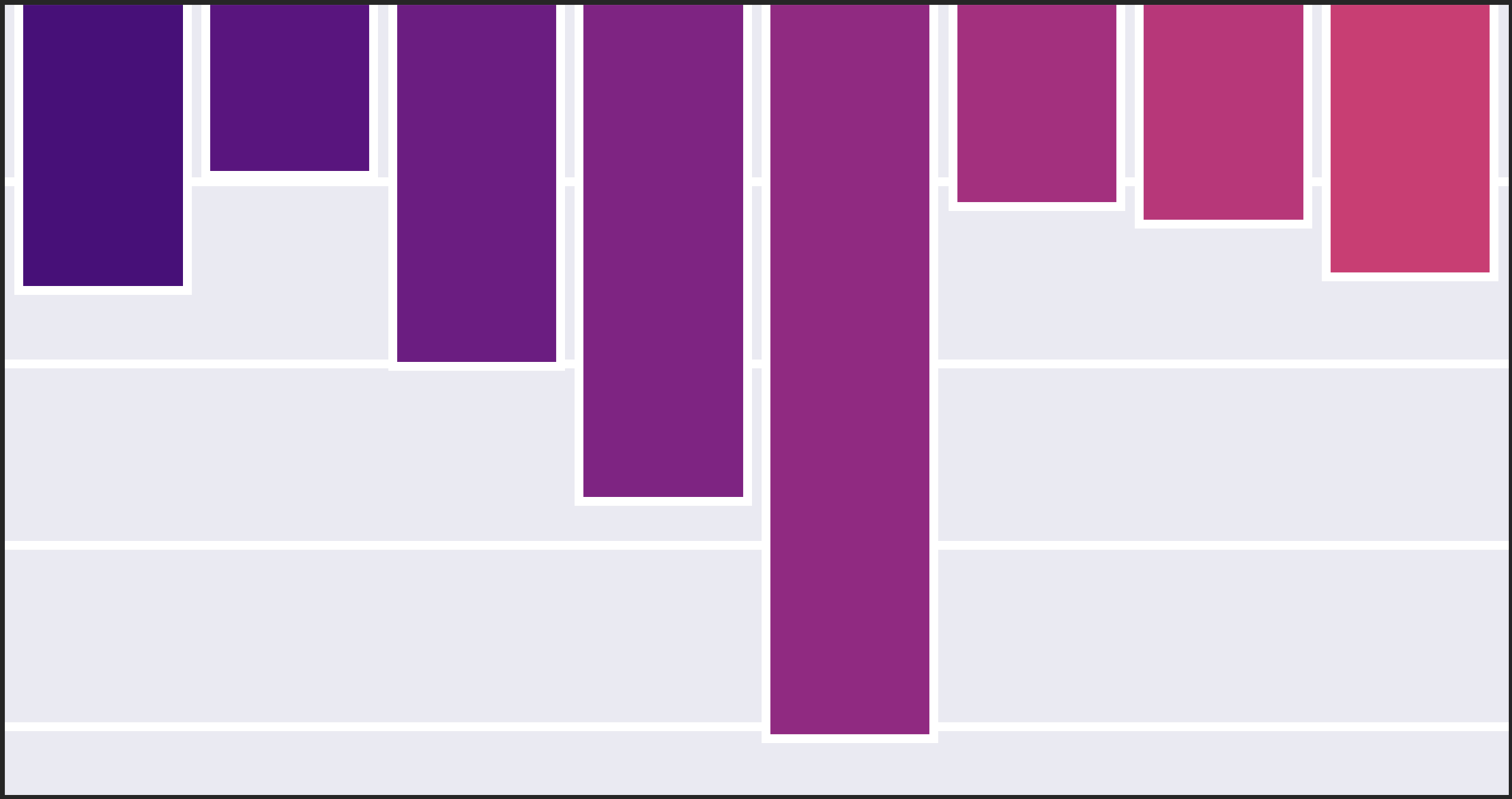};
\node[anchor=mid] at (0, -11.0) {\shortstack{\scriptsize 8.0\\{\tiny (4.4)}}};
\node[anchor=mid] at (1, -7.8) {\shortstack{\scriptsize 4.8\\{\tiny (4.3)}}};
\node[anchor=mid] at (2, -13.1) {\shortstack{\scriptsize 10.1\\{\tiny (6.5)}}};
\node[anchor=mid] at (3, -16.8) {\shortstack{\scriptsize 13.8\\{\tiny (4.6)}}};
\node[anchor=mid] at (4, -18.3) {\color{white} \shortstack{\scriptsize 20.3\\{\tiny (5.8)}}};
\node[anchor=mid] at (5, -8.7) {\shortstack{\scriptsize 5.7\\{\tiny (6.5)}}};
\node[anchor=mid] at (6, -9.2) {\shortstack{\scriptsize 6.2\\{\tiny (4.7)}}};
\node[anchor=mid] at (7, -10.6) {\shortstack{\scriptsize 7.6\\{\tiny (2.8)}}};
\draw[line width=0.5pt] (0,0) -- (0,1.0);
\draw[line width=0.5pt] (1,0) -- (1,1.0);
\draw[line width=0.5pt] (2,0) -- (2,1.0);
\draw[line width=0.5pt] (3,0) -- (3,1.0);
\draw[line width=0.5pt] (4,0) -- (4,1.0);
\draw[line width=0.5pt] (5,0) -- (5,1.0);
\draw[line width=0.5pt] (6,0) -- (6,1.0);
\draw[line width=0.5pt] (7,0) -- (7,1.0);
\end{axis}
\end{tikzpicture}
\hspace{-8pt}
\begin{tikzpicture}[trim axis right]
\begin{axis}[
    height=0.2\textwidth,
    width=0.257\textwidth,
    title style={yshift=36pt},
    title={\shortstack{\vphantom{p}Normalization\\{\scriptsize \vphantom{$(z^{s'}$}Default: AvgL1Norm}}},
    ytick={-15, -10, -5},
    yticklabels={},
    xtick={0, 1, 2, 3, 4, 5},
    xticklabels={$z^s$ only, None, Norm $z^{sa}$, {\tiny BatchNorm}, {\tiny LayerNorm}, {\tiny Cosine loss}},
    xticklabel style={rotate=60, font=\tiny},
    xticklabel pos=right,
    clip=false,
]
\addplot graphics [
xmin=-0.55, xmax=5.55,
ymin=-22, ymax=0,
]{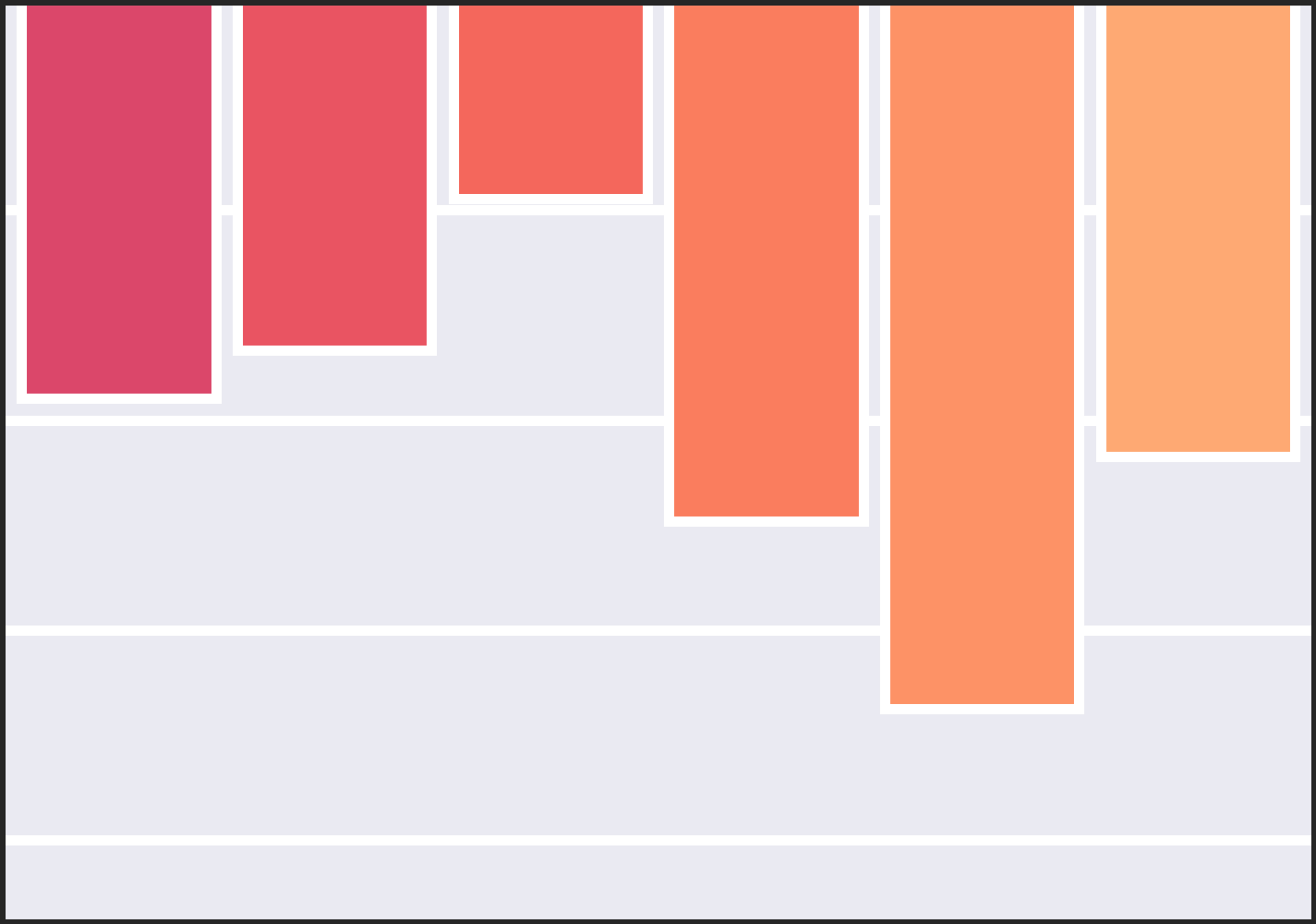};
\node[anchor=mid] at (0, -12.5) {\shortstack{\scriptsize 9.5\\{\tiny (7.9)}}};
\node[anchor=mid] at (1, -11.3) {\shortstack{\scriptsize 8.3\\{\tiny (5.1)}}};
\node[anchor=mid] at (2, -7.7) {\shortstack{\scriptsize 4.7\\{\tiny (4.7)}}};
\node[anchor=mid] at (3, -15.4) {\shortstack{\scriptsize 12.4\\{\tiny (5.4)}}};
\node[anchor=mid] at (4, -19.9) {\shortstack{\scriptsize 16.9\\{\tiny (10.8)}}};
\node[anchor=mid] at (5, -13.9) {\shortstack{\scriptsize 10.9\\{\tiny (4.9)}}};
\draw[line width=0.5pt] (0,0) -- (0,1.0);
\draw[line width=0.5pt] (1,0) -- (1,1.0);
\draw[line width=0.5pt] (2,0) -- (2,1.0);
\draw[line width=0.5pt] (3,0) -- (3,1.0);
\draw[line width=0.5pt] (4,0) -- (4,1.0);
\draw[line width=0.5pt] (5,0) -- (5,1.0);
\end{axis}
\end{tikzpicture}
\hspace{-8pt}
\begin{tikzpicture}[trim axis right]
\begin{axis}[
    height=0.2\textwidth,
    width=0.13\textwidth,
    title style={yshift=36pt},
    title={\shortstack{\vphantom{p}End-to-End\\{\scriptsize \vphantom{$(z^{s'}$}Default: Decoupled}}},
    ytick={-15, -10, -5},
    yticklabels={},
    xtick={0, 1, 2},
    xticklabels={$0.1$, $1$, $10$},
    xticklabel style={rotate=60, font=\tiny},
    xticklabel pos=right,
    clip=false,
]
\addplot graphics [
xmin=-0.55, xmax=2.55,
ymin=-22, ymax=0,
]{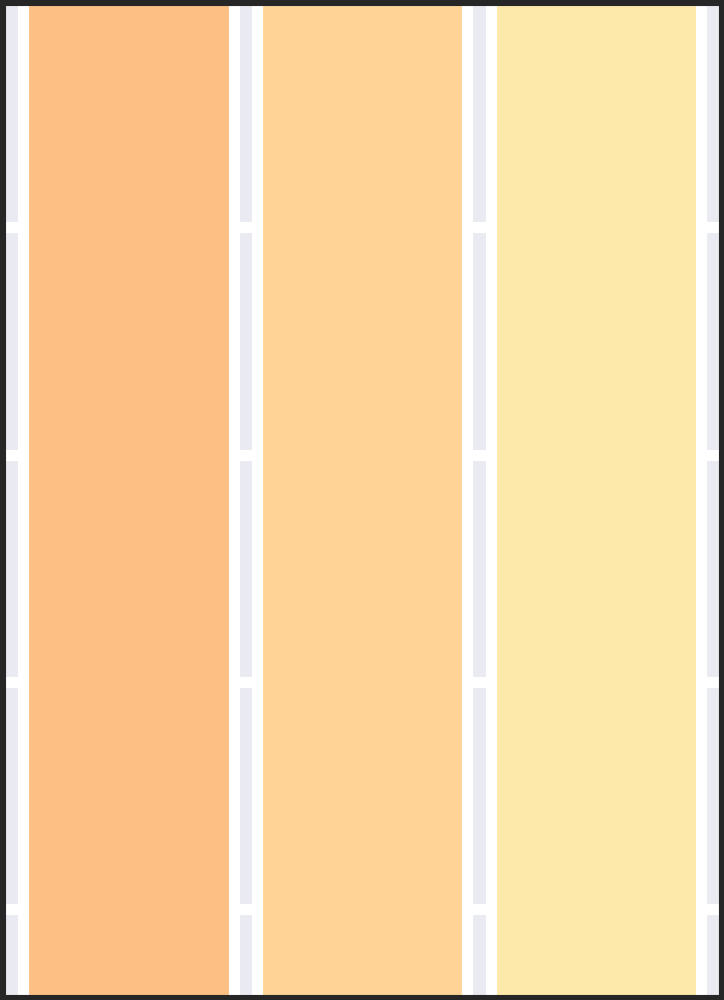};
\node[anchor=mid] at (0, -19) {\shortstack{\scriptsize 23.6\\{\tiny (7.6)}}};
\node[anchor=mid] at (1, -19) {\shortstack{\scriptsize 22.6\\{\tiny (7.3)}}};
\node[anchor=mid] at (2, -19) {\shortstack{\scriptsize 24.8\\{\tiny (3.7)}}};
\draw[line width=0.5pt] (0,0) -- (0,1.0);
\draw[line width=0.5pt] (1,0) -- (1,1.0);
\draw[line width=0.5pt] (2,0) -- (2,1.0);
\end{axis}
\end{tikzpicture}
\caption{The mean percent loss from using alternate design choices in TD7 at 1M time steps, over 10 seeds and the five benchmark MuJoCo environments. Bracketed values describe the range of the 95\% confidence interval around the mean. Percent loss is computed against TD7 where the default choices correspond to a percent loss of $0$. 
See \autoref{subsec:design} for a description of each design choice and key observations. See the Appendix for further implementation-level details.} \label{fig:design}
\vspace{-8pt}
\end{figure}

In \autoref{fig:design} we display the mean percent loss when modifying SALE in the TD7 algorithm (to be fully introduced in \autoref{subsec:TD7}). The percent loss is determined from the average performance at 1M time steps, over 10 seeds and five benchmark environments (HalfCheetah, Hopper, Walker2d, Ant, Humanoid)~\citep{OpenAIGym}. A more detailed description of each variation and complete learning curves can be found in \autoref{appendix:sec:design}.%

\textbf{Learning target.} TD7 trains the encoders by minimizing the MSE between the state-action embedding~$z^{sa}$ and a learning target of the next state embedding $z^{s'}$~(\autoref{eqn:encoder_loss}). We test several alternate learning targets. OFENet uses the next state~$s'$ as the target~\citep{ota2020can} while SPR~\citep{schwarzer2020data} uses the embedding $z_\text{target}^{s'}$ from a target network obtained with an exponential moving average with weight $0.01$. Drawing inspiration from Bisimulation metrics~\citep{ferns2011bisimulation}, DeepMDP~\citep{gelada2019deepmdp} use an objective that considers both the next state embedding $z^{s'}$ and the reward~$r$. We test including a prediction loss on the reward by having the encoder~$g$ output both $z^{sa}$ and $r^\text{pred}$ where $r^\text{pred}$ is trained with the MSE to the reward~$r$. Finally, we test the next state-action embedding $z^{s'a'}$ as the target, where the action~$a'$ is sampled from the target policy. 

\begin{tcolorbox}[notitle,boxrule=0pt,colback=gray!10,colframe=gray!10, left=2pt, right=2pt, top=2pt, bottom=2pt]
	$\Rightarrow$ All learning targets based on the next state~$s'$ perform similarly, although using the embedding~$z^{s'}$ further improves the performance. On the contrary, the next state-action embedding~$z^{s'a'}$ performs much worse as a target, highlighting that signal based on the non-stationary policy can harm learning. Including the reward as a signal has little impact on performance. 
\end{tcolorbox}

\textbf{Network input.} In our approach, the learned embeddings $(z^{sa}, z^s)$ are appended to the state and action input to the value function~$Q(z^{sa}, z^s, s, a)$ and policy~$\pi(z^s, s)$~(\autoref{eqn:input}). We attempt different combinations of input to both networks. We also evaluate replacing the fixed embeddings~(Equations~\ref{eqn:critic_loss}~\&~\ref{eqn:actor_loss}), with the non-static current embeddings~$(z^{sa}_{t+1}, z^{s}_{t+1})$. 

\begin{tcolorbox}[notitle, boxrule=0pt, colback=gray!10, colframe=gray!10, left=2pt, right=2pt, top=2pt, bottom=2pt]
	$\Rightarrow$ The added features have a greater impact on the value function than the policy, but are beneficial for both networks. All components of the value function input~$(z^{sa}, z^s, s, a)$, are necessary to achieve the highest performance. 
	While the state-action embedding~$z^{sa}$ is a useful representation for value learning, it is only trained to predict the next state and may overlook other relevant aspects of the original state-action input~$(s, a)$. Solely using the state-action embedding~$z^{sa}$ as input leads to poor results, but combining it with the original input~$(s, a)$ significantly improves performance. 
	
\end{tcolorbox}

\textbf{Normalization.} TD7 uses AvgL1Norm (\autoref{eqn:AvgL1Norm}) to normalize the scale of the state embedding~$z^s$, as well as on the state-action input $(s,a)$, following a linear layer (\autoref{eqn:norm_input}). We attempt removing AvgL1Norm on $(s,a)$, removing it entirely, and adding it to the state-action embedding $z^{sa}$. We additionally test swapping AvgL1Norm for BatchNorm~\citep{ioffe2015batch} and LayerNorm~\citep{ba2016layer}. Finally, instead of directly applying normalization to the embeddings, we replace the MSE in the encoder loss~(\autoref{eqn:encoder_loss}) by the cosine loss from~\citet{schwarzer2020data}.

\begin{tcolorbox}[notitle,boxrule=0pt,colback=gray!10,colframe=gray!10, left=2pt, right=2pt, top=2pt, bottom=2pt]
	$\Rightarrow$ The usage of AvgL1Norm is beneficial and related alternate approaches do not achieve the same performance. 
\end{tcolorbox}

\textbf{End-to-end.} Embeddings can be trained independently or end-to-end with the downstream task. 
We test our approach as an auxiliary loss to the value function. The encoders and the value function are trained end-to-end, thus allowing the value loss to affect the embeddings $(z^{sa}, z^s)$, where the encoder loss (\autoref{eqn:encoder_loss}) is multiplied by a constant to weigh its importance versus the value loss. 

\begin{tcolorbox}[notitle,boxrule=0pt,colback=gray!10,colframe=gray!10, left=2pt, right=2pt, top=2pt, bottom=2pt]
	$\Rightarrow$ Learning the embeddings end-to-end with the value function performs signficantly worse than decoupled representation learning.
\end{tcolorbox}

\section{Stabilizing RL with Decoupled Representation Learning} \label{sec:stabilizing}

In this section, we present the TD7 algorithm (TD3\texttt{+}4 additions). We begin by introducing the use of checkpoints in RL to improve the stability of RL agents. We then combine SALE with checkpoints and various previous algorithmic modifications to TD3~\citep{fujimoto2018addressing} to create a single RL algorithm for both the online and offline setting.

\subsection{Policy Checkpoints} \label{subsec:checkpoints}

Deep RL algorithms are notoriously unstable~\citep{henderson2017deep}. The unreliable nature of deep RL algorithms suggest a need for stabilizing techniques. While we can often directly address the source of instability, some amount of instability is inherent to the combination of function approximation and RL. In this section, we propose the use of checkpoints, to preserve evaluation performance, irrespective of the quality of the current learned policy. 

A checkpoint is a snapshot of the parameters of a model, captured at a specific time during training. In supervised learning, checkpoints are often used to recall a previous set of high-performing parameters based on validation error, and maintain a consistent performance across evaluations~\citep{vaswani2017attention, kenton2019bert}. Yet this technique is surprisingly absent from the deep RL toolkit for stabilizing policy performance. 

In RL, using the checkpoint of a policy that obtained a high reward during training, instead of the current policy, could improve the stability of the performance at test time.

For off-policy deep RL algorithms, the standard training paradigm is to train after each time step (typically at a one-to-one ratio: one gradient step for one data point). However, this means that the policy changes throughout each episode, making it hard to evaluate the performance. Similar to many on-policy algorithms~\citep{williams1992reinforce, PPO}, we propose to keep the policy fixed for several \textit{assessment} episodes, then batch the training that would have occurred. 
\begin{itemize}[nosep]
	\item Standard off-policy RL: Collect a data point $\rightarrow$ train once.
	\item Proposed: Collect $N$ data points over several assessment episodes $\rightarrow$ train $N$ times.
\end{itemize}
In a similar manner to evolutionary approaches~\citep{salimans2017evolution}, we can use these assessment episodes to judge if the current policy outperforms the previous best policy and checkpoint accordingly. At evaluation time, the checkpoint policy is used, rather than the current policy. 

We make two additional modifications to this basic strategy.

\textbf{Minimum over mean.} Setting aside practical considerations, the optimal approach would be to evaluate the average performance of each policy using as many trials as possible. However, to preserve learning speed and sample efficiency, it is only sensible to use a handful of trials. As such, to penalize unstable policies using a finite number of assessment episodes, we use the minimum performance, rather than the mean performance. This approach also means that extra assessment episodes do not need to be wasted on poorly performing policies, since training can resume early if the performance of any episode falls below the checkpoint performance.

\textbf{Variable assessment length.} In \autoref{appendix:sec:checkpoints}, we examine the caliber of policies trained with a varied number of assessment episodes and observe that a surprisingly high number of episodes (20\texttt{+}) can be used without compromising the performance of the final policy. However, the use of many assessment episodes negatively impacts the early performance of the agent. Freezing training for many episodes means that the environment is explored by a stale policy, reducing data diversity, and delaying feedback from policy updates.  
To counteract this effect, we restrict the number of assessment episodes used during the initial phase of training before increasing it. 

Additional details of our approach to policy checkpoints can be found in \autoref{appendix:sec:checkpoints}.

\subsection{TD7} \label{subsec:TD7}

TD7 is based on TD3~\citep{fujimoto2018addressing} with LAP~\citep{fujimoto2020equivalence}, a behavior cloning term for offline RL~\citep{fujimoto2021minimalist}, SALE~(\autoref{subsec:SALE}), and policy checkpoints~(\autoref{subsec:checkpoints}).

\textbf{LAP.} Gathered experience is stored in a replay buffer~\citep{expreplay1992} and sampled according to LAP~\citep{fujimoto2020equivalence}, a prioritized replay buffer $D$~\citep{PrioritizedExpReplay} where a transition tuple~$i:=(s,a,r,s')$ is sampled with probability 
\begin{align} \label{eqn:LAP}
	p(i) = \frac{\max \lp |\delta(i)|^\al, 1\rp}{\sum_{j \in D} \max \lp |\delta(j)|^\al, 1\rp}, \qquad \text{ where } \delta(i) := Q(s,a) - y,
\end{align}
where $y$ is the learning target. 
The amount of prioritization used is controlled by a hyperparameter $\al$. Furthermore, the value function loss uses the Huber loss~\citep{huber1964robust}, rather than the MSE.  

\textbf{Offline RL.} To make TD7 amenable to the offline RL setting, we add a behavior cloning loss to the policy update~\citep{DPG}, inspired by TD3+BC~\citep{fujimoto2021minimalist}:
\begin{equation} \label{eqn:TD3+BC}
	\pi \approx\argmax_\pi \E_{(s,a) \sim D} \lb Q(s, \pi(s)) - \lambda \lvert \E_{s \sim D} \lb Q(s, \pi(s)) \rb \rvert_\times \lp \pi(s) - a \rp^2 \rb.
\end{equation}
The same loss function is used for both offline and online RL, where $\lambda=0$ for the online setting. $|\cdot|_\times$ denotes the stop-gradient operation. Unlike TD3+BC, we do not normalize the state vectors. Checkpoints are not used in the offline setting, as there is no interaction with the environment. 

Both the value function and policy use the SALE embeddings as input, which we omit from the equations above for simplicity. Pseudocode for TD7 is described in \autoref{alg:TD7}.

\tikzexternaldisable
\begin{algorithm}[t]
\small
   \caption{Online TD7} \label{alg:TD7}
\begin{algorithmic}[1]
    \BeginBox[fill=gray!20]
    \State {\bfseries Initialize:} 
    \Comment{Before training}
    \Statex \quad $\cdot$ Policy $\pi_{t+1}$, value function $Q_{t+1}$, encoders $(f_{t+1}, g_{t+1})$.
    \Statex \quad $\cdot$ Target policy $\pi_t$, target value function $Q_t$, fixed encoders~$(f_t,g_t)$, target fixed encoders~$(f_{t-1}, g_{t-1})$.
    \Statex \quad $\cdot$ Checkpoint policy $\pi_c$, checkpoint encoder $f_c$.
    \EndBox
    \BeginBox[fill=gray!10]
    \For{$\texttt{episode}=1$ {\bfseries to} $\texttt{final\_episode}$}
    \Comment{Data collection}
    \State Using current policy $\pi_{t+1}$, collect transitions and store in the LAP replay buffer.
    \EndBox
    \BeginBox[fill=gray!20]
    \If{$\texttt{checkpoint\_condition}$} 
    \Comment{Checkpointing}
    \If{actor $\pi_{t+1}$ \texttt{outperforms} checkpoint policy $\pi_c$} 
    \State Update checkpoint networks $\pi_c \leftarrow \pi_{t+1}$, $f_c \leftarrow f_{t}$. 
    \EndIf
    \EndBox
    \BeginBox[fill=gray!10]
    \For{$i=1$ {\bfseries to} $\texttt{timesteps\_since\_training}$}
    \Comment{Training}
    \State Sample transitions from LAP replay buffer (\autoref{eqn:LAP}).
    \State Train encoder (\autoref{eqn:encoder_loss}), value function (Equations \ref{eqn:critic_loss} \& \ref{eqn:clipped}), and policy (\autoref{eqn:TD3+BC}).
    \If{$\texttt{target\_update\_frequency}$ steps have passed}
    \State Update target networks (\autoref{eqn:target_networks}). 
    \EndIf
    \EndFor
    \EndIf
    \EndFor
    \EndBox
    \Statex \BoxedString[fill=gray!10]{\textcolor{gray}{$\triangleright$ Detailed hyperparameter explanations found in the Appendix.}}
\end{algorithmic}
\end{algorithm}
\tikzexternalenable

\section{Results}

In this section, we evaluate the performance of TD7 in both the online and offline regimes. A detailed description of the experimental setup, baselines, and hyperparameters can be found in the \hyperref[appendix]{Appendix}, along with additional learning curves and ablation studies. 

\begin{figure}[ht]
	\begin{tikzpicture}[trim axis right]
		\begin{axis}[
			width=0.275\textwidth,
			title={\phantom{p}HalfCheetah\phantom{p}},
			ylabel={Total Reward (1k)},
			xtick={0, 1, 2, 3, 4, 5},
			xticklabels={0, 1, 2, 3, 4, 5},
			ytick={0, 4, 8, 12, 16, 20},
			yticklabels={0, 4, 8, 12, 16, 20},
			]
			\addplot graphics [
			ymin=-0.6, ymax=20.6,
			xmin=-0.15, xmax=5.15,
			]{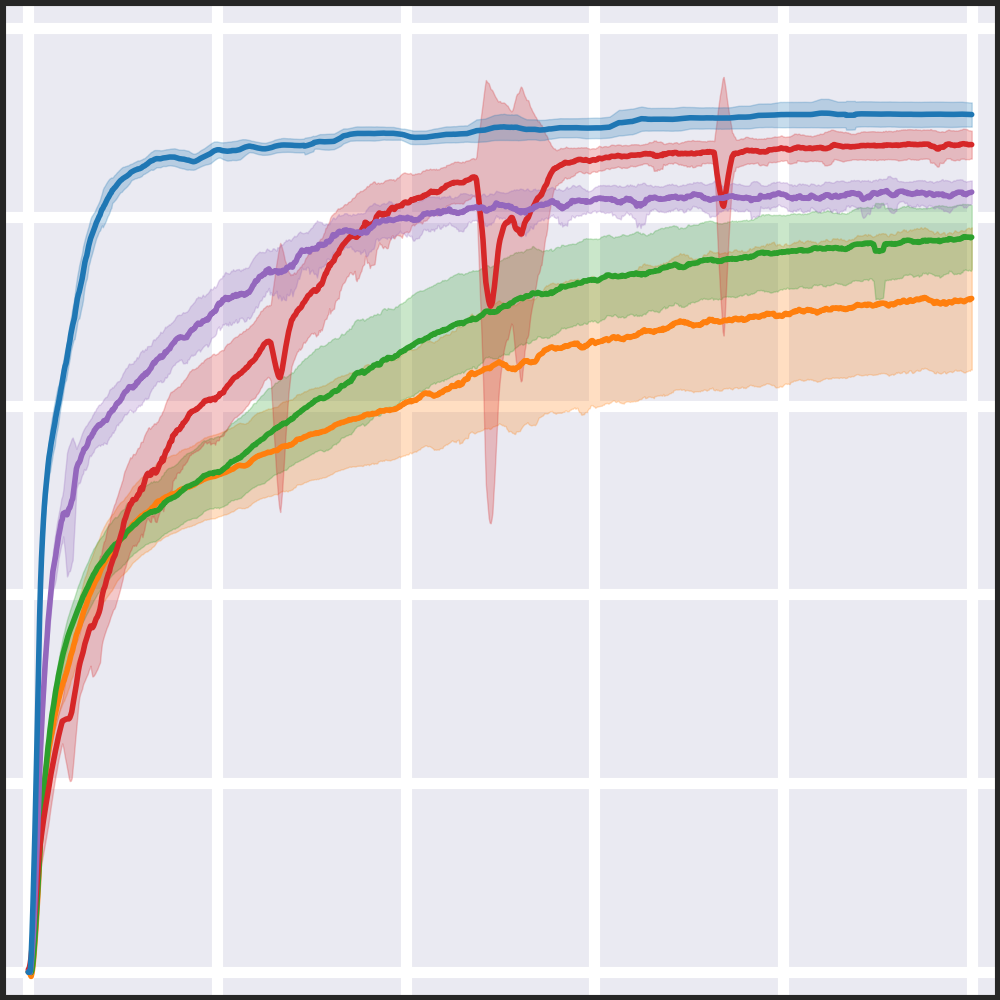};
		\end{axis}
	\end{tikzpicture}
	\begin{tikzpicture}[trim axis right]
		\begin{axis}[
			width=0.275\textwidth,
			title={Hopper},
			xtick={0, 1, 2, 3, 4, 5},
			xticklabels={0, 1, 2, 3, 4, 5},
			ytick={0, 1, 2, 3, 4},
			yticklabels={0, \phantom{0}1, 2, 3, 4},
			]
			\addplot graphics [
			ymin=-0.135, ymax=4.635,
			xmin=-0.15, xmax=5.15,
			]{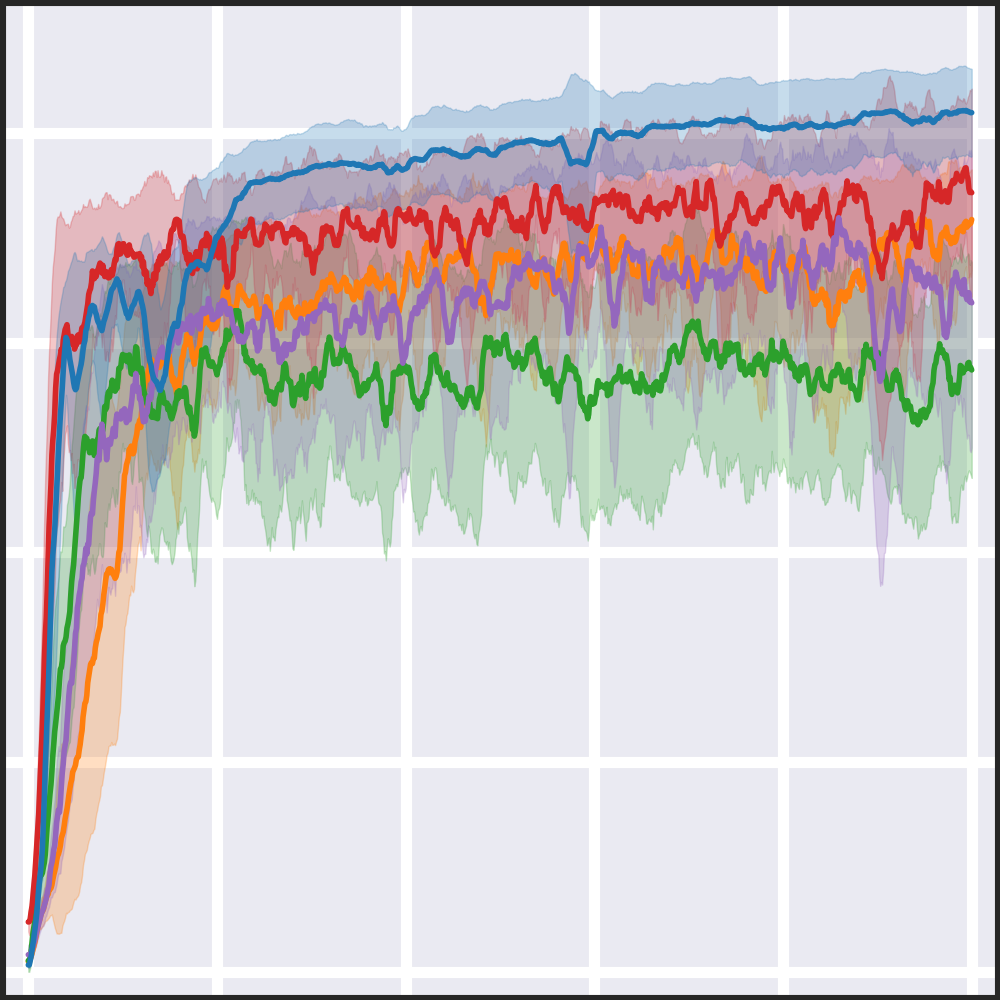};
		\end{axis}
	\end{tikzpicture}
	\begin{tikzpicture}[trim axis right]
		\begin{axis}[
			width=0.275\textwidth,
			title={\phantom{p}Walker2d\phantom{p}},
			xtick={0, 1, 2, 3, 4, 5},
			xticklabels={0, 1, 2, 3, 4, 5},
			ytick={0, 2, 4, 6, 8},
			yticklabels={0, \phantom{0}2, 4, 6, 8},
			]
			\addplot graphics [
			ymin=-0.24, ymax=8.24,
			xmin=-0.15, xmax=5.15,
			]{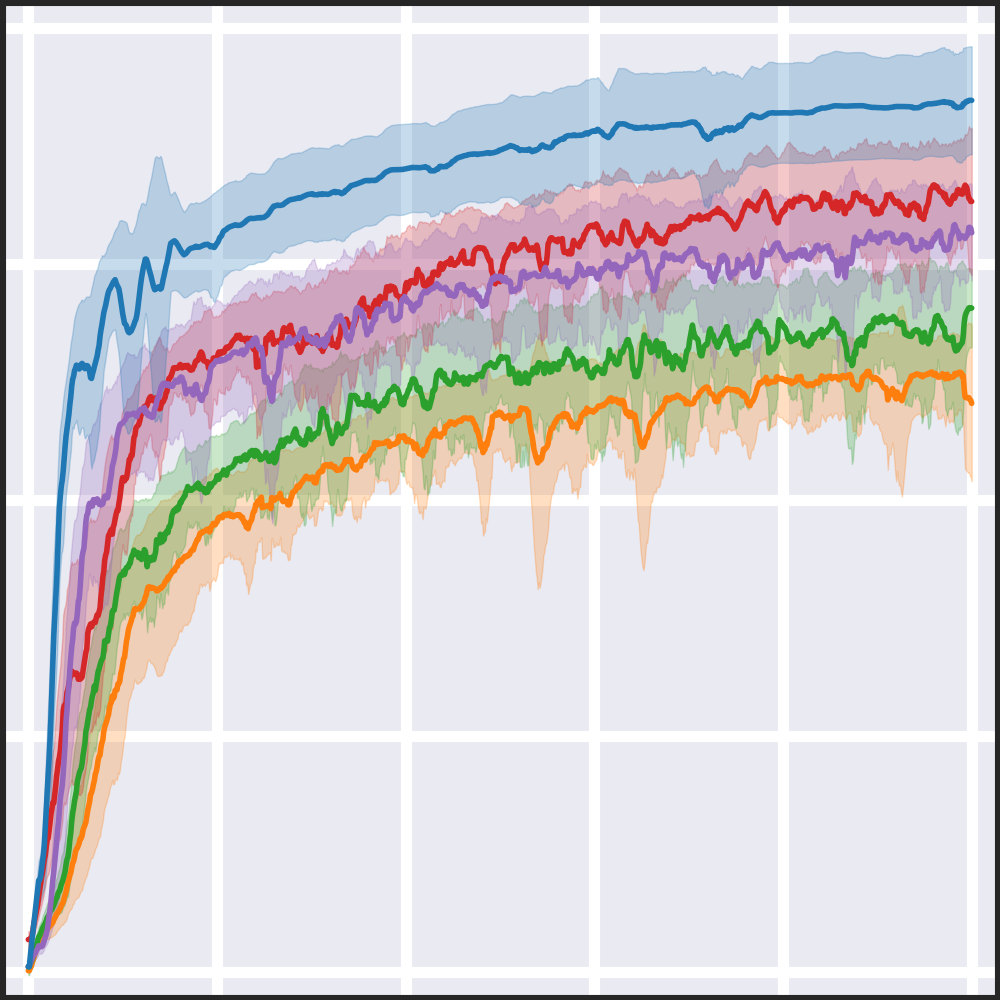};
		\end{axis}
	\end{tikzpicture}
	
	\begin{minipage}{0.7\textwidth}
		
		\begin{tikzpicture}[trim axis right]
			\begin{axis}[
				width=0.393\textwidth,
				ylabel={Total Reward (1k)},
				title={\phantom{p}Ant\phantom{p}},
				xlabel={Time steps (1M)},
				xtick={0, 1, 2, 3, 4, 5},
				xticklabels={0, 1, 2, 3, 4, 5},
				ytick={0, 3, 6, 9, 12},
				yticklabels={0, 3, 6, 9, 12},
				]
				\addplot graphics [
				ymin=-0.36, ymax=12.36,
				xmin=-0.15, xmax=5.15,
				]{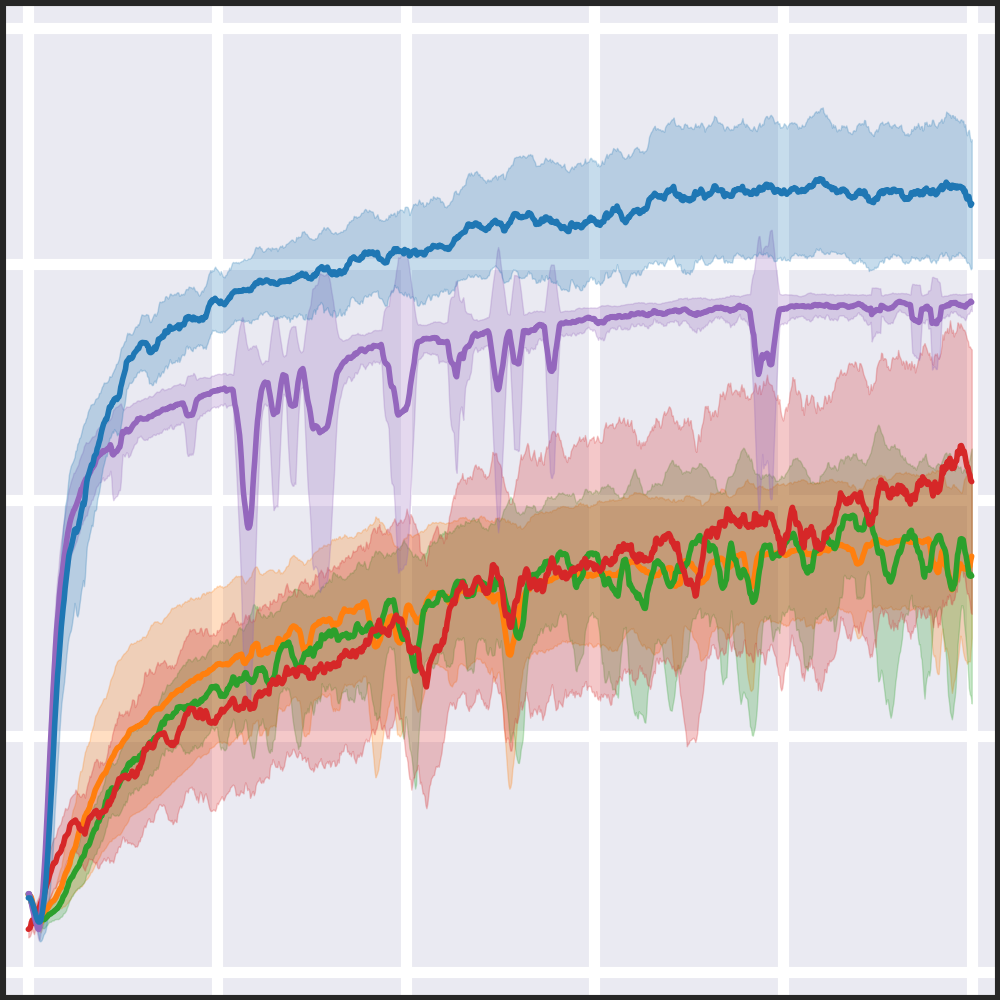};
			\end{axis}
		\end{tikzpicture}
		\begin{tikzpicture}[trim axis right]
			\begin{axis}[
				width=0.393\textwidth,
				title={\phantom{p}Humanoid\phantom{p}},
				xlabel={Time steps (1M)},
				xtick={0, 1, 2, 3, 4, 5},
				xticklabels={0, 1, 2, 3, 4, 5},
				ytick={0, 2, 4, 6, 8, 10},
				yticklabels={0, 2, 4, 6, 8, 10},
				]
				\addplot graphics [
				ymin=-0.33, ymax=11.33,
				xmin=-0.15, xmax=5.15,
				]{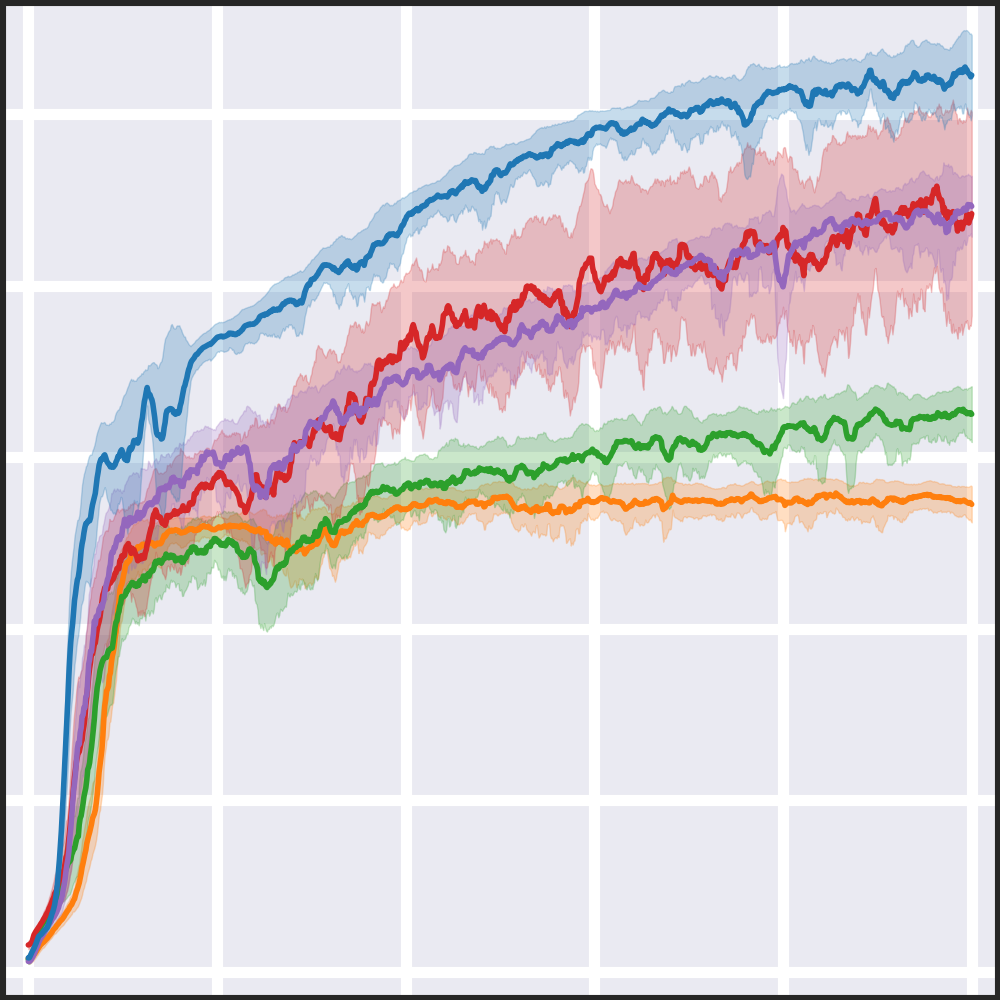};
			\end{axis}
		\end{tikzpicture}
	\end{minipage}
	\hspace{-0.035\textwidth}
	\begin{minipage}{0.3\textwidth}
		\centering
		\small
		\fcolorbox{gray}{gray!10}{
			\begin{tabular}{ll}
				\cblock{sb_blue}~TD7 & \cblock{sb_orange}~TD3 \\
				\cblock{sb_green}~SAC & \cblock{sb_red}~TQC \\
				\cblock{sb_purple}~TD3+OFE 
			\end{tabular}	
		} 
		\captionof{figure}{Learning curves on the MuJoCo benchmark. Results are averaged over 10 seeds. The shaded area captures a 95\% confidence interval around the average performance.} \label{fig:online_curves}    
	\end{minipage}
\end{figure}

\textbf{Online.} Using OpenAI gym~\citep{OpenAIGym}, we benchmark TD7 against TD3~\citep{fujimoto2018addressing}, SAC~\citep{haarnoja2018soft}, TQC~\citep{kuznetsov2020controlling}, and TD3+OFE~\citep{ota2020can} on the MuJoCo environments~\citep{mujoco}. SAC and TD3+OFE results are from re-implementations based on author descriptions~\citep{haarnoja2018soft, ota2020can}. TD3 and TQC results use author-provided code~\citep{fujimoto2018addressing, kuznetsov2020controlling}, with a consistent evaluation protocol for all methods. 
Learning curves are displayed in \autoref{fig:online_curves} 
and final and intermediate results are listed in \autoref{table:online_results}. 

Although TQC and TD3+OFE use per-environment hyperparameters along with larger and more computationally expensive architectures, TD7 outperforms these baselines significantly in terms of both early (300k time steps) and final performance (5M time steps). At 300k time steps, TD7 often surpasses the performance of TD3 at 5M time steps, highlighting the considerable performance gains. %

\begin{table*}[ht]
\centering
\small
\setlength{\tabcolsep}{4pt}
\newcolumntype{Y}{>{\centering\arraybackslash}X} %
\caption{Average performance on the MuJoCo benchmark at 300k, 1M, and 5M time steps, over 10 trials, where $\pm$~captures a 95\% confidence interval. The \mhl{highest performance} is highlighted. Any performance which is \mlhl{not statistically significantly} worse than the highest performance (according to a Welch's $t$-test with significance level $0.05$) is highlighted.}
 \label{table:online_results}
\begin{tabularx}{\textwidth}{lcYYYYY}
\toprule
Environment                  & Time step & TD3             & SAC                  & TQC              & TD3+OFE     & TD7 \\ 
\midrule
\multirow{3}{*}{HalfCheetah} & 300k & \po7715 \textcolor{gray}{$\pm$ 633\po} & \po8052 \textcolor{gray}{$\pm$ 515\po} & \po7006 \textcolor{gray}{$\pm$ 891\po} & 11294 \textcolor{gray}{$\pm$ 247\po} & \mhl{15031}\textcolor{gray}{\mhl{ $\pm$ 401\po}} \\
                             & 1M   & 10574 \textcolor{gray}{$\pm$ 897\po} & 10484 \textcolor{gray}{$\pm$ 659\po} & 12349 \textcolor{gray}{$\pm$ 878\po} & 13758 \textcolor{gray}{$\pm$ 544\po} & \mhl{17434}\textcolor{gray}{\mhl{ $\pm$ 155\po}} \\
                             & 5M   & 14337 \textcolor{gray}{$\pm$ 1491} & 15526 \textcolor{gray}{$\pm$ 697\po} & 17459 \textcolor{gray}{$\pm$ 258\po} & 16596 \textcolor{gray}{$\pm$ 164\po} & \mhl{18165}\textcolor{gray}{\mhl{ $\pm$ 255\po}} \\
                             \midrule
\multirow{3}{*}{Hopper}      & 300k & \po1289 \textcolor{gray}{$\pm$ 768\po} & \po2370 \textcolor{gray}{$\pm$ 626\po} & \mhl{\po3251}\textcolor{gray}{\mhl{ $\pm$ 461\po}} & \po1581 \textcolor{gray}{$\pm$ 682\po} & \mlhl{\po2948}\textcolor{gray}{\mlhl{ $\pm$ 464\po}} \\
                             & 1M   & \po3226 \textcolor{gray}{$\pm$ 315\po} & \po2785 \textcolor{gray}{$\pm$ 634\po} & \mhl{\po3526}\textcolor{gray}{\mhl{ $\pm$ 244\po}} & \po3121 \textcolor{gray}{$\pm$ 506\po} & \mlhl{\po3512}\textcolor{gray}{\mlhl{ $\pm$ 315\po}} \\
                             & 5M   & \po3682 \textcolor{gray}{$\pm$ 83\po\po} & \po3167 \textcolor{gray}{$\pm$ 485\po} & \mlhl{\po3462}\textcolor{gray}{\mlhl{ $\pm$ 818\po}} & \po3423 \textcolor{gray}{$\pm$ 584\po} & \mhl{\po4075}\textcolor{gray}{\mhl{ $\pm$ 225\po}} \\
                             \midrule
\multirow{3}{*}{Walker2d}    & 300k & \po1101 \textcolor{gray}{$\pm$ 386\po} & \po1989 \textcolor{gray}{$\pm$ 500\po} & \po2812 \textcolor{gray}{$\pm$ 838\po} & \po4018 \textcolor{gray}{$\pm$ 570\po} & \mhl{\po5379}\textcolor{gray}{\mhl{ $\pm$ 328\po}} \\
                             & 1M   & \po3946 \textcolor{gray}{$\pm$ 292\po} & \po4314 \textcolor{gray}{$\pm$ 256\po} & \po5321 \textcolor{gray}{$\pm$ 322\po} & \po5195 \textcolor{gray}{$\pm$ 512\po} & \mhl{\po6097}\textcolor{gray}{\mhl{ $\pm$ 570\po}} \\
                             & 5M   & \po5078 \textcolor{gray}{$\pm$ 343\po} & \po5681 \textcolor{gray}{$\pm$ 329\po} & \po6137 \textcolor{gray}{$\pm$ 1194} & \po6379 \textcolor{gray}{$\pm$ 332\po} & \mhl{\po7397}\textcolor{gray}{\mhl{ $\pm$ 454\po}} \\
                             \midrule
\multirow{3}{*}{Ant}         & 300k & \po1704 \textcolor{gray}{$\pm$ 655\po} & \po1478 \textcolor{gray}{$\pm$ 354\po} & \po1830 \textcolor{gray}{$\pm$ 572\po} & \mhl{\po6348}\textcolor{gray}{\mhl{ $\pm$ 441\po}} & \mlhl{\po6171}\textcolor{gray}{\mlhl{ $\pm$ 831\po}} \\
                             & 1M   & \po3942 \textcolor{gray}{$\pm$ 1030} & \po3681 \textcolor{gray}{$\pm$ 506\po} & \po3582 \textcolor{gray}{$\pm$ 1093} & \po7398 \textcolor{gray}{$\pm$ 118\po} & 
                             \mhl{\po8509}\textcolor{gray}{\mhl{ $\pm$ 422\po}} \\
                             & 5M   & \po5589 \textcolor{gray}{$\pm$ 758\po} & \po4615 \textcolor{gray}{$\pm$ 2022} & \po6329 \textcolor{gray}{$\pm$ 1510} & \po8547 \textcolor{gray}{$\pm$ 84\po\po} & \mhl{10133}\textcolor{gray}{\mhl{ $\pm$ 966\po}} \\
                             \midrule
\multirow{3}{*}{Humanoid}    & 300k & \po1344 \textcolor{gray}{$\pm$ 365\po} & \po1997 \textcolor{gray}{$\pm$ 483\po} & \po3117 \textcolor{gray}{$\pm$ 910\po} & \po3181 \textcolor{gray}{$\pm$ 771\po} & \mhl{\po5332}\textcolor{gray}{\mhl{ $\pm$ 714\po}} \\
                             & 1M   & \po5165 \textcolor{gray}{$\pm$ 145\po} & \po4909 \textcolor{gray}{$\pm$ 364\po} & \po6029 \textcolor{gray}{$\pm$ 531\po} & \po6032 \textcolor{gray}{$\pm$ 334\po} & \mhl{\po7429}\textcolor{gray}{\mhl{ $\pm$ 153\po}} \\
                             & 5M   & \po5433 \textcolor{gray}{$\pm$ 245\po} & \po6555 \textcolor{gray}{$\pm$ 279\po} & \po8361 \textcolor{gray}{$\pm$ 1364} & \po8951 \textcolor{gray}{$\pm$ 246\po} & \mhl{10281}\textcolor{gray}{\mhl{ $\pm$ 588\po}} \\
                             \bottomrule
\end{tabularx}
\vspace{-8pt}
\end{table*}

\begin{table*}[t]
\centering
\small
\setlength{\tabcolsep}{2pt}
\newcolumntype{Y}{>{\centering\arraybackslash}X} %
\caption[]{Average final performance on the D4RL benchmark after training for 1M time steps.  over 10 trials, where $\pm$~captures a 95\% confidence interval. The \mhl{highest performance} is highlighted. Any performance which is \mlhl{not statistically significantly} worse than the highest performance (according to a Welch's $t$-test with significance level $0.05$) is highlighted.}
\label{table:offline_results}
\begin{tabularx}{\textwidth}{llYYYYY}
\toprule
\multicolumn{1}{l}{Environment} & Dataset       & CQL   & TD3+BC & IQL & $\mathcal{X}$-QL & TD7   \\ 
\midrule
\multirow{3}{*}{HalfCheetah} & Medium & \po46.7 \textcolor{gray}{$\pm$ 0.3\po} & \po48.1 \textcolor{gray}{$\pm$ 0.1\po} & \po47.4 \textcolor{gray}{$\pm$ 0.2\po} & \po47.4 \textcolor{gray}{$\pm$ 0.1\po} & \mhl{\po58.0}\textcolor{gray}{\mhl{ $\pm$ 0.4\po}}\\
 & Medium-Replay & \po45.5 \textcolor{gray}{$\pm$ 0.3\po} & \po44.6 \textcolor{gray}{$\pm$ 0.4\po} & \po43.9 \textcolor{gray}{$\pm$ 1.3\po} & \po44.2 \textcolor{gray}{$\pm$ 0.7\po} & \mhl{\po53.8}\textcolor{gray}{\mhl{ $\pm$ 0.8\po}}\\
 & Medium-Expert & \po76.8 \textcolor{gray}{$\pm$ 7.4\po} & \po93.7 \textcolor{gray}{$\pm$ 0.9\po} & \po89.6 \textcolor{gray}{$\pm$ 3.5\po} & \po90.2 \textcolor{gray}{$\pm$ 2.7\po} & \mhl{104.6}\textcolor{gray}{\mhl{ $\pm$ 1.6\po}}\\
\midrule
\multirow{3}{*}{Hopper} & Medium & \po59.3 \textcolor{gray}{$\pm$ 3.3\po} & \po59.1 \textcolor{gray}{$\pm$ 3.0\po} & \po63.9 \textcolor{gray}{$\pm$ 4.9\po} & \po67.7 \textcolor{gray}{$\pm$ 3.6\po} & \mhl{\po76.1}\textcolor{gray}{\mhl{ $\pm$ 5.1\po}}\\
 & Medium-Replay & \po78.8 \textcolor{gray}{$\pm$ 10.9} & \po52.0 \textcolor{gray}{$\pm$ 10.6} & \mhl{\po93.4}\textcolor{gray}{\mhl{ $\pm$ 7.8\po}} & \mlhl{\po82.0}\textcolor{gray}{\mlhl{ $\pm$ 14.9}} & \mlhl{\po91.1}\textcolor{gray}{\mlhl{ $\pm$ 8.0\po}}\\
 & Medium-Expert & \po79.9 \textcolor{gray}{$\pm$ 19.8} & \mlhl{\po98.1}\textcolor{gray}{\mlhl{ $\pm$ 10.7}} & \po64.2 \textcolor{gray}{$\pm$ 32.0} & \po92.0 \textcolor{gray}{$\pm$ 10.0} & \mhl{108.2}\textcolor{gray}{\mhl{ $\pm$ 4.8\po}}\\
\midrule
\multirow{3}{*}{Walker2d} & Medium & \po81.4 \textcolor{gray}{$\pm$ 1.7\po} & \mlhl{\po84.3}\textcolor{gray}{\mlhl{ $\pm$ 0.8\po}} & \mlhl{\po84.2}\textcolor{gray}{\mlhl{ $\pm$ 1.6\po}} & \po79.2 \textcolor{gray}{$\pm$ 4.0\po} & \mhl{\po91.1}\textcolor{gray}{\mhl{ $\pm$ 7.8\po}}\\
 & Medium-Replay & \po79.9 \textcolor{gray}{$\pm$ 3.6\po} & \po81.0 \textcolor{gray}{$\pm$ 3.4\po} & \po71.2 \textcolor{gray}{$\pm$ 8.3\po} & \po61.8 \textcolor{gray}{$\pm$ 7.7\po} & \mhl{\po89.7}\textcolor{gray}{\mhl{ $\pm$ 4.7\po}}\\
 & Medium-Expert & 108.5 \textcolor{gray}{$\pm$ 1.2\po} & 110.5 \textcolor{gray}{$\pm$ 0.4\po} & 108.9 \textcolor{gray}{$\pm$ 1.4\po} & 110.3 \textcolor{gray}{$\pm$ 0.2\po} & \mhl{111.8}\textcolor{gray}{\mhl{ $\pm$ 0.6\po}}\\
\midrule
\multicolumn{2}{l}{Total} & 656.7 \textcolor{gray}{$\pm$ 24.3} & 671.3 \textcolor{gray}{$\pm$ 15.7} & 666.7 \textcolor{gray}{$\pm$ 34.6} & 674.9 \textcolor{gray}{$\pm$ 20.4} & \mhl{784.4}\textcolor{gray}{\mhl{ $\pm$ 14.1}}\\
\bottomrule 
\end{tabularx}
\end{table*}

\textbf{Offline.} We benchmark TD7 against CQL~\citep{kumar2020conservative}, TD3+BC~\citep{fujimoto2021minimalist}, IQL~\citep{kostrikov2021offline} and $\mathcal{X}$-QL~\citep{garg2023extreme} using the MuJoCo datasets in D4RL~\citep{mujoco, fu2021benchmarks}. While there are methods that use per-dataset hyperparameters to attain higher total results, we omit these methods because it makes it difficult to directly compare. Baseline results are obtained by re-running author-provided code with a single set of hyperparameters and a consistent evaluation protocol. Final performance is reported in \autoref{table:offline_results}. 

TD7 outperforms all baselines. Since TD7 and TD3+BC employ the same approach to offline RL, the significant performance gap highlights the effectiveness of SALE in the offline setting. 

\begin{wrapfigure}{r}{0.45\textwidth} %
\vspace{-36pt}
\centering
\hspace{-6pt}
\begin{tikzpicture}[trim axis right]
\begin{axis}[
    width=0.28\textwidth,
    title={Ablation},
    title style={yshift=2pt},
    ylabel={Normalized Performance},
    xlabel={Time steps (1M)},
    xtick={0, 1, 2, 3, 4, 5},
    xticklabels={0, 1, 2, 3, 4, 5},
    ytick={0, 0.2, 0.4, 0.6, 0.8, 1.0},
    yticklabels={0, 0.2, 0.4, 0.6, 0.8, 1.0},
]
\addplot graphics [
ymin=-0.03, ymax=1.08,
xmin=-0.15, xmax=5.15,
]{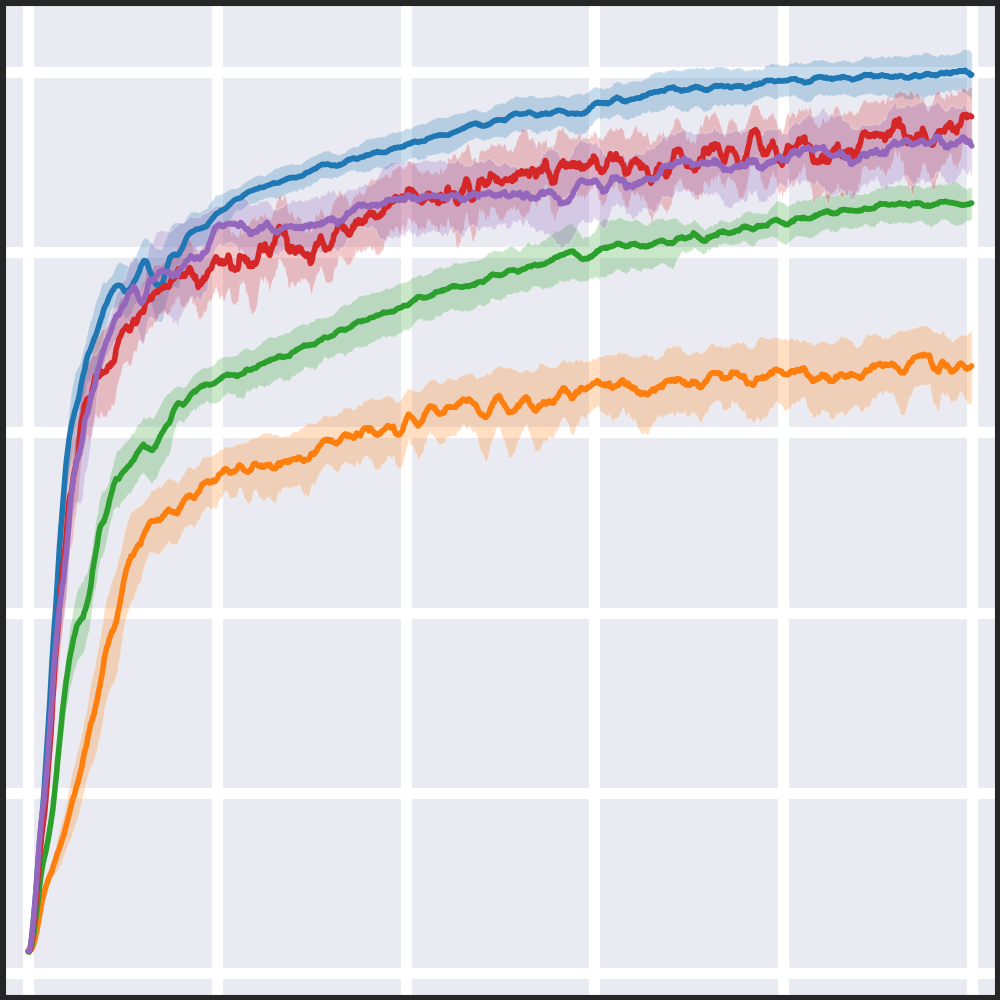};
\end{axis}
\end{tikzpicture}

\hspace{8pt}
\fcolorbox{gray}{gray!10}{
\small
\begin{tabular}{@{\hspace{2pt}}lll@{}}
\cblock{sb_blue}~TD7 & \cblock{sb_orange}~TD3 & \cblock{sb_green}~No SALE \\ 
\multicolumn{2}{@{\hspace{2pt}}l}{\cblock{sb_red}~No checkpoints} & \cblock{sb_purple}~No LAP %
\end{tabular}
} 
\captionof{figure}{Ablation study over the components of TD7. The y-axis corresponds to the average performance over all five MuJoCo tasks, normalized with respect to the performance of TD7 at 5M time steps. The shaded area captures a 95\% confidence interval.}
\label{fig:ablation}
\vspace{8pt}
\hspace{-10pt}
\begin{tikzpicture}[trim axis right]
    \begin{axis}[
        width=0.36\textwidth,
        height=0.2\textwidth, 	
        title={Run Time},
        title style={yshift=2pt},
        ylabel={Hours},
        xtick={1, 2, 3, 4, 5},
        xticklabels={TD3, SAC, TQC, TD3+OFE, TD7},
        ytick={60, 120, 180, 240},
        yticklabels={1, 2, 3, 4},
        clip=false
        ]
        \addplot graphics [
        ymin=0, ymax=260,
        xmin=0.4, xmax=5.6,
        ]{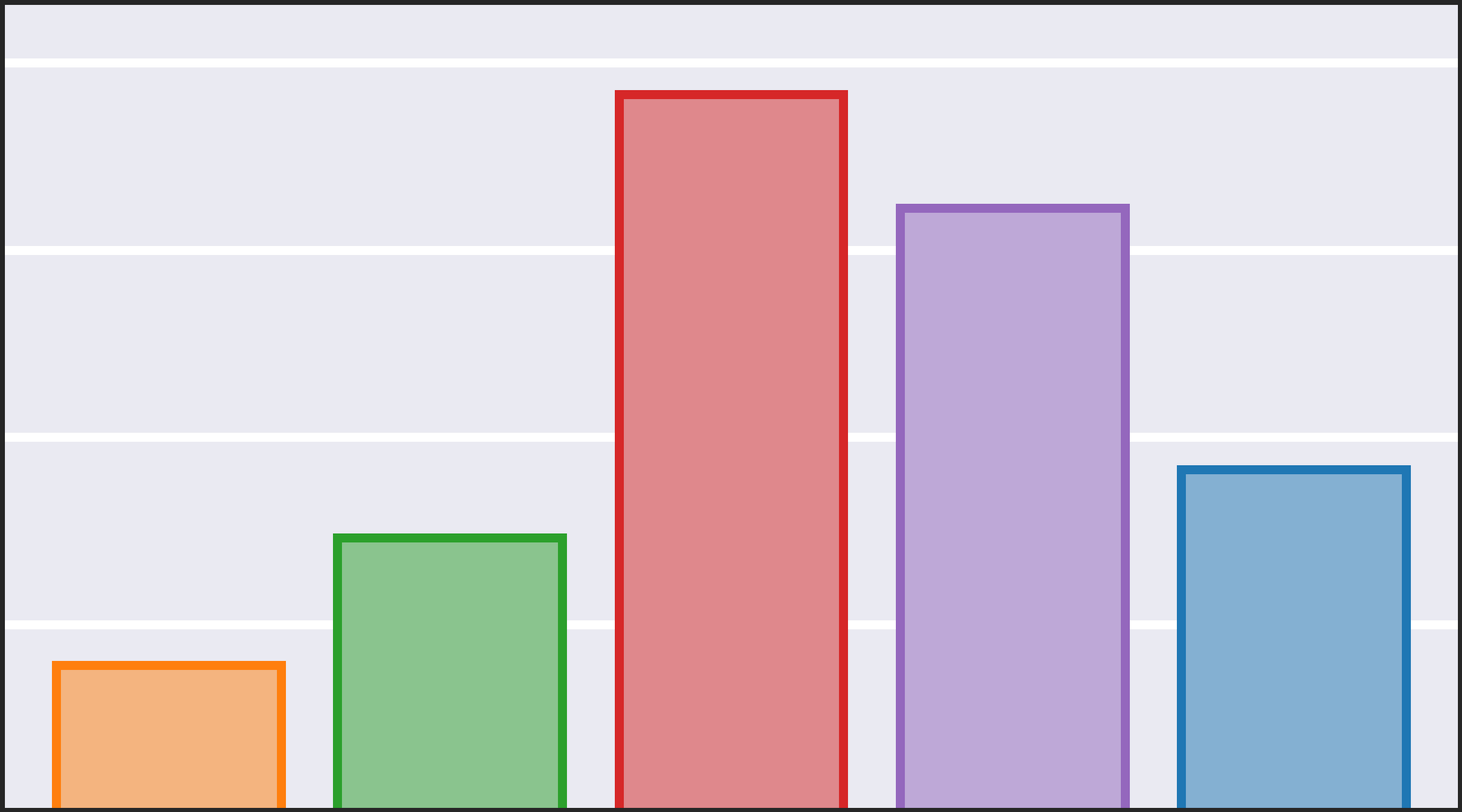};
        \node[anchor=mid] at (1, 57.06) {\scriptsize 47m};
        \node[anchor=mid] at (2, 97.88) {\scriptsize 1h 28m};
        \node[anchor=mid] at (3, 239.73) {\scriptsize 3h 50m};
        \node[anchor=mid] at (4, 203.57) {\scriptsize 3h 14m};
        \node[anchor=mid] at (5, 119.85) {\scriptsize 1h 50m};
        \draw[line width=0.5pt] (1,0) -- (1,-5.0);
        \draw[line width=0.5pt] (2,0) -- (2,-5.0);
        \draw[line width=0.5pt] (3,0) -- (3,-5.0);
        \draw[line width=0.5pt] (4,0) -- (4,-5.0);
        \draw[line width=0.5pt] (5,0) -- (5,-5.0);
    \end{axis}
\end{tikzpicture}    
\captionof{figure}{Run time of each method for 1M time steps on the HalfCheetah environment, using the same hardware and deep learning framework (PyTorch~\citep{paszke2019pytorch}). 
}\label{fig:runtime}
\vspace{-10pt}
\end{wrapfigure}

\textbf{Ablation study}. In \autoref{fig:ablation} we report the results of an ablation study over the components of TD7 (SALE, checkpoints, LAP). The interaction between components is explored further in \autoref{appendix:sec:ablation}. %

\textbf{Run time.} To understand the computational cost of using SALE and the TD7 algorithm, we benchmark the run time of each of the online baselines with identical computational resources and deep learning framework.  
The results are reported in \autoref{fig:runtime}.

\FloatBarrier

\section{Conclusion}

Representation learning has been typically reserved for image-based tasks, where the observations are large and unstructured. However, by learning embeddings which consider the interaction between state and action, we make representation learning more broadly applicable to low-level states. We introduce SALE, a method for learning state-action embeddings by considering a latent space dynamics model. 
Through an extensive empirical evaluation, we investigate various design choices in SALE. 

We highlight the risk of extrapolation error~\citep{fujimoto2019off} due to the increase in input dimensions from using state-action embeddings, but show this instability can be corrected by clipping the target with seen values. We further introduce stability by including policy checkpoints. 

While both SALE and policy checkpoints are general-purpose techniques that can be included with most RL methods, we combine them with TD3 and several other recent improvements~\citep{fujimoto2020equivalence, fujimoto2021minimalist} to introduce the TD7 algorithm. We find our TD7 algorithm is able to match the performance of expensive offline algorithms and significantly outperform the state-of-the-art continuous control methods 
in both final performance and early learning.

\section*{Acknowledgments and Disclosure of Funding}

This research was enabled in part by support provided by Calcul Qu\'ebec and the Digital Research Alliance of Canada.

\clearpage

\bibliography{references}

\begin{thebibliography}{80}
\providecommand{\natexlab}[1]{#1}
\providecommand{\url}[1]{\texttt{#1}}
\expandafter\ifx\csname urlstyle\endcsname\relax
  \providecommand{\doi}[1]{doi: #1}\else
  \providecommand{\doi}{doi: \begingroup \urlstyle{rm}\Url}\fi

\bibitem[Anand et~al.(2019)Anand, Racah, Ozair, Bengio, C{\^o}t{\'e}, and
  Hjelm]{anand2019unsupervised}
Ankesh Anand, Evan Racah, Sherjil Ozair, Yoshua Bengio, Marc-Alexandre
  C{\^o}t{\'e}, and R~Devon Hjelm.
\newblock Unsupervised state representation learning in atari.
\newblock \emph{Advances in neural information processing systems}, 32, 2019.

\bibitem[Ba et~al.(2016)Ba, Kiros, and Hinton]{ba2016layer}
Jimmy~Lei Ba, Jamie~Ryan Kiros, and Geoffrey~E Hinton.
\newblock Layer normalization.
\newblock \emph{arXiv preprint arXiv:1607.06450}, 2016.

\bibitem[Brockman et~al.(2016)Brockman, Cheung, Pettersson, Schneider,
  Schulman, Tang, and Zaremba]{OpenAIGym}
Greg Brockman, Vicki Cheung, Ludwig Pettersson, Jonas Schneider, John Schulman,
  Jie Tang, and Wojciech Zaremba.
\newblock Openai gym, 2016.

\bibitem[Cetin et~al.(2022)Cetin, Ball, Roberts, and
  Celiktutan]{cetin2022stabilizing}
Edoardo Cetin, Philip~J Ball, Stephen Roberts, and Oya Celiktutan.
\newblock Stabilizing off-policy deep reinforcement learning from pixels.
\newblock In \emph{International Conference on Machine Learning}, pages
  2784--2810. PMLR, 2022.

\bibitem[Chandak et~al.(2019)Chandak, Theocharous, Kostas, Jordan, and
  Thomas]{chandak2019learning}
Yash Chandak, Georgios Theocharous, James Kostas, Scott Jordan, and Philip
  Thomas.
\newblock Learning action representations for reinforcement learning.
\newblock In \emph{International conference on machine learning}, pages
  941--950. PMLR, 2019.

\bibitem[Clevert et~al.(2015)Clevert, Unterthiner, and
  Hochreiter]{clevert2015fast}
Djork-Arn{\'e} Clevert, Thomas Unterthiner, and Sepp Hochreiter.
\newblock Fast and accurate deep network learning by exponential linear units
  (elus).
\newblock \emph{arXiv preprint arXiv:1511.07289}, 2015.

\bibitem[Dayan(1993)]{dayan1993improving}
Peter Dayan.
\newblock Improving generalization for temporal difference learning: The
  successor representation.
\newblock \emph{Neural Computation}, 5\penalty0 (4):\penalty0 613--624, 1993.

\bibitem[Engstrom et~al.(2019)Engstrom, Ilyas, Santurkar, Tsipras, Janoos,
  Rudolph, and Madry]{engstrom2019implementation}
Logan Engstrom, Andrew Ilyas, Shibani Santurkar, Dimitris Tsipras, Firdaus
  Janoos, Larry Rudolph, and Aleksander Madry.
\newblock Implementation matters in deep rl: A case study on ppo and trpo.
\newblock In \emph{International Conference on Learning Representations}, 2019.

\bibitem[Ferns et~al.(2011)Ferns, Panangaden, and
  Precup]{ferns2011bisimulation}
Norm Ferns, Prakash Panangaden, and Doina Precup.
\newblock Bisimulation metrics for continuous markov decision processes.
\newblock \emph{SIAM Journal on Computing}, 40\penalty0 (6):\penalty0
  1662--1714, 2011.

\bibitem[Finn et~al.(2016)Finn, Tan, Duan, Darrell, Levine, and
  Abbeel]{finn2016deep}
Chelsea Finn, Xin~Yu Tan, Yan Duan, Trevor Darrell, Sergey Levine, and Pieter
  Abbeel.
\newblock Deep spatial autoencoders for visuomotor learning.
\newblock In \emph{2016 IEEE International Conference on Robotics and
  Automation (ICRA)}, pages 512--519. IEEE, 2016.

\bibitem[Fu et~al.(2021)Fu, Norouzi, Nachum, Tucker, Wang, Novikov, Yang,
  Zhang, Chen, Kumar, Paduraru, Levine, and Paine]{fu2021benchmarks}
Justin Fu, Mohammad Norouzi, Ofir Nachum, George Tucker, Ziyu Wang, Alexander
  Novikov, Mengjiao Yang, Michael~R Zhang, Yutian Chen, Aviral Kumar, Cosmin
  Paduraru, Sergey Levine, and Thomas Paine.
\newblock Benchmarks for deep off-policy evaluation.
\newblock In \emph{International Conference on Learning Representations}, 2021.

\bibitem[Fujimoto and Gu(2021)]{fujimoto2021minimalist}
Scott Fujimoto and Shixiang~Shane Gu.
\newblock A minimalist approach to offline reinforcement learning.
\newblock In \emph{Thirty-Fifth Conference on Neural Information Processing
  Systems}, 2021.

\bibitem[Fujimoto et~al.(2018)Fujimoto, van Hoof, and
  Meger]{fujimoto2018addressing}
Scott Fujimoto, Herke van Hoof, and David Meger.
\newblock Addressing function approximation error in actor-critic methods.
\newblock In \emph{International Conference on Machine Learning}, volume~80,
  pages 1587--1596. PMLR, 2018.

\bibitem[Fujimoto et~al.(2019)Fujimoto, Meger, and Precup]{fujimoto2019off}
Scott Fujimoto, David Meger, and Doina Precup.
\newblock Off-policy deep reinforcement learning without exploration.
\newblock In \emph{International Conference on Machine Learning}, pages
  2052--2062, 2019.

\bibitem[Fujimoto et~al.(2020)Fujimoto, Meger, and
  Precup]{fujimoto2020equivalence}
Scott Fujimoto, David Meger, and Doina Precup.
\newblock An equivalence between loss functions and non-uniform sampling in
  experience replay.
\newblock \emph{Advances in Neural Information Processing Systems}, 33, 2020.

\bibitem[Fujimoto et~al.(2021)Fujimoto, Meger, and Precup]{fujimoto2021srdice}
Scott Fujimoto, David Meger, and Doina Precup.
\newblock A deep reinforcement learning approach to marginalized importance
  sampling with the successor representation.
\newblock In \emph{Proceedings of the 38th International Conference on Machine
  Learning}, volume 139, pages 3518--3529. PMLR, 2021.

\bibitem[Fujimoto et~al.(2022)Fujimoto, Meger, Precup, Nachum, and
  Gu]{fujimoto2022should}
Scott Fujimoto, David Meger, Doina Precup, Ofir Nachum, and Shixiang~Shane Gu.
\newblock Why should i trust you, bellman? {T}he {B}ellman error is a poor
  replacement for value error.
\newblock In \emph{International Conference on Machine Learning}, volume 162,
  pages 6918--6943. PMLR, 2022.

\bibitem[Garg et~al.(2023)Garg, Hejna, Geist, and Ermon]{garg2023extreme}
Divyansh Garg, Joey Hejna, Matthieu Geist, and Stefano Ermon.
\newblock Extreme q-learning: Maxent {RL} without entropy.
\newblock In \emph{The Eleventh International Conference on Learning
  Representations}, 2023.

\bibitem[Gelada et~al.(2019)Gelada, Kumar, Buckman, Nachum, and
  Bellemare]{gelada2019deepmdp}
Carles Gelada, Saurabh Kumar, Jacob Buckman, Ofir Nachum, and Marc~G Bellemare.
\newblock Deepmdp: Learning continuous latent space models for representation
  learning.
\newblock In \emph{International Conference on Machine Learning}, pages
  2170--2179. PMLR, 2019.

\bibitem[Grill et~al.(2020)Grill, Strub, Altch{\'e}, Tallec, Richemond,
  Buchatskaya, Doersch, Avila~Pires, Guo, Gheshlaghi~Azar,
  et~al.]{grill2020bootstrap}
Jean-Bastien Grill, Florian Strub, Florent Altch{\'e}, Corentin Tallec, Pierre
  Richemond, Elena Buchatskaya, Carl Doersch, Bernardo Avila~Pires, Zhaohan
  Guo, Mohammad Gheshlaghi~Azar, et~al.
\newblock Bootstrap your own latent-a new approach to self-supervised learning.
\newblock \emph{Advances in neural information processing systems},
  33:\penalty0 21271--21284, 2020.

\bibitem[Ha and Schmidhuber(2018)]{ha2018world}
David Ha and J{\"u}rgen Schmidhuber.
\newblock World models.
\newblock \emph{arXiv preprint arXiv:1803.10122}, 2018.

\bibitem[Haarnoja et~al.(2018)Haarnoja, Zhou, Abbeel, and
  Levine]{haarnoja2018soft}
Tuomas Haarnoja, Aurick Zhou, Pieter Abbeel, and Sergey Levine.
\newblock Soft actor-critic: Off-policy maximum entropy deep reinforcement
  learning with a stochastic actor.
\newblock In \emph{International Conference on Machine Learning}, volume~80,
  pages 1861--1870. PMLR, 2018.

\bibitem[Hafner et~al.(2019)Hafner, Lillicrap, Fischer, Villegas, Ha, Lee, and
  Davidson]{hafner2019learning}
Danijar Hafner, Timothy Lillicrap, Ian Fischer, Ruben Villegas, David Ha,
  Honglak Lee, and James Davidson.
\newblock Learning latent dynamics for planning from pixels.
\newblock In \emph{International conference on machine learning}, pages
  2555--2565. PMLR, 2019.

\bibitem[Hafner et~al.(2023)Hafner, Pasukonis, Ba, and
  Lillicrap]{hafner2023mastering}
Danijar Hafner, Jurgis Pasukonis, Jimmy Ba, and Timothy Lillicrap.
\newblock Mastering diverse domains through world models.
\newblock \emph{arXiv preprint arXiv:2301.04104}, 2023.

\bibitem[Hansen et~al.(2022)Hansen, Su, and Wang]{hansen2022temporal}
Nicklas~A Hansen, Hao Su, and Xiaolong Wang.
\newblock Temporal difference learning for model predictive control.
\newblock In \emph{International Conference on Machine Learning}, pages
  8387--8406. PMLR, 2022.

\bibitem[Henderson et~al.(2017)Henderson, Islam, Bachman, Pineau, Precup, and
  Meger]{henderson2017deep}
Peter Henderson, Riashat Islam, Philip Bachman, Joelle Pineau, Doina Precup,
  and David Meger.
\newblock Deep reinforcement learning that matters.
\newblock In \emph{AAAI Conference on Artificial Intelligence}, 2017.

\bibitem[Huber et~al.(1964)]{huber1964robust}
Peter~J Huber et~al.
\newblock Robust estimation of a location parameter.
\newblock \emph{The annals of mathematical statistics}, 35\penalty0
  (1):\penalty0 73--101, 1964.

\bibitem[Ioffe and Szegedy(2015)]{ioffe2015batch}
Sergey Ioffe and Christian Szegedy.
\newblock Batch normalization: Accelerating deep network training by reducing
  internal covariate shift.
\newblock In \emph{International conference on machine learning}, pages
  448--456. pmlr, 2015.

\bibitem[Jaderberg et~al.(2017)Jaderberg, Mnih, Czarnecki, Schaul, Leibo,
  Silver, and Kavukcuoglu]{jaderberg2017reinforcement}
Max Jaderberg, Volodymyr Mnih, Wojciech~Marian Czarnecki, Tom Schaul, Joel~Z
  Leibo, David Silver, and Koray Kavukcuoglu.
\newblock Reinforcement learning with unsupervised auxiliary tasks.
\newblock In \emph{International Conference on Learning Representations}, 2017.

\bibitem[Karl et~al.(2017)Karl, Soelch, Bayer, and van~der Smagt]{karl2017deep}
Maximilian Karl, Maximilian Soelch, Justin Bayer, and Patrick van~der Smagt.
\newblock Deep variational bayes filters: Unsupervised learning of state space
  models from raw data.
\newblock \emph{stat}, 1050:\penalty0 3, 2017.

\bibitem[Kenton and Toutanova(2019)]{kenton2019bert}
Jacob Devlin Ming-Wei~Chang Kenton and Lee~Kristina Toutanova.
\newblock Bert: Pre-training of deep bidirectional transformers for language
  understanding.
\newblock In \emph{Proceedings of NAACL-HLT}, pages 4171--4186, 2019.

\bibitem[Khadka and Tumer(2018)]{khadka2018evolution}
Shauharda Khadka and Kagan Tumer.
\newblock Evolution-guided policy gradient in reinforcement learning.
\newblock In \emph{Advances in Neural Information Processing Systems 31}, pages
  1188--1200, 2018.

\bibitem[Kingma and Ba(2014)]{adam}
Diederik Kingma and Jimmy Ba.
\newblock Adam: A method for stochastic optimization.
\newblock \emph{arXiv preprint arXiv:1412.6980}, 2014.

\bibitem[Kostrikov et~al.(2020)Kostrikov, Yarats, and
  Fergus]{kostrikov2020image}
Ilya Kostrikov, Denis Yarats, and Rob Fergus.
\newblock Image augmentation is all you need: Regularizing deep reinforcement
  learning from pixels.
\newblock \emph{arXiv preprint arXiv:2004.13649}, 2020.

\bibitem[Kostrikov et~al.(2021)Kostrikov, Nair, and
  Levine]{kostrikov2021offline}
Ilya Kostrikov, Ashvin Nair, and Sergey Levine.
\newblock Offline reinforcement learning with implicit q-learning.
\newblock In \emph{International Conference on Learning Representations}, 2021.

\bibitem[Kumar et~al.(2020{\natexlab{a}})Kumar, Agarwal, Ghosh, and
  Levine]{kumar2020implicit}
Aviral Kumar, Rishabh Agarwal, Dibya Ghosh, and Sergey Levine.
\newblock Implicit under-parameterization inhibits data-efficient deep
  reinforcement learning.
\newblock In \emph{International Conference on Learning Representations},
  2020{\natexlab{a}}.

\bibitem[Kumar et~al.(2020{\natexlab{b}})Kumar, Zhou, Tucker, and
  Levine]{kumar2020conservative}
Aviral Kumar, Aurick Zhou, George Tucker, and Sergey Levine.
\newblock Conservative q-learning for offline reinforcement learning.
\newblock \emph{Advances in Neural Information Processing Systems},
  33:\penalty0 1179--1191, 2020{\natexlab{b}}.

\bibitem[Kuznetsov et~al.(2020)Kuznetsov, Shvechikov, Grishin, and
  Vetrov]{kuznetsov2020controlling}
Arsenii Kuznetsov, Pavel Shvechikov, Alexander Grishin, and Dmitry Vetrov.
\newblock Controlling overestimation bias with truncated mixture of continuous
  distributional quantile critics.
\newblock In \emph{International Conference on Machine Learning}, pages
  5556--5566. PMLR, 2020.

\bibitem[Kuznetsov et~al.(2021)Kuznetsov, Grishin, Tsypin, Ashukha, and
  Vetrov]{kuznetsov2021automating}
Arsenii Kuznetsov, Alexander Grishin, Artem Tsypin, Arsenii Ashukha, and
  Dmitry~P Vetrov.
\newblock Automating control of overestimation bias for continuous
  reinforcement learning.
\newblock \emph{ArXiv abs/2110.13523}, 2021.

\bibitem[Laroche et~al.(2019)Laroche, Trichelair, and
  Des~Combes]{laroche2019safe}
Romain Laroche, Paul Trichelair, and Remi~Tachet Des~Combes.
\newblock Safe policy improvement with baseline bootstrapping.
\newblock In \emph{International Conference on Machine Learning}, pages
  3652--3661. PMLR, 2019.

\bibitem[Laskin et~al.(2020)Laskin, Srinivas, and Abbeel]{laskin2020curl}
Michael Laskin, Aravind Srinivas, and Pieter Abbeel.
\newblock Curl: Contrastive unsupervised representations for reinforcement
  learning.
\newblock In \emph{International Conference on Machine Learning}, pages
  5639--5650. PMLR, 2020.

\bibitem[Li et~al.(2006)Li, Walsh, and Littman]{li2006towards}
Lihong Li, Thomas~J Walsh, and Michael~L Littman.
\newblock Towards a unified theory of state abstraction for mdps.
\newblock In \emph{AI\&M}, 2006.

\bibitem[Lillicrap et~al.(2015)Lillicrap, Hunt, Pritzel, Heess, Erez, Tassa,
  Silver, and Wierstra]{DDPG}
Timothy~P Lillicrap, Jonathan~J Hunt, Alexander Pritzel, Nicolas Heess, Tom
  Erez, Yuval Tassa, David Silver, and Daan Wierstra.
\newblock Continuous control with deep reinforcement learning.
\newblock \emph{arXiv preprint arXiv:1509.02971}, 2015.

\bibitem[Lin(1992)]{expreplay1992}
Long-Ji Lin.
\newblock Self-improving reactive agents based on reinforcement learning,
  planning and teaching.
\newblock \emph{Machine learning}, 8\penalty0 (3-4):\penalty0 293--321, 1992.

\bibitem[Lin et~al.(2019)Lin, Baweja, Kantor, and Held]{lin2019adaptive}
Xingyu Lin, Harjatin Baweja, George Kantor, and David Held.
\newblock Adaptive auxiliary task weighting for reinforcement learning.
\newblock \emph{Advances in neural information processing systems}, 32, 2019.

\bibitem[Littman and Sutton(2001)]{littman2001predictive}
Michael Littman and Richard~S Sutton.
\newblock Predictive representations of state.
\newblock \emph{Advances in neural information processing systems}, 14, 2001.

\bibitem[Liu et~al.(2021)Liu, Zhang, Zhao, Qin, Zhu, Jian, Yu, and
  Liu]{liu2021return}
Guoqing Liu, Chuheng Zhang, Li~Zhao, Tao Qin, Jinhua Zhu, Li~Jian, Nenghai Yu,
  and Tie-Yan Liu.
\newblock Return-based contrastive representation learning for reinforcement
  learning.
\newblock In \emph{International Conference on Learning Representations}, 2021.

\bibitem[Lyle et~al.(2021)Lyle, Rowland, and Dabney]{lyle2021understanding}
Clare Lyle, Mark Rowland, and Will Dabney.
\newblock Understanding and preventing capacity loss in reinforcement learning.
\newblock In \emph{International Conference on Learning Representations}, 2021.

\bibitem[Mania et~al.(2018)Mania, Guy, and Recht]{mania2018simple}
Horia Mania, Aurelia Guy, and Benjamin Recht.
\newblock Simple random search provides a competitive approach to reinforcement
  learning.
\newblock \emph{arXiv preprint arXiv:1803.07055}, 2018.

\bibitem[Mnih et~al.(2015)Mnih, Kavukcuoglu, Silver, Rusu, Veness, Bellemare,
  Graves, Riedmiller, Fidjeland, Ostrovski, et~al.]{DQN}
Volodymyr Mnih, Koray Kavukcuoglu, David Silver, Andrei~A Rusu, Joel Veness,
  Marc~G Bellemare, Alex Graves, Martin Riedmiller, Andreas~K Fidjeland, Georg
  Ostrovski, et~al.
\newblock Human-level control through deep reinforcement learning.
\newblock \emph{Nature}, 518\penalty0 (7540):\penalty0 529--533, 2015.

\bibitem[Munk et~al.(2016)Munk, Kober, and Babu{\v{s}}ka]{munk2016learning}
Jelle Munk, Jens Kober, and Robert Babu{\v{s}}ka.
\newblock Learning state representation for deep actor-critic control.
\newblock In \emph{2016 IEEE 55th Conference on Decision and Control (CDC)},
  pages 4667--4673. IEEE, 2016.

\bibitem[Oord et~al.(2018)Oord, Li, and Vinyals]{oord2018representation}
Aaron van~den Oord, Yazhe Li, and Oriol Vinyals.
\newblock Representation learning with contrastive predictive coding.
\newblock \emph{arXiv preprint arXiv:1807.03748}, 2018.

\bibitem[Ota et~al.(2020)Ota, Oiki, Jha, Mariyama, and Nikovski]{ota2020can}
Kei Ota, Tomoaki Oiki, Devesh Jha, Toshisada Mariyama, and Daniel Nikovski.
\newblock Can increasing input dimensionality improve deep reinforcement
  learning?
\newblock In \emph{International Conference on Machine Learning}, pages
  7424--7433. PMLR, 2020.

\bibitem[Ota et~al.(2021)Ota, Jha, and Kanezaki]{ota2021training}
Kei Ota, Devesh~K Jha, and Asako Kanezaki.
\newblock Training larger networks for deep reinforcement learning.
\newblock \emph{arXiv preprint arXiv:2102.07920}, 2021.

\bibitem[Paszke et~al.(2019)Paszke, Gross, Massa, Lerer, Bradbury, Chanan,
  Killeen, Lin, Gimelshein, Antiga, et~al.]{paszke2019pytorch}
Adam Paszke, Sam Gross, Francisco Massa, Adam Lerer, James Bradbury, Gregory
  Chanan, Trevor Killeen, Zeming Lin, Natalia Gimelshein, Luca Antiga, et~al.
\newblock Pytorch: An imperative style, high-performance deep learning library.
\newblock In \emph{Advances in Neural Information Processing Systems}, pages
  8024--8035, 2019.

\bibitem[Pourchot and Sigaud(2018)]{pourchot2018cem}
Alo{\"\i}s Pourchot and Olivier Sigaud.
\newblock Cem-rl: Combining evolutionary and gradient-based methods for policy
  search.
\newblock \emph{arXiv preprint arXiv:1810.01222}, 2018.

\bibitem[Ravindran(2004)]{ravindran2004algebraic}
Balaraman Ravindran.
\newblock \emph{An algebraic approach to abstraction in reinforcement
  learning}.
\newblock University of Massachusetts Amherst, 2004.

\bibitem[Rezaei-Shoshtari et~al.(2022)Rezaei-Shoshtari, Zhao, Panangaden,
  Meger, and Precup]{rezaei2022continuous}
Sahand Rezaei-Shoshtari, Rosie Zhao, Prakash Panangaden, David Meger, and Doina
  Precup.
\newblock Continuous mdp homomorphisms and homomorphic policy gradient.
\newblock In \emph{Advances in Neural Information Processing Systems}, 2022.

\bibitem[Riedmiller et~al.(2018)Riedmiller, Hafner, Lampe, Neunert, Degrave,
  Wiele, Mnih, Heess, and Springenberg]{riedmiller2018learning}
Martin Riedmiller, Roland Hafner, Thomas Lampe, Michael Neunert, Jonas Degrave,
  Tom Wiele, Vlad Mnih, Nicolas Heess, and Jost~Tobias Springenberg.
\newblock Learning by playing solving sparse reward tasks from scratch.
\newblock In \emph{International conference on machine learning}, pages
  4344--4353. PMLR, 2018.

\bibitem[Salimans et~al.(2017)Salimans, Ho, Chen, Sidor, and
  Sutskever]{salimans2017evolution}
Tim Salimans, Jonathan Ho, Xi~Chen, Szymon Sidor, and Ilya Sutskever.
\newblock Evolution strategies as a scalable alternative to reinforcement
  learning.
\newblock \emph{arXiv preprint arXiv:1703.03864}, 2017.

\bibitem[Schaul et~al.(2016)Schaul, Quan, Antonoglou, and
  Silver]{PrioritizedExpReplay}
Tom Schaul, John Quan, Ioannis Antonoglou, and David Silver.
\newblock Prioritized experience replay.
\newblock In \emph{International Conference on Learning Representations},
  Puerto Rico, 2016.

\bibitem[Schulman et~al.(2015)Schulman, Levine, Abbeel, Jordan, and
  Moritz]{trpo}
John Schulman, Sergey Levine, Pieter Abbeel, Michael Jordan, and Philipp
  Moritz.
\newblock Trust region policy optimization.
\newblock In \emph{International Conference on Machine Learning}, pages
  1889--1897, 2015.

\bibitem[Schulman et~al.(2017)Schulman, Wolski, Dhariwal, Radford, and
  Klimov]{PPO}
John Schulman, Filip Wolski, Prafulla Dhariwal, Alec Radford, and Oleg Klimov.
\newblock Proximal policy optimization algorithms.
\newblock \emph{arXiv preprint arXiv:1707.06347}, 2017.

\bibitem[Schwarzer et~al.(2020)Schwarzer, Anand, Goel, Hjelm, Courville, and
  Bachman]{schwarzer2020data}
Max Schwarzer, Ankesh Anand, Rishab Goel, R~Devon Hjelm, Aaron Courville, and
  Philip Bachman.
\newblock Data-efficient reinforcement learning with self-predictive
  representations.
\newblock In \emph{International Conference on Learning Representations}, 2020.

\bibitem[Silver et~al.(2014)Silver, Lever, Heess, Degris, Wierstra, and
  Riedmiller]{DPG}
David Silver, Guy Lever, Nicolas Heess, Thomas Degris, Daan Wierstra, and
  Martin Riedmiller.
\newblock Deterministic policy gradient algorithms.
\newblock In \emph{International Conference on Machine Learning}, pages
  387--395, 2014.

\bibitem[Stooke et~al.(2021)Stooke, Lee, Abbeel, and
  Laskin]{stooke2021decoupling}
Adam Stooke, Kimin Lee, Pieter Abbeel, and Michael Laskin.
\newblock Decoupling representation learning from reinforcement learning.
\newblock In \emph{International Conference on Machine Learning}, pages
  9870--9879. PMLR, 2021.

\bibitem[Sutton et~al.(2011)Sutton, Modayil, Delp, Degris, Pilarski, White, and
  Precup]{sutton2011horde}
Richard~S Sutton, Joseph Modayil, Michael Delp, Thomas Degris, Patrick~M
  Pilarski, Adam White, and Doina Precup.
\newblock Horde: a scalable real-time architecture for learning knowledge from
  unsupervised sensorimotor interaction.
\newblock In \emph{The 10th International Conference on Autonomous Agents and
  Multiagent Systems-Volume 2}, pages 761--768, 2011.

\bibitem[Tennenholtz and Mannor(2019)]{tennenholtz2019natural}
Guy Tennenholtz and Shie Mannor.
\newblock The natural language of actions.
\newblock In \emph{International Conference on Machine Learning}, pages
  6196--6205. PMLR, 2019.

\bibitem[Todorov et~al.(2012)Todorov, Erez, and Tassa]{mujoco}
Emanuel Todorov, Tom Erez, and Yuval Tassa.
\newblock Mujoco: A physics engine for model-based control.
\newblock In \emph{IEEE/RSJ International Conference on Intelligent Robots and
  Systems (IROS)}, pages 5026--5033. IEEE, 2012.

\bibitem[van~der Pol et~al.(2020{\natexlab{a}})van~der Pol, Kipf, Oliehoek, and
  Welling]{van2020plannable}
Elise van~der Pol, Thomas Kipf, Frans~A Oliehoek, and Max Welling.
\newblock Plannable approximations to mdp homomorphisms: Equivariance under
  actions.
\newblock In \emph{Proceedings of the 19th International Conference on
  Autonomous Agents and MultiAgent Systems}, pages 1431--1439,
  2020{\natexlab{a}}.

\bibitem[van~der Pol et~al.(2020{\natexlab{b}})van~der Pol, Worrall, van Hoof,
  Oliehoek, and Welling]{van2020mdp}
Elise van~der Pol, Daniel Worrall, Herke van Hoof, Frans Oliehoek, and Max
  Welling.
\newblock Mdp homomorphic networks: Group symmetries in reinforcement learning.
\newblock \emph{Advances in Neural Information Processing Systems},
  33:\penalty0 4199--4210, 2020{\natexlab{b}}.

\bibitem[Van~Hoof et~al.(2016)Van~Hoof, Chen, Karl, van~der Smagt, and
  Peters]{van2016stable}
Herke Van~Hoof, Nutan Chen, Maximilian Karl, Patrick van~der Smagt, and Jan
  Peters.
\newblock Stable reinforcement learning with autoencoders for tactile and
  visual data.
\newblock In \emph{2016 IEEE/RSJ international conference on intelligent robots
  and systems (IROS)}, pages 3928--3934. IEEE, 2016.

\bibitem[Vaswani et~al.(2017)Vaswani, Shazeer, Parmar, Uszkoreit, Jones, Gomez,
  Kaiser, and Polosukhin]{vaswani2017attention}
Ashish Vaswani, Noam Shazeer, Niki Parmar, Jakob Uszkoreit, Llion Jones,
  Aidan~N Gomez, {\L}ukasz Kaiser, and Illia Polosukhin.
\newblock Attention is all you need.
\newblock \emph{Advances in neural information processing systems}, 30, 2017.

\bibitem[Watter et~al.(2015)Watter, Springenberg, Boedecker, and
  Riedmiller]{watter2015embed}
Manuel Watter, Jost Springenberg, Joschka Boedecker, and Martin Riedmiller.
\newblock Embed to control: A locally linear latent dynamics model for control
  from raw images.
\newblock \emph{Advances in neural information processing systems}, 28, 2015.

\bibitem[Whitney et~al.(2020)Whitney, Agarwal, Cho, and
  Gupta]{whitney2020dynamics}
William Whitney, Rajat Agarwal, Kyunghyun Cho, and Abhinav Gupta.
\newblock Dynamics-aware embeddings.
\newblock In \emph{International Conference on Learning Representations}, 2020.

\bibitem[Williams(1992)]{williams1992reinforce}
Ronald~J Williams.
\newblock Simple statistical gradient-following algorithms for connectionist
  reinforcement learning.
\newblock \emph{Machine learning}, 8\penalty0 (3-4):\penalty0 229--256, 1992.

\bibitem[Yarats et~al.(2021)Yarats, Zhang, Kostrikov, Amos, Pineau, and
  Fergus]{yarats2021improving}
Denis Yarats, Amy Zhang, Ilya Kostrikov, Brandon Amos, Joelle Pineau, and Rob
  Fergus.
\newblock Improving sample efficiency in model-free reinforcement learning from
  images.
\newblock In \emph{Proceedings of the AAAI Conference on Artificial
  Intelligence}, volume~35, pages 10674--10681, 2021.

\bibitem[Yarats et~al.(2022)Yarats, Fergus, Lazaric, and
  Pinto]{yarats2022mastering}
Denis Yarats, Rob Fergus, Alessandro Lazaric, and Lerrel Pinto.
\newblock Mastering visual continuous control: Improved data-augmented
  reinforcement learning.
\newblock In \emph{International Conference on Learning Representations}, 2022.

\bibitem[Zhang et~al.(2018)Zhang, Satija, and Pineau]{zhang2018decoupling}
Amy Zhang, Harsh Satija, and Joelle Pineau.
\newblock Decoupling dynamics and reward for transfer learning.
\newblock \emph{arXiv preprint arXiv:1804.10689}, 2018.

\bibitem[Zhang et~al.(2020)Zhang, McAllister, Calandra, Gal, and
  Levine]{zhang2020learning}
Amy Zhang, Rowan~Thomas McAllister, Roberto Calandra, Yarin Gal, and Sergey
  Levine.
\newblock Learning invariant representations for reinforcement learning without
  reconstruction.
\newblock In \emph{International Conference on Learning Representations}, 2020.

\end{thebibliography}
\bibliographystyle{plainnat}

\clearpage

\iftrue

\appendix

\doparttoc %
\faketableofcontents %

\part{Appendix}\label{appendix} %
\parttoc %

\clearpage

\section{TD7 Additional Details}  \label{appendix:sec:TD7}

\subsection{Algorithm} \label{appendix:sec:algorithm}

TD7 (TD3\texttt{+}4 additions) has several networks and sub-components:\
\begin{itemize}[nosep]
    \item Two value functions $(Q_{t+1, 1}, Q_{t+1, 2})$.
    \item Two target value functions $(Q_{t, 1}, Q_{t, 2})$.
    \item A policy network $\pi_{t+1}$.
    \item A target policy network $\pi_t$.
    \item An encoder, with sub-components $(f_{t+1}, g_{t+1})$.
    \item A fixed encoder, with sub-components $(f_t, g_t)$.
    \item A target fixed encoder with sub-components $(f_{t-1}, g_{t-1})$.
    \item A checkpoint policy $\pi_c$ and checkpoint encoder $f_c$ ($g$ is not needed). 
\end{itemize}

\textbf{Encoder:} The encoder is composed of two sub-networks~$(f_{t+1}(s), g_{t+1})(z^s, a)$, where each network outputs an embedding:
\begin{align}
    z^s := f(s), \qquad z^{sa} := g(z^s,a).
\end{align}
At each training step, the encoder is updated with the following loss:
\begin{align} \label{appendix:eqn:encoder_loss}
\Loss(f_{t+1},g_{t+1}) :=&~\Bigl( g_{t+1}(f_{t+1}(s),a) - |f_{t+1}(s')|_\times \Bigr)^2 \\
=&~\lp z_{t+1}^{sa} - |z_{t+1}^{s'}|_\times \rp^2,
\end{align}
where $|\cdot|_\times$ is the stop-gradient operation.

\textbf{Value function:} TD7 uses a pair of value functions (as motivated by TD3~\citep{fujimoto2018addressing}) $(Q_{t+1, 1}, Q_{t+1, 2})$, each taking input $[z_{t-1}^{s'a'}, z_{t-1}^{s'}, s', a']$. At each training step, both value functions are updated with the following loss: 
\begin{align} \label{appendix:eqn:critic_loss} 
\Loss(Q_{t+1}) :=&~\text{Huber}\Bigl( \texttt{target} - Q_{t+1}(z_{t}^{sa}, z_{t}^s, s, a) \Bigr),\\
\texttt{target} :=&~r + \y \text{ clip} \bigl( \min \lp Q_{t,1} (x), Q_{t,2} (x) \rp, Q_\text{min}, Q_\text{max} \bigr), \label{appendix:eqn:Qtarget}\\
x :=&~[z_{t-1}^{s'a'}, z_{t-1}^{s'}, s', a'],\\
a' :=&~\pi_{t}(z^{s'}_{t-1}, s') + \e, \\
\e \sim&~\text{clip}(\N(0,\sigma^2), -c, c). 
\end{align}
Taking the minimum of the value functions is from TD3's Clipped Double Q-learning (CDQ)~\citep{fujimoto2018addressing}). 
The use of Huber loss~\citep{huber1964robust} is in accordance to TD3 with the Loss-Adjusted Prioritized (LAP) experience replay~\citep{fujimoto2020equivalence}. $a'$ is sampled and clipped in the same manner as TD3~\citep{fujimoto2018addressing}. 
The same embeddings~$(z_{t}^{sa}, z_{t}^s)$ are used for each value function. $Q_\text{min}$ and $Q_\text{max}$ are updated at each time step:
\begin{align} 
    Q_\text{min} \leftarrow&~\min \lp Q_\text{min}, \texttt{target} \rp, \label{appendix:eqn:clipping_update1} \\
    Q_\text{max} \leftarrow&~\max \lp Q_\text{max}, \texttt{target} \rp, \label{appendix:eqn:clipping_update2} 
\end{align}
where \texttt{target} is defined by \autoref{appendix:eqn:Qtarget}.

\textbf{Policy:} TD7 uses a single policy network which takes input $[z^s, s]$. 
On every second training step (according to TD3's delayed policy updates~\citep{fujimoto2018addressing}) the policy~$\pi_{t+1}$ is updated with the following loss:  
\begin{align} \label{appendix:eqn:actor_loss}
\Loss(\pi_{t+1}) :=&~-Q + \lambda \lvert \E_{s \sim D} \lb Q \rb \rvert_\times \lp a_\pi - a \rp^2, \\
Q :=&~0.5 \lp Q_{t+1,1}(x) + Q_{t+1,2}(x) \rp \\
x :=&~[z_{t}^{sa_\pi}, z_{t}^s, s, a_\pi], \\
a_\pi :=&~\pi_{t+1}(z_{t}^s, s).
\end{align}
The policy loss is the deterministic policy gradient (DPG)~\citep{DPG} with a behavior cloning term to regularize~\citep{fujimoto2021minimalist}. $\lambda$ is set to $0$ for online RL. 

After every \texttt{target\_update\_frequency}~(250) training steps, the iteration is updated and each target (and fixed) network copies the network of the higher iteration:
\begin{align} \label{appendix:eqn:target}
    (Q_{t, 1}, Q_{t, 2}) &\leftarrow (Q_{t+1, 1}, Q_{t+1, 2}), \\
    \pi_t &\leftarrow \pi_{t+1}, \\
    (f_{t-1}, g_{t-1}) &\leftarrow (f_t, g_t), \\
    (f_t, g_t) &\leftarrow (f_{t+1}, g_{t+1}).
\end{align}

The checkpoint policy and checkpoint encoder are only used at test time (see \autoref{appendix:sec:checkpoints}). 

\textbf{LAP:} Gathered experience is stored in a replay buffer~\citep{expreplay1992} and sampled according to LAP~\citep{fujimoto2020equivalence}, a prioritized replay buffer $D$~\citep{PrioritizedExpReplay} where a transition tuple $i:=(s,a,r,s')$ is sampled with probability 
\begin{align} \label{appendix:eqn:LAP}
    p(i) =&~\frac{\max \lp |\delta(i)|^\al, 1\rp}{\sum_{j \in D} \max \lp |\delta(j)|^\al, 1\rp}, \\
    |\delta(i)| :=&~ \max \Bigl( \left| Q_{t+1,1}(z_{t}^{sa}, z_{t}^s, s, a) - \texttt{target} \right|, \left| Q_{t+1,2}(z_{t}^{sa}, z_{t}^s, s, a) - \texttt{target} \right| \Bigr),
\end{align}
where \texttt{target} is defined by \autoref{appendix:eqn:Qtarget}. As suggested by \cite{fujimoto2020equivalence}, $|\delta(i)|$ is defined by the maximum absolute error of both value functions. The amount of prioritization used is controlled by a hyperparameter $\al$. New transitions are assigned the maximum priority of any sample in the replay buffer. 

We outline the train function of TD7 in \autoref{appendix:alg:TD7}. There is no difference between the train function of online TD7 and offline TD7 other than the value of $\lambda$, which is $0$ for online and $0.1$ for offline. 

\begin{algorithm}[ht]
\small
   \caption{TD7 Train Function} \label{appendix:alg:TD7}
\begin{algorithmic}[1]
    \State Sample transition from LAP replay buffer with probability~(\autoref{appendix:eqn:LAP}).
    \State Train encoder (\autoref{appendix:eqn:encoder_loss}).
    \State Train value function (\autoref{appendix:eqn:critic_loss}). 
    \State Update $(Q_\text{min}, Q_\text{max})$ (Equations~\ref{appendix:eqn:clipping_update1}~\&~\ref{appendix:eqn:clipping_update2}).
    \If{$i \text{ mod } \texttt{policy\_update\_frequency} = 0$}
    \State Train policy (\autoref{appendix:eqn:actor_loss}).
    \EndIf
    \If{$i \text{ mod } \texttt{target\_update\_frequency} = 0$}
    \State Update target networks (\autoref{appendix:eqn:target}). 
    \EndIf
\end{algorithmic}
\end{algorithm}

\clearpage

\subsection{Hyperparameters}  \label{appendix:sec:TD7_hyperparameters}

The action space is assumed to be in the range~$[-1,1]$ (and is normalized if otherwise). Besides a few exceptions mentioned below, hyperparameters (and architecture) are taken directly from TD3~\url{https://github.com/sfujim/TD3}. 

\begin{table}[ht]
\caption{TD7 Hyperparameters.}\label{table:hyperparameters}
\centering
\begin{tabular}{cll}
\toprule
& Hyperparameter & Value \\
\midrule
\multirow{3}{*}{\shortstack{TD3\\\citep{fujimoto2018addressing}}} 
& Target policy noise~$\sigma$      & $\N(0,0.2^2)$ \\
& Target policy noise clipping~$c$  & $(-0.5, 0.5)$ \\
& Policy update frequency           & $2$ \\
\midrule
\multirow{2}{*}{\shortstack{LAP\\\citep{fujimoto2020equivalence}}} 
& Probability smoothing $\alpha$    & $0.4$ \\
& Minimum priority                  & $1$ \\
\midrule
\multirow{2}{*}{\shortstack{TD3+BC\\\citep{fujimoto2021minimalist}}} 
& Behavior cloning weight $\lambda$ (Online)  & $0.0$ \\
& Behavior cloning weight $\lambda$ (Offline) & $0.1$ \\
\midrule
\multirow{5}{*}{\shortstack{Policy Checkpoints\\(\autoref{appendix:sec:checkpoints})}}& Checkpoint criteria & minimum \\
& Early assessment episodes & $1$ \\
& Late assessment episodes & $20$ \\
& Early time steps & $750$k \\
& Criteria reset weight & $0.9$ \\
\midrule
\multirow{2}{*}{Exploration} & Initial random exploration time steps & $25$k \\
& Exploration noise    & $\N(0,0.1^2)$ \\
\midrule
\multirow{4}{*}{Common}
& Discount factor~$\y$      & $0.99$ \\
& Replay buffer capacity    & $1$M \\
& Mini-batch size           & $256$ \\
& Target update frequency   & $250$ \\
\midrule
\multirow{2}{*}{Optimizer} 
& (Shared) Optimizer        & Adam~\citep{adam} \\
& (Shared) Learning rate    & $3\text{e}-4$ \\
\bottomrule
\end{tabular}
\end{table} %

Besides algorithmic difference from TD3, there are three implementation-level changes: 
\begin{enumerate}[nosep]
    \item Rather than only using $Q_1$, both value functions are used when updating the policy~(\autoref{appendix:eqn:actor_loss}).
    \item The value function uses ELU activation functions~\citep{clevert2015fast} rather than ReLU activation functions. 
    \item The target network is updated periodically (every 250 time steps) rather than using an exponential moving average at every time step. This change is necessary due to the use of fixed encoders. 
\end{enumerate}

We evaluate the importance of these changes in our ablation study~(\autoref{appendix:sec:ablation}). Network architecture details are described in \autoref{pseudo:TD7}. 

\clearpage

\begin{tcolorbox}[title={\begin{pseudo}\label{pseudo:TD7}TD7 Network Details\end{pseudo}}]
\textbf{Variables:}
\begin{verbatim}
zs_dim = 256
\end{verbatim}

\vspace{-8pt}
\hrulefill

\textbf{Value $Q$ Network:}

$\triangleright$ TD7 uses two value networks each with the same network and forward pass. 
\begin{verbatim}
l0 = Linear(state_dim + action_dim, 256)
l1 = Linear(zs_dim * 2 + 256, 256)
l2 = Linear(256, 256)
l3 = Linear(256, 1)
\end{verbatim} 

\textbf{Value $Q$ Forward Pass:}
\begin{verbatim}
input = concatenate([state, action])
x = AvgL1Norm(l0(inuput))
x = concatenate([zsa, zs, x])
x = ELU(l1(x))
x = ELU(l2(x))
value = l3(x)
\end{verbatim} 

\vspace{-8pt}
\hrulefill

\textbf{Policy $\pi$ Network:}
\begin{verbatim}
l0 = Linear(state_dim, 256)
l1 = Linear(zs_dim + 256, 256)
l2 = Linear(256, 256)
l3 = Linear(256, action_dim)
\end{verbatim} 

\textbf{Policy $\pi$ Forward Pass:}
\begin{verbatim}
input = state
x = AvgL1Norm(l0(input))
x = concatenate([zs, x])
x = ReLU(l1(x))
x = ReLU(l2(x))
action = tanh(l3(x))
\end{verbatim} 

\vspace{-8pt}
\hrulefill

\textbf{State Encoder $f$ Network:}
\begin{verbatim}
l1 = Linear(state_dim, 256)
l2 = Linear(256, 256)
l3 = Linear(256, zs_dim)
\end{verbatim} 

\textbf{State Encoder $f$ Forward Pass:}
\begin{verbatim}
input = state
x = ELU(l1(input))
x = ELU(l2(x))
zs = AvgL1Norm(l3(x))
\end{verbatim} 

\vspace{-8pt}
\hrulefill

\textbf{State-Action Encoder $g$ Network:}
\begin{verbatim}
l1 = Linear(action_dim + zs_dim, 256)
l2 = Linear(256, 256)
l3 = Linear(256, zs_dim)
\end{verbatim} 

\textbf{State-Action Encoder $g$ Forward Pass:}
\begin{verbatim}
input = concatenate([action, zs])
x = ELU(l1(input))
x = ELU(l2(x))
zsa = l3(x)
\end{verbatim} 
\end{tcolorbox}

\clearpage

\section{Experimental Details} \label{appendix:sec:experimental}

\textbf{Environment.} Our experimental evaluation is based on the MuJoCo simulator~\citep{mujoco} with tasks defined by OpenAI gym~\citep{OpenAIGym} using the v4 environments. No modification are made to the state, action, or reward space. 

\textbf{Terminal transitions.} All methods use a discount factor $\y=0.99$ for non-terminal transitions and $\y=0$ for terminal transitions. The final transition from an episode which ends due to a time limit is not considered terminal. 

\textbf{Exploration.} To fill the replay buffer before training, all methods initially collect data by following a uniformly random policy for the first $25$k time steps ($256$ for TQC). This initial data collection is accounted for in all graphs and tables, meaning ``time steps'' refers to the number of environment interactions, rather than the number of training steps. Methods based on the deterministic TD3 add Gaussian noise to the policy. Methods based on the stochastic SAC do not add noise.

\textbf{Evaluation.} Agents are evaluated every $5000$ time steps, taking the average undiscounted sum of rewards over $10$ episodes. Each experiment is repeated over $10$ seeds. Evaluations use a deterministic policy, meaning there is no noise added to the policy and stochastic methods use the mean action. Methods with our proposed policy checkpoints use the checkpoint policy during evaluations. Evaluations are considered entirely independent from training, meaning no data is saved, nor is any network updated.

\textbf{Visualization and tables.} As aforementioned, time steps in figures and tables refers to the number of environment interactions (all methods are trained once per environment interaction, other than the initial random data collection phase). The value reported by the table corresponds to the average evaluation over the $10$ seeds, at the given time step (300k, 1M or 5M for online results and 1M for offline results). The shaded area in figures and the $\pm$ term in tables refers to a $97.5\%$ confidence interval. Given that methods are evaluated over $10$ seeds, this confidence interval $\texttt{CI}$ is computed by
\begin{equation}
    \texttt{CI} = \frac{1.96}{\sqrt{10}} \sigma,
\end{equation}
where $\sigma$ is the sample standard deviation with Bessel's correction at the corresponding evaluation. Unless stated otherwise, all curves are smoothed uniformly over a window of $10$ evaluations.

\textbf{Offline Experiments.} Offline results are based on the D4RL datasets~\citep{fu2021benchmarks}, using the v2 version. No modification are made to the state, action, or reward space. The reported performance is from a final evaluation occurring after 1M training steps, which uses the average D4RL score over $10$ episodes. TD7 results are repeated over $10$ seeds. Baseline results are taken from other papers, and may not be based on an identical evaluation protocol in terms of number of episodes and seeds used. 

\textbf{Software.} 
We use the following software versions:
\begin{itemize}[nosep]
    \item Python 3.9.13
    \item Pytorch 2.0.0~\citep{paszke2019pytorch}
    \item CUDA version 11.8
    \item Gym 0.25.0~\citep{OpenAIGym}
    \item MuJoCo 2.3.3~\citep{mujoco}
\end{itemize}

\clearpage

\section{Baselines}
 
\subsection{TD3 \& TD3+OFE Hyperparameters}

Our TD3 baseline~\citep{fujimoto2018addressing} uses the author implementation \url{https://github.com/sfujim/TD3}. Our TD3+OFE baseline~\citep{ota2020can} uses the aforementioned TD3 code alongside a re-implementation of OFENet. For hyperparameters, the action space is assumed to be in the range~$[-1,1]$. 

\textbf{Environment-specific hyperparameters.} OFE uses a variable number of encoder layers depending on the environment. However, the authors also provide an offline approach for hyperparameter tuning based on the representation loss. Our choice of layers per environment is the outcome of their hyperparameter tuning. Additionally, when training the state-action encoder, OFE drops certain state dimensions which correspond to external forces which are hard to predict. In practice this only affects Humanoid, which uses $292$ out of a possible state $376$ dimensions on Humanoid. All other tasks use the full state space.

Hyperparameters of both methods are listed in \autoref{table:TD3_hyperparameters}. Network architecture details of TD3 are described in \autoref{pseudo:TD3}. Network architecture details of TD3+OFE are described in \autoref{pseudo:TD3+OFE} and \autoref{pseudo:TD3+OFE2}.

\begin{table}[ht]
\caption{TD3 \& TD3+OFE Hyperparameters.}\label{table:TD3_hyperparameters}
\centering
\begin{tabular}{cll}
\toprule
& Hyperparameter & Value \\
\midrule
\multirow{5}{*}{\shortstack{OFE\\\citep{ota2020can}}} 
& Encoder layers & \\
\cmidrule{2-3}
& HalfCheetah & $8$ \\
& Hopper & $6$ \\
& Walker2d & $6$ \\
& Ant & $6$ \\
& Humanoid & $8$ \\
\midrule
\multirow{3}{*}{\shortstack{TD3\\\citep{fujimoto2018addressing}}} 
& Target policy noise~$\sigma$      & $\N(0,0.2^2)$ \\
& Target policy noise clipping~$c$  & $(-0.5, 0.5)$ \\
& Policy update frequency           & $2$ \\
\midrule
\multirow{2}{*}{Exploration} & Initial random exploration time steps & $25$k \\
& Exploration noise    & $\N(0,0.1^2)$ \\
\midrule
\multirow{4}{*}{Common}
& Discount factor~$\y$      & $0.99$ \\
& Replay buffer capacity    & $1$M \\
& Mini-batch size           & $256$ \\
& Target update rate~$\tau$ & $0.005$ \\
\midrule
\multirow{2}{*}{Optimizer} 
& (Shared) Optimizer        & Adam~\citep{adam} \\
& (Shared) Learning rate    & $3\text{e}-4$ \\
\bottomrule
\end{tabular}
\end{table} 

\clearpage

\begin{tcolorbox}[title={\begin{pseudo}\label{pseudo:TD3}TD3 Network Details\end{pseudo}}]
\textbf{Value $Q$ Network:}

$\triangleright$ TD3 uses two value networks each with the same network and forward pass. 
\begin{verbatim}
l1 = Linear(state_dim + action_dim, 256)
l2 = Linear(256, 256)
l3 = Linear(256, 1)
\end{verbatim} 

\textbf{Value $Q$ Forward Pass:}
\begin{verbatim}
input = concatenate([state, action])
x = ReLU(l1(input))
x = ReLU(l2(x))
value = l3(x)
\end{verbatim} 

\vspace{-8pt}
\hrulefill

\textbf{Policy $\pi$ Network:}
\begin{verbatim}
l1 = Linear(state_dim, 256)
l2 = Linear(256, 256)
l3 = Linear(256, action_dim)
\end{verbatim} 

\textbf{Policy $\pi$ Forward Pass:}
\begin{verbatim}
input = state
x = ReLU(l1(input))
x = ReLU(l2(x))
action = tanh(l3(x))
\end{verbatim} 
\end{tcolorbox}

\begin{tcolorbox}[title={\begin{pseudo}\label{pseudo:TD3+OFE}TD3+OFE Network Details (TD3)\end{pseudo}}]
\textbf{Variables:}
\begin{verbatim}
zs_dim = 240
zsa_dim = 240
num_layers = 6 (or 8)
zs_hdim = zs_dim/num_layers = 40 (or 30)
zsa_hdim = zsa_dim/num_layers = 40 (or 30)
target_dim = state_dim if not Humanoid else 292
\end{verbatim} 

\vspace{-8pt}
\hrulefill

\textbf{Value $Q$ Network:}

$\triangleright$ TD3+OFE uses two value networks each with the same network and forward pass. 
\begin{verbatim}
l1 = Linear(state_dim + action_dim + zs_dim + zsa_dim, 256)
l2 = Linear(256, 256)
l3 = Linear(256, 1)
\end{verbatim} 

\textbf{Value $Q$ Forward Pass:}
\begin{verbatim}
input = zsa
x = ReLU(l1(input))
x = ReLU(l2(x))
value = l3(x)
\end{verbatim} 

\vspace{-8pt}
\hrulefill

\textbf{Policy $\pi$ Network:}
\begin{verbatim}
l1 = Linear(state_dim + zs_dim, 256)
l2 = Linear(256, 256)
l3 = Linear(256, action_dim)
\end{verbatim} 

\textbf{Policy $\pi$ Forward Pass:}
\begin{verbatim}
input = zs
x = ReLU(l1(input))
x = ReLU(l2(x))
action = tanh(l3(x))
\end{verbatim} 
\end{tcolorbox}

\begin{tcolorbox}[title={\begin{pseudo}\label{pseudo:TD3+OFE2}TD3+OFE Network Details (OFE)\end{pseudo}}]
\textbf{State Encoder $f^s$ Network:}
\begin{verbatim}
l1 = Linear(state_dim, zs_hdim)
l2 = Linear(state_dim + zs_hdim, zs_hdim)
l3 = Linear(state_dim + zs_hdim * 2, zs_hdim)
...
l_num_layers = Linear(state_dim + zs_hdim * (num_layers-1), zs_hdim)
\end{verbatim} 

\textbf{State Encoder $f^s$ Forward Pass:}
\begin{verbatim}
input = state
x = swish(batchnorm(l1(input)))
input = concatenate([input, x])
x = swish(batchnorm(l2(input)))
...
x = swish(batchnorm(l_num_layers(input)))
output = concatenate([input, x])
\end{verbatim} 

\vspace{-8pt}
\hrulefill

\textbf{State-Action Encoder $f^{sa}$ Network:}
\begin{verbatim}

l1 = Linear(state_dim + action_dim + zs_dim, zsa_hdim)
l2 = Linear(state_dim + action_dim + zs_dim + zsa_hdim, zsa_hdim)
l3 = Linear(state_dim + action_dim + zs_dim + zsa_hdim * 2, zsa_hdim)
...
l_num_layers = Linear(state_dim + action_dim 
    + zs_dim + zsa_hdim * (num_layers-1), zsa_hdim)
    
final_layer = Linear(state_dim + action_dim + zs_dim + zsa_dim, target_dim)
\end{verbatim} 

\textbf{State-Action Encoder $f^{sa}$ Forward Pass:}
\begin{verbatim}
input = concatenate([action, zs])
x = swish(batchnorm(l1(input)))
input = concatenate([input, x])
x = swish(batchnorm(l2(input)))
...
x = swish(batchnorm(l_num_layers(input)))
output = concatenate([input, x])
\end{verbatim} 

\textbf{Final Layer $t$ Forward Pass:}
\begin{verbatim}
input = zsa
output = final_layer(zsa)
\end{verbatim} 
\end{tcolorbox}

\clearpage

\subsection{SAC \& TQC Hyperparameters}

Our SAC baseline~\citep{haarnoja2018soft} is based on our TD3 implementation, keeping hyperparameters constant when possible. Remaining details are based on the author implementation \url{https://github.com/haarnoja/sac}. Our TQC baseline~\citep{kuznetsov2020controlling} uses the author PyTorch implementation \url{https://github.com/SamsungLabs/tqc_pytorch} (with evaluation code kept consistent for all methods). For hyperparameters, the action space is assumed to be in the range~$[-1,1]$. 

TQC is based on SAC and uses similar hyperparameters. One exception is the use of an $\e$ offset. As suggested in the appendix of \cite{haarnoja2018soft}, the log probability of the Tanh Normal distribution is calculated as follows:
\begin{equation}
    \log \pi(a|s) = \log N(u|s) - \log(1 - \text{tanh}(u)^2 + \e),
\end{equation}
where $u$ is the pre-activation value of the action, sampled from $N$, where $N$ is distributed according to $\N(\mu, \sigma^2)$, from the outputs $\mu$ and $\log \sigma$ of the actor network. \cite{kuznetsov2020controlling} use an alternate calculation of the log of the Tanh Normal distribution which eliminates the need for an $\e$ offset: 
\begin{equation}
    \log \pi(a|s) = \log N(u|s) - \lp 2 \log(2) + \log(\text{sigmoid}(2u)) + \log(\text{sigmoid}(-2u)) \rp.
\end{equation}

\textbf{Environment-specific hyperparameters.} TQC varies the number of dropped quantiles depending on the environment. While it is possible to select this quantity based on heuristics~\citep{kuznetsov2021automating}, we use the author suggested values for each environment.

Hyperparameters of SAC are listed in \autoref{table:SAC_hyperparameters}. Network architecture details of SAC are described in \autoref{pseudo:SAC}.

Hyperparameters of TQC are listed in \autoref{table:TQC_hyperparameters}. Network architecture details of TQC are described in \autoref{pseudo:TQC}.

\clearpage

\begin{table}[ht]
\caption{SAC Hyperparameters.}\label{table:SAC_hyperparameters}
\centering
\begin{tabular}{cll}
\toprule
& Hyperparameter & Value \\
\midrule
\multirow{3}{*}{\shortstack{SAC\\\citep{haarnoja2018soft}}} & Target Entropy & $-\texttt{action\_dim}$ \\
& Policy $\log$ standard deviation clamp & $[-20, 2]$ \\
& Numerical stability offset $\e$ & $1\text{e}-6$ \\
\midrule
Exploration & Initial random exploration time steps & $25$k \\
\midrule
\multirow{4}{*}{Common}
& Discount factor~$\y$      & $0.99$ \\
& Replay buffer capacity    & $1$M \\
& Mini-batch size           & $256$ \\
& Target update rate~$\tau$ & $0.005$ \\
\midrule
\multirow{2}{*}{Optimizer} 
& (Shared) Optimizer        & Adam~\citep{adam} \\
& (Shared) Learning rate    & $3\text{e}-4$ \\
\bottomrule
\end{tabular}
\end{table} 

\begin{tcolorbox}[title={\begin{pseudo}\label{pseudo:SAC}SAC Network Details\end{pseudo}}]
\textbf{Value $Q$ Network:}

$\triangleright$ SAC uses two value networks each with the same network and forward pass. 
\begin{verbatim}
l1 = Linear(state_dim + action_dim, 256)
l2 = Linear(256, 256)
l3 = Linear(256, 1)
\end{verbatim} 

\textbf{Value $Q$ Forward Pass:}
\begin{verbatim}
input = concatenate([state, action])
x = ReLU(l1(input))
x = ReLU(l2(x))
value = l3(x)
\end{verbatim} 

\vspace{-8pt}
\hrulefill

\textbf{Policy $\pi$ Network:}
\begin{verbatim}
l1 = Linear(state_dim, 256)
l2 = Linear(256, 256)
l3 = Linear(256, action_dim * 2)
\end{verbatim} 

\textbf{Policy $\pi$ Forward Pass:}
\begin{verbatim}
input = state
x = ReLU(l1(input))
x = ReLU(l2(x))
mean, log_std = l3(x)
x = Normal(mean, exp(log_std.clip(-20, 2))).sample()
action = tanh(x)
\end{verbatim} 
\end{tcolorbox}

\clearpage

\begin{table}[ht]
\setlength{\tabcolsep}{4pt}
\caption{TQC Hyperparameters.}\label{table:TQC_hyperparameters}
\centering
\begin{tabular}{cll}
\toprule
& Hyperparameter & Value \\
\midrule
\multirow{7}{*}{\shortstack{TQC\\\citep{kuznetsov2020controlling}}} & Number of networks & $5$ \\
& Quantiles per network & $25$ \\
\cmidrule{2-3}
& Quantiles dropped per network & \\
\cmidrule{2-3}
& HalfCheetah & $0$ \\
& Hopper & $5$ \\
& Walker2d & $2$ \\
& Ant & $2$ \\
& Humanoid & $2$ \\
\midrule
\multirow{2}{*}{\shortstack{SAC\\\citep{haarnoja2018soft}}} & Target Entropy & $-\texttt{action\_dim}$ \\
& Policy $\log$ standard deviation clamp & $[-20, 2]$ \\
\midrule
Exploration & Initial random exploration time steps & $256$ \\
\midrule
\multirow{4}{*}{Common}
& Discount factor~$\y$      & $0.99$ \\
& Replay buffer capacity    & $1$M \\
& Mini-batch size           & $256$ \\
& Target update rate~$\tau$ & $0.005$ \\
\midrule
\multirow{2}{*}{Optimizer} 
& (Shared) Optimizer        & Adam~\citep{adam} \\
& (Shared) Learning rate    & $3\text{e}-4$ \\
\bottomrule
\end{tabular}
\end{table} 

\begin{tcolorbox}[title={\begin{pseudo}\label{pseudo:TQC}TQC Network Details\end{pseudo}}]
\textbf{Value $Q$ Network:}

$\triangleright$ TQC uses five value networks each with the same network and forward pass. 
\begin{verbatim}
l1 = Linear(state_dim + action_dim, 512)
l2 = Linear(512, 512)
l3 = Linear(512, 512)
l4 = Linear(512, 25)
\end{verbatim} 

\textbf{Value $Q$ Forward Pass:}
\begin{verbatim}
input = concatenate([state, action])
x = ReLU(l1(input))
x = ReLU(l2(x))
x = ReLU(l3(x))
value = l4(x)
\end{verbatim} 

\vspace{-8pt}
\hrulefill

\textbf{Policy $\pi$ Network:}
\begin{verbatim}
l1 = Linear(state_dim, 256)
l2 = Linear(256, 256)
l3 = Linear(256, action_dim * 2)
\end{verbatim} 

\textbf{Policy $\pi$ Forward Pass:}
\begin{verbatim}
input = state
x = ReLU(l1(input))
x = ReLU(l2(x))
mean, log_std = l3(x)
x = Normal(mean, exp(log_std.clip(-20, 2))).sample()
action = tanh(x)
\end{verbatim} 
\end{tcolorbox}

\clearpage

\subsection{Offline RL Baselines}

We use four offline RL baseline methods. The results of each method are obtained by re-running the author-provided code with the following commands:

\textbf{CQL}~\citep{kumar2020conservative}. \url{https://github.com/aviralkumar2907/CQL} \texttt{commit d67dbe9}

\begin{lcverbatim}
python examples/cql_mujoco_new.py --env=ENV \
--policy_lr=3e-5 --langrange_thresh=-1 --min_q_weight=10
\end{lcverbatim}

Setting the Lagrange threshold below 0 means the Lagrange version of the code is not used (as dictated by the settings defined in the CQL paper). Default hyperparameters in the GitHub performed substantially worse (total normalized score: 221.3 \textcolor{gray}{$\pm$ 34.4}). Some modifications were made to the evaluation code to match the evaluation protocol used by TD7. 

\textbf{TD3+BC}~\citep{fujimoto2021minimalist}. \url{https://github.com/sfujim/TD3_BC} \texttt{commit 8791ad7}

\begin{lcverbatim}
python main.py --env=ENV
\end{lcverbatim}

\textbf{IQL}~\citep{kostrikov2021offline}. \url{https://github.com/ikostrikov/implicit_q_learning} \texttt{commit 09d7002}

\begin{lcverbatim}
python train_offline.py --env_name=ENV --config=configs/mujoco_config.py
\end{lcverbatim}

\textbf{$\mathcal{X}$-QL}~\citep{garg2023extreme}. \url{https://github.com/Div99/XQL} \texttt{commit dff09af}

\begin{lcverbatim}
python train_offline.py --env_name=ENV --config=configs/mujoco_config.py \
--max_clip=7 --double=True --temp=2 --batch_size=256
\end{lcverbatim}
This setting corresponds to the $\mathcal{X}$-QL-C version of the method where  hyperparameters are fixed across datasets. 

\clearpage

\section{Design Study} \label{appendix:sec:design}

In this section we analyze the space of design choices for SALE. All results are based on modifying TD7 without policy checkpoints. A summary of all results is contained in \autoref{appendix:table:main_design}. Each choice is described in detail in the subsections that follow. 

\begin{table*}[ht]
\centering
\small
\setlength{\tabcolsep}{4pt}
\newcolumntype{Y}{>{\centering\arraybackslash}X} %
\caption{Average performance on the MuJoCo benchmark at 1M time steps. $\pm$~captures a $95\%$ confidence interval around the average performance. Results are over $10$ seeds.} \label{appendix:table:main_design}
\begin{tabularx}{\textwidth}{lYYYYY}
\toprule
Algorithm & HalfCheetah & Hopper & Walker2d & Ant & Humanoid \\ 
\midrule
TD7 (no checkpoints) & 17123 \textcolor{gray}{$\pm$ 296\po} & \po3361 \textcolor{gray}{$\pm$ 429\po} & \po5718 \textcolor{gray}{$\pm$ 308\po} & \po8605 \textcolor{gray}{$\pm$ 1008} & \po7381 \textcolor{gray}{$\pm$ 172\po} \\
TD3 & 10574 \textcolor{gray}{$\pm$ 897\po} & \po3226 \textcolor{gray}{$\pm$ 315\po} & \po3946 \textcolor{gray}{$\pm$ 292\po} & \po3942 \textcolor{gray}{$\pm$ 1030} & \po5165 \textcolor{gray}{$\pm$ 145\po} \\
\midrule
\multicolumn{6}{c}{Learning Target (\autoref{appendix:subsec:learning_target})} \\
\midrule
$s'$ & 16438 \textcolor{gray}{$\pm$ 219\po} & \po3507 \textcolor{gray}{$\pm$ 313\po} & \po5882 \textcolor{gray}{$\pm$ 262\po} & \po7415 \textcolor{gray}{$\pm$ 753\po} & \po6337 \textcolor{gray}{$\pm$ 387\po} \\
$z^{s'}_\text{target}$ & 17317 \textcolor{gray}{$\pm$ 65\po\po} & \po3034 \textcolor{gray}{$\pm$ 647\po} & \po6227 \textcolor{gray}{$\pm$ 218\po} & \po7939 \textcolor{gray}{$\pm$ 334\po} & \po6984 \textcolor{gray}{$\pm$ 578\po} \\
$z^s$ and $r$ & 17089 \textcolor{gray}{$\pm$ 219\po} & \po3264 \textcolor{gray}{$\pm$ 670\po} & \po6255 \textcolor{gray}{$\pm$ 220\po} & \po8300 \textcolor{gray}{$\pm$ 675\po} & \po6841 \textcolor{gray}{$\pm$ 252\po} \\
$z^{s'a'}$ & 12445 \textcolor{gray}{$\pm$ 417\po} & \po3009 \textcolor{gray}{$\pm$ 528\po} & \po4757 \textcolor{gray}{$\pm$ 352\po} & \po6288 \textcolor{gray}{$\pm$ 570\po} & \po5657 \textcolor{gray}{$\pm$ 110\po} \\
\midrule
\multicolumn{6}{c}{Network Input (\autoref{appendix:subsec:network_input})} \\
\midrule
$Q$, remove $z^{sa}$ & 16948 \textcolor{gray}{$\pm$ 250\po} & \po2914 \textcolor{gray}{$\pm$ 756\po} & \po5746 \textcolor{gray}{$\pm$ 254\po} & \po6911 \textcolor{gray}{$\pm$ 552\po} & \po6907 \textcolor{gray}{$\pm$ 443\po} \\
$Q$, remove $z^{s}$ & 16856 \textcolor{gray}{$\pm$ 241\po} & \po2783 \textcolor{gray}{$\pm$ 710\po} & \po5987 \textcolor{gray}{$\pm$ 407\po} & \po7954 \textcolor{gray}{$\pm$ 248\po} & \po7196 \textcolor{gray}{$\pm$ 363\po} \\
$Q$, remove $s,a$ & 17042 \textcolor{gray}{$\pm$ 208\po} & \po2268 \textcolor{gray}{$\pm$ 723\po} & \po5486 \textcolor{gray}{$\pm$ 466\po} & \po8546 \textcolor{gray}{$\pm$ 655\po} & \po6449 \textcolor{gray}{$\pm$ 478\po} \\
$Q$, $z^{sa}$ only & 15410 \textcolor{gray}{$\pm$ 691\po} & \po2781 \textcolor{gray}{$\pm$ 585\po} & \po5205 \textcolor{gray}{$\pm$ 409\po} & \po7916 \textcolor{gray}{$\pm$ 976\po} & \po5551 \textcolor{gray}{$\pm$ 1144} \\
$Q$, $s,a$ only & 12776 \textcolor{gray}{$\pm$ 774\po} & \po2929 \textcolor{gray}{$\pm$ 631\po} & \po4938 \textcolor{gray}{$\pm$ 422\po} & \po5695 \textcolor{gray}{$\pm$ 875\po} & \po6209 \textcolor{gray}{$\pm$ 384\po} \\
$\pi$, $s$ only & 16997 \textcolor{gray}{$\pm$ 265\po} & \po3206 \textcolor{gray}{$\pm$ 419\po} & \po5744 \textcolor{gray}{$\pm$ 430\po} & \po7710 \textcolor{gray}{$\pm$ 593\po} & \po6413 \textcolor{gray}{$\pm$ 1521} \\
$\pi$, $z^s$ only & 17283 \textcolor{gray}{$\pm$ 388\po} & \po2429 \textcolor{gray}{$\pm$ 705\po} & \po6319 \textcolor{gray}{$\pm$ 331\po} & \po8378 \textcolor{gray}{$\pm$ 732\po} & \po6507 \textcolor{gray}{$\pm$ 496\po} \\
No fixed embeddings & 17116 \textcolor{gray}{$\pm$ 149\po} & \po2262 \textcolor{gray}{$\pm$ 590\po} & \po5835 \textcolor{gray}{$\pm$ 481\po} & \po8027 \textcolor{gray}{$\pm$ 551\po} & \po7334 \textcolor{gray}{$\pm$ 153\po} \\
\midrule
\multicolumn{6}{c}{Normalization (\autoref{appendix:subsec:normalization})} \\
\midrule
No normalization on $\phi$ & 17231 \textcolor{gray}{$\pm$ 246\po} & \po2647 \textcolor{gray}{$\pm$ 466\po} & \po5639 \textcolor{gray}{$\pm$ 1248} & \po8191 \textcolor{gray}{$\pm$ 846\po} & \po5862 \textcolor{gray}{$\pm$ 1471} \\
No normalization & 17275 \textcolor{gray}{$\pm$ 288\po} & \po3359 \textcolor{gray}{$\pm$ 479\po} & \po6168 \textcolor{gray}{$\pm$ 164\po} & \po7274 \textcolor{gray}{$\pm$ 662\po} & \po4803 \textcolor{gray}{$\pm$ 1706} \\
Normalization on $z^{sa}$ & 16947 \textcolor{gray}{$\pm$ 284\po} & \po3383 \textcolor{gray}{$\pm$ 262\po} & \po5502 \textcolor{gray}{$\pm$ 1142} & \po8049 \textcolor{gray}{$\pm$ 538\po} & \po6418 \textcolor{gray}{$\pm$ 302\po} \\
BatchNorm & 17299 \textcolor{gray}{$\pm$ 218\po} & \po2318 \textcolor{gray}{$\pm$ 473\po} & \po4839 \textcolor{gray}{$\pm$ 987\po} & \po7636 \textcolor{gray}{$\pm$ 769\po} & \po6979 \textcolor{gray}{$\pm$ 325\po} \\
LayerNorm & 17132 \textcolor{gray}{$\pm$ 360\po} & \po2451 \textcolor{gray}{$\pm$ 902\po} & \po4470 \textcolor{gray}{$\pm$ 1096} & \po6331 \textcolor{gray}{$\pm$ 1024} & \po6712 \textcolor{gray}{$\pm$ 1297} \\
Cosine similarity loss & 16897 \textcolor{gray}{$\pm$ 372\po} & \po3324 \textcolor{gray}{$\pm$ 249\po} & \po5566 \textcolor{gray}{$\pm$ 353\po} & \po7873 \textcolor{gray}{$\pm$ 511\po} & \po4370 \textcolor{gray}{$\pm$ 1573} \\
\midrule
\multicolumn{6}{c}{End-to-end (\autoref{appendix:subsec:endtoend})} \\
\midrule
End-to-end, $0.1$ & 16186 \textcolor{gray}{$\pm$ 360\po} & \po1820 \textcolor{gray}{$\pm$ 674\po} & \po5013 \textcolor{gray}{$\pm$ 729\po} & \po6601 \textcolor{gray}{$\pm$ 1251} & \po5076 \textcolor{gray}{$\pm$ 897\po} \\
End-to-end, $1$ & 15775 \textcolor{gray}{$\pm$ 658\po} & \po1779 \textcolor{gray}{$\pm$ 537\po} & \po4882 \textcolor{gray}{$\pm$ 710\po} & \po6604 \textcolor{gray}{$\pm$ 1135} & \po5880 \textcolor{gray}{$\pm$ 703\po} \\
End-to-end, $10$ & 16472 \textcolor{gray}{$\pm$ 365\po} & \po1534 \textcolor{gray}{$\pm$ 341\po} & \po4900 \textcolor{gray}{$\pm$ 760\po} & \po6626 \textcolor{gray}{$\pm$ 1253} & \po5279 \textcolor{gray}{$\pm$ 1095} \\
\bottomrule
\end{tabularx}
\end{table*}

\clearpage

\subsection{Learning Target} \label{appendix:subsec:learning_target}

The encoder~$(f_{t+1}, g_{t+1})$ in TD7 uses the following update:
\begin{align} %
\Loss(f_{t+1},g_{t+1}) :=&~\Bigl( g_{t+1}(f_{t+1}(s),a) - |f_{t+1}(s')|_\times \Bigr)^2 \\
=&~\lp z_{t+1}^{sa} - |z_{t+1}^{s'}|_\times \rp^2.
\end{align}
In this section, we vary the learning target (originally $z_{t+1}^{s'}$).

(\autoref{appendix:fig:design:baseline_target}) \textbf{Target 1:} $s'$. Sets the learning target to the next state $s'$, making the encoder loss:
\begin{align} %
\Loss(f_{t+1},g_{t+1}) := \lp z_{t+1}^{sa} - s' \rp^2.
\end{align}
This target is inspired by OFENet~\citep{ota2020can}.%

(\autoref{appendix:fig:design:baseline_target}) \textbf{Target 2:} $z^{s'}_\text{target}$. Sets the learning target to the embedding of the next state~$z^{s'}_\text{target}$ taken from a slow-moving target network. This makes the encoder loss:
\begin{align} %
\Loss(f_{t+1},g_{t+1}) := \lp z_{t+1}^{sa} - |z_{t+1}^{sa}|_\times \rp^2.
\end{align}
$z^{s'}_\text{target}$ is taken from a target network $f_\text{target}$ where $f_{t+1}$ has parameters $\ta$ and $f_\text{target}$ has parameters $\ta'$, and $\ta'$ is updated at each time step by:
\begin{equation}
    \ta' \leftarrow (0.99) \ta' + (0.01) \ta.
\end{equation}
This target is inspired by SPR~\citep{schwarzer2020data}.%

(\autoref{appendix:fig:design:baseline_target}) \textbf{Target 3:} $z^s$ and $r$. Sets the learning target to the embedding of the next state~$z^s_{t+1}$ (which is the same as in the original loss) but adds a second loss towards the target~$r$. This is achieved by having the output of $g$ be two dimensional, with the first dimension being $z_{t+1}^{sa}$ and the second $r^\text{pred}_{t+1}$, making the encoder loss:
\begin{align} %
\Loss(f_{t+1},g_{t+1}) :=&~\Bigl( g_{t+1}(f_{t+1}(s),a)[0] - |f_{t+1}(s')|_\times \Bigr)^2 + \Bigl( g_{t+1}(f_{t+1}(s),a)[1] - r \Bigr)^2\\
=&~\lp z_{t+1}^{sa} - |z_{t+1}^{s'}|_\times \rp^2 + \lp r^\text{pred}_{t+1} - r \rp^2. 
\end{align}
This target is inspired by DeepMDP~\citep{gelada2019deepmdp}.%

(\autoref{appendix:fig:design:alternate_target}) \textbf{Target 4:} $z^{s'a'}$. Sets the learning target to the next state-action embedding $z_{t+1}^{s'a'}$, where the action $a'$ is sampled according to the target policy with added noise and clipping (as done in TD3~\citep{fujimoto2018addressing}). This makes the encoder loss:
\begin{align} %
\Loss(f_{t+1},g_{t+1}) :=&~\lp z_{t+1}^{sa} - |z_{t+1}^{s'a'}|_\times \rp^2, \\
a' :=&~\pi_{t}(z^{s'}_{t-1}, s') + \e, \\
\e \sim&~\text{clip}(\N(0,\sigma^2), -c, c). 
\end{align}

\clearpage

\begin{table*}[ht]
	\centering
	\small
	\setlength{\tabcolsep}{4pt}
	\newcolumntype{Y}{>{\centering\arraybackslash}X} %
	\caption{Average performance on the MuJoCo benchmark at 1M time steps. $\pm$~captures a $95\%$ confidence interval around the average performance. Results are over $10$ seeds.} %
	\begin{tabularx}{\textwidth}{lYYYYY}
		\toprule
		Algorithm & HalfCheetah & Hopper & Walker2d & Ant & Humanoid \\ 
		\midrule
		TD7 (no checkpoints) & 17123 \textcolor{gray}{$\pm$ 296\po} & \po3361 \textcolor{gray}{$\pm$ 429\po} & \po5718 \textcolor{gray}{$\pm$ 308\po} & \po8605 \textcolor{gray}{$\pm$ 1008} & \po7381 \textcolor{gray}{$\pm$ 172\po} \\
		TD3 & 10574 \textcolor{gray}{$\pm$ 897\po} & \po3226 \textcolor{gray}{$\pm$ 315\po} & \po3946 \textcolor{gray}{$\pm$ 292\po} & \po3942 \textcolor{gray}{$\pm$ 1030} & \po5165 \textcolor{gray}{$\pm$ 145\po} \\
		\midrule
		$s'$ & 16438 \textcolor{gray}{$\pm$ 219\po} & \po3507 \textcolor{gray}{$\pm$ 313\po} & \po5882 \textcolor{gray}{$\pm$ 262\po} & \po7415 \textcolor{gray}{$\pm$ 753\po} & \po6337 \textcolor{gray}{$\pm$ 387\po} \\
		$z^{s'}_\text{target}$ & 17317 \textcolor{gray}{$\pm$ 65\po\po} & \po3034 \textcolor{gray}{$\pm$ 647\po} & \po6227 \textcolor{gray}{$\pm$ 218\po} & \po7939 \textcolor{gray}{$\pm$ 334\po} & \po6984 \textcolor{gray}{$\pm$ 578\po} \\
		$z^s$ and $r$ & 17089 \textcolor{gray}{$\pm$ 219\po} & \po3264 \textcolor{gray}{$\pm$ 670\po} & \po6255 \textcolor{gray}{$\pm$ 220\po} & \po8300 \textcolor{gray}{$\pm$ 675\po} & \po6841 \textcolor{gray}{$\pm$ 252\po} \\
		$z^{s'a'}$ & 12445 \textcolor{gray}{$\pm$ 417\po} & \po3009 \textcolor{gray}{$\pm$ 528\po} & \po4757 \textcolor{gray}{$\pm$ 352\po} & \po6288 \textcolor{gray}{$\pm$ 570\po} & \po5657 \textcolor{gray}{$\pm$ 110\po} \\
		\bottomrule
	\end{tabularx}
\end{table*}

\begin{figure}[ht]
\appendixfigure{design_0}
\fcolorbox{gray}{gray!10}{
\small
\cblock{sb_blue}~TD7 \quad \cblock{sb_orange}~TD3 \quad \cblock{sb_green}~$s'$ \quad \cblock{sb_red}~$z^{s'}_\text{target}$ \quad \cblock{sb_purple} $z^{s'}$ and $r$ 
} 
\caption{\textbf{Learning targets from baseline methods.} Learning curves on the MuJoCo benchmark, using learning targets inspired by baseline methods, the next state~$s'$ from OFENet~\citep{ota2020can}, the next state embedding~$z^{s'}_\text{target}$ from a slow-moving target network from SPR~\citep{schwarzer2020data}, and including the reward term ($z^{s'}$ and $r$) as in DeepMDP~\citep{gelada2019deepmdp}. Note that this study only accounts for changes to the learning target and does not encompass all aspects of the baseline algorithms. Results are averaged over $10$ seeds. The shaded area captures a $95\%$ confidence interval around the average performance.} \label{appendix:fig:design:baseline_target}
\end{figure}

\begin{figure}[ht]
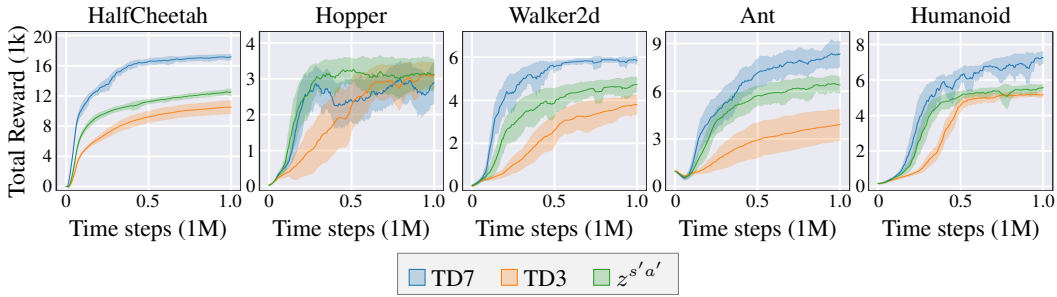

\appendixfigure{design_1}
\fcolorbox{gray}{gray!10}{
\small
\cblock{sb_blue}~TD7 \quad \cblock{sb_orange}~TD3 \quad \cblock{sb_green}~$z^{s'a'}$
} 
\caption{\textbf{Alternate learning target.} Learning curves on the MuJoCo benchmark, using a learning target of the next state-action embedding~$z^{s'a'}$. Results are averaged over $10$ seeds. The shaded area captures a $95\%$ confidence interval around the average performance.} \label{appendix:fig:design:alternate_target}
\end{figure}

\clearpage

\subsection{Network Input} \label{appendix:subsec:network_input}

SALE modifies the value function and policy to take the embeddings as additional input:
\begin{align}
    Q_{t+1}(s,a) \rightarrow Q_{t+1}(z^{sa}_t, z^s_t, s, a), \qquad \pi_{t+1}(s) \rightarrow \pi_{t+1}(z^s, s).
\end{align}
In this section, we vary the input of the value function and policy. In all cases, changes are made independently (i.e.\ we fix the policy to the above (using embeddings) and change the policy, or fix the value function to the above (using embeddings) and change the policy). 

(\autoref{appendix:fig:design:input_remove}) \textbf{Input 1:} $Q$, remove $z^{sa}$. We remove $z^{sa}$ from the value function input. This makes the input: $Q_{t+1}(z^s_t, s, a)$. 

(\autoref{appendix:fig:design:input_remove}) \textbf{Input 2:} $Q$, remove $z^{s}$. We remove $z^{s}$ from the value function input. This makes the input: $Q_{t+1}(z^{sa}_t, s, a)$. 

(\autoref{appendix:fig:design:input_remove}) \textbf{Input 3:} $Q$, remove $s,a$. We remove $s,a$ from the value function input. This makes the input: $Q_{t+1}(z^{sa}_t, z^s)$. 

(\autoref{appendix:fig:design:input_exclusive}) \textbf{Input 4:} $Q$, $z^{sa}$ only. We only use $z^{sa}$ in the value function input. This makes the input: $Q_{t+1}(z^{sa}_t)$. 

(\autoref{appendix:fig:design:input_exclusive}) \textbf{Input 5:} $Q$, $s,a$ only. We only use $s,a$ in the value function input. This makes the input: $Q_{t+1}(s,a)$. As for all cases, the policy still takes in the state embedding~$z^s$. 

(\autoref{appendix:fig:design:input_policy}) \textbf{Input 6:} $\pi$, $s$ only. We remove $z^s$ from the policy input, using only the state~$s$ as input. This makes the input: $\pi_{t+1}(s)$. Note that the value function still uses the embeddings as input.

(\autoref{appendix:fig:design:input_policy}) \textbf{Input 7:} $\pi$, $z^s$ only. We remove $s$ from the policy input, using only the embedding~$z^s$ as input. This makes the input: $\pi_{t+1}(z_t^s)$. 

(\autoref{appendix:fig:design:fixed}) \textbf{Input 8:} No fixed embeddings. We remove the fixed embeddings from the value function and the policy. This means that the networks use embeddings from the current encoder, rather than the fixed encoder from the previous iteration. This makes the input: $Q_{t+1}(z_{t+1}^{sa}, z^s_{t+1}, s, a)$ and $\pi_{t+1}(z^s_{t+1}, s)$. 

\begin{table*}[ht]
	\centering
	\small
	\setlength{\tabcolsep}{4pt}
	\newcolumntype{Y}{>{\centering\arraybackslash}X} %
	\caption{Average performance on the MuJoCo benchmark at 1M time steps. $\pm$~captures a $95\%$ confidence interval around the average performance. Results are over $10$ seeds.} %
	\begin{tabularx}{\textwidth}{lYYYYY}
		\toprule
		Algorithm & HalfCheetah & Hopper & Walker2d & Ant & Humanoid \\ 
		\midrule
		TD7 (no checkpoints) & 17123 \textcolor{gray}{$\pm$ 296\po} & \po3361 \textcolor{gray}{$\pm$ 429\po} & \po5718 \textcolor{gray}{$\pm$ 308\po} & \po8605 \textcolor{gray}{$\pm$ 1008} & \po7381 \textcolor{gray}{$\pm$ 172\po} \\
		TD3 & 10574 \textcolor{gray}{$\pm$ 897\po} & \po3226 \textcolor{gray}{$\pm$ 315\po} & \po3946 \textcolor{gray}{$\pm$ 292\po} & \po3942 \textcolor{gray}{$\pm$ 1030} & \po5165 \textcolor{gray}{$\pm$ 145\po} \\
		\midrule
		$Q$, remove $z^{sa}$ & 16948 \textcolor{gray}{$\pm$ 250\po} & \po2914 \textcolor{gray}{$\pm$ 756\po} & \po5746 \textcolor{gray}{$\pm$ 254\po} & \po6911 \textcolor{gray}{$\pm$ 552\po} & \po6907 \textcolor{gray}{$\pm$ 443\po} \\
		$Q$, remove $z^{s}$ & 16856 \textcolor{gray}{$\pm$ 241\po} & \po2783 \textcolor{gray}{$\pm$ 710\po} & \po5987 \textcolor{gray}{$\pm$ 407\po} & \po7954 \textcolor{gray}{$\pm$ 248\po} & \po7196 \textcolor{gray}{$\pm$ 363\po} \\
		$Q$, remove $s,a$ & 17042 \textcolor{gray}{$\pm$ 208\po} & \po2268 \textcolor{gray}{$\pm$ 723\po} & \po5486 \textcolor{gray}{$\pm$ 466\po} & \po8546 \textcolor{gray}{$\pm$ 655\po} & \po6449 \textcolor{gray}{$\pm$ 478\po} \\
		$Q$, $z^{sa}$ only & 15410 \textcolor{gray}{$\pm$ 691\po} & \po2781 \textcolor{gray}{$\pm$ 585\po} & \po5205 \textcolor{gray}{$\pm$ 409\po} & \po7916 \textcolor{gray}{$\pm$ 976\po} & \po5551 \textcolor{gray}{$\pm$ 1144} \\
		$Q$, $s,a$ only & 12776 \textcolor{gray}{$\pm$ 774\po} & \po2929 \textcolor{gray}{$\pm$ 631\po} & \po4938 \textcolor{gray}{$\pm$ 422\po} & \po5695 \textcolor{gray}{$\pm$ 875\po} & \po6209 \textcolor{gray}{$\pm$ 384\po} \\
		$\pi$, $s$ only & 16997 \textcolor{gray}{$\pm$ 265\po} & \po3206 \textcolor{gray}{$\pm$ 419\po} & \po5744 \textcolor{gray}{$\pm$ 430\po} & \po7710 \textcolor{gray}{$\pm$ 593\po} & \po6413 \textcolor{gray}{$\pm$ 1521} \\
		$\pi$, $z^s$ only & 17283 \textcolor{gray}{$\pm$ 388\po} & \po2429 \textcolor{gray}{$\pm$ 705\po} & \po6319 \textcolor{gray}{$\pm$ 331\po} & \po8378 \textcolor{gray}{$\pm$ 732\po} & \po6507 \textcolor{gray}{$\pm$ 496\po} \\
		No fixed embeddings & 17116 \textcolor{gray}{$\pm$ 149\po} & \po2262 \textcolor{gray}{$\pm$ 590\po} & \po5835 \textcolor{gray}{$\pm$ 481\po} & \po8027 \textcolor{gray}{$\pm$ 551\po} & \po7334 \textcolor{gray}{$\pm$ 153\po} \\
		\bottomrule
	\end{tabularx}
\end{table*}

\begin{figure}[ht]
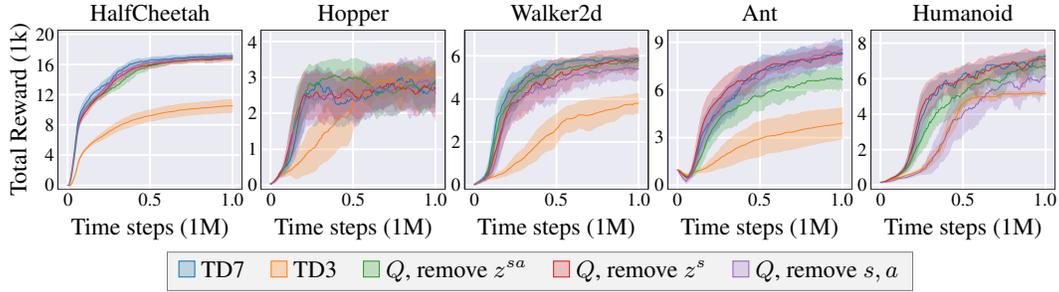

\appendixfigure{design_2}
\fcolorbox{gray}{gray!10}{
\small
\cblock{sb_blue}~TD7 \quad \cblock{sb_orange}~TD3 \quad \cblock{sb_green}~$Q$, remove $z^{sa}$ \quad \cblock{sb_red}~$Q$, remove $z^{s}$ \quad \cblock{sb_purple}~$Q$, remove $s,a$
} 
\caption{\textbf{Removing terms from the value function input.} Learning curves on the MuJoCo benchmark, when removing components from the input to the value function. Results are averaged over $10$ seeds. The shaded area captures a $95\%$ confidence interval around the average performance.} \label{appendix:fig:design:input_remove}
\end{figure}
\vspace{-64pt}
\begin{figure}[ht]
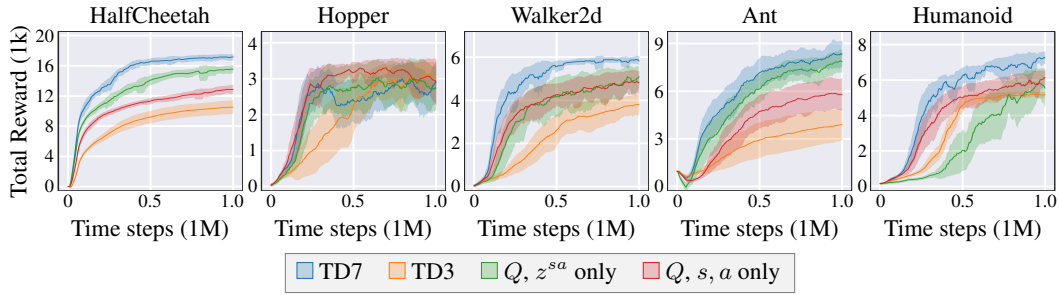

\appendixfigure{design_3}
\fcolorbox{gray}{gray!10}{
\small
\cblock{sb_blue}~TD7 \quad \cblock{sb_orange}~TD3 \quad \cblock{sb_green}~$Q$, $z^{sa}$ only \quad \cblock{sb_red}~$Q$, $s,a$ only
} 
\caption{\textbf{Exclusive value function input.} Learning curves on the MuJoCo benchmark, changing the value function input to a single component. Results are averaged over $10$ seeds. The shaded area captures a $95\%$ confidence interval around the average performance.} \label{appendix:fig:design:input_exclusive}
\end{figure}
\vspace{-64pt}
\begin{figure}[ht]
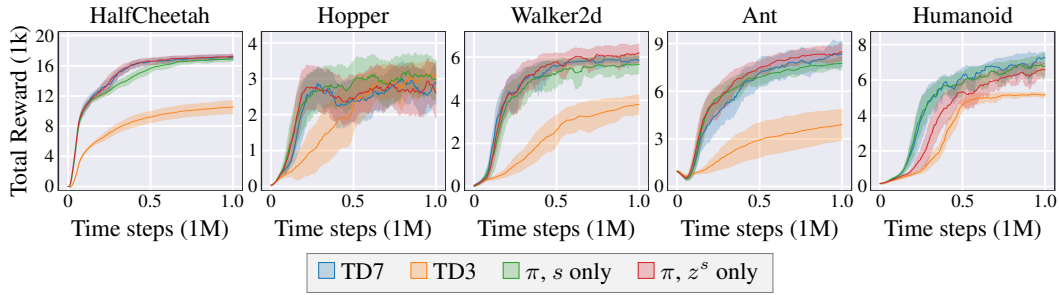

\appendixfigure{design_4}
\fcolorbox{gray}{gray!10}{
\small
\cblock{sb_blue}~TD7 \quad \cblock{sb_orange}~TD3 \quad \cblock{sb_green}~$\pi$, $s$ only \quad \cblock{sb_red}~$\pi$, $z^s$ only
} 
\caption{\textbf{Modified policy input.} Learning curves on the MuJoCo benchmark, changing the policy input to a single component. Results are averaged over $10$ seeds. The shaded area captures a $95\%$ confidence interval around the average performance.} \label{appendix:fig:design:input_policy}
\end{figure}
\vspace{-64pt}
\begin{figure}[ht]
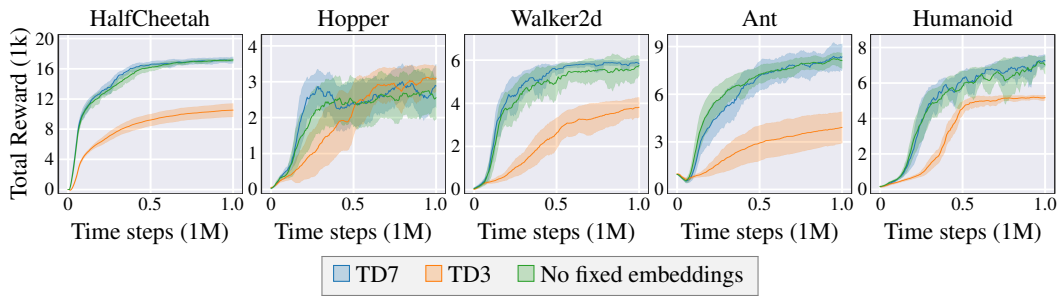

\appendixfigure{design_5}
\fcolorbox{gray}{gray!10}{
\small
\cblock{sb_blue}~TD7 \quad \cblock{sb_orange}~TD3 \quad \cblock{sb_green}~No fixed embeddings
} 
\caption{\textbf{Fixed embeddings.} Learning curves on the MuJoCo benchmark, without fixed embeddings for the value function and policy, so that the current encoder is used rather than the fixed encoder. Results are averaged over $10$ seeds. The shaded area captures a $95\%$ confidence interval around the average performance.} \label{appendix:fig:design:fixed}
\end{figure}

\clearpage

\subsection{Normalization} \label{appendix:subsec:normalization}

SALE normalizes embeddings and input through our proposed AvgL1Norm:
\begin{equation}
    \text{AvgL1Norm}(x) := \frac{x}{\frac{1}{N} \sum_i |x_i|}.
\end{equation}

AvgL1Norm is embedded into the architecture used by SALE (see \autoref{pseudo:TD7}), applied to the output of the state encoder~$f$ to give the state embedding~$z^s = \text{AvgL1Norm}(f(s))$. It is also used following a linear layer in the input of both the value function and policy. This makes the input to the value function and policy: 
\begin{align} 
    &Q(z^{sa}, \text{AvgL1Norm}(f(s)), \text{AvgL1Norm}(\phi^{sa})), \\
    &\pi(\text{AvgL1Norm}(f(s)), \text{AvgL1Norm}(\phi^s)),
\end{align}
where $\phi^{sa}$ is the output from a linear layer on the state-action input, $\text{Linear}(s,a)$, and $\phi^s$ is the output from a linear layer on the state input, $\text{Linear}(s)$. 

In this section, we vary how normalization is used, swapping which components it is used on and what normalization function is used. 

(\autoref{appendix:fig:design:remove_norm}) \textbf{Normalization 1:} No normalization on $\phi$. Normalization is not applied to either $\phi$. This makes the input to the value function and policy: 
\begin{align} 
    &Q(z^{sa}, \text{AvgL1Norm}(f(s)), \phi^{sa}), \\
    &\pi(\text{AvgL1Norm}(f(s)), \phi^s).
\end{align}

(\autoref{appendix:fig:design:remove_norm}) \textbf{Normalization 2:} No normalization. Normalization is not applied anywhere. This makes the input to the value function and policy: 
\begin{align} 
    &Q(z^{sa}, f(s), \phi^{sa}), \\
    &\pi(f(s), \phi^s).
\end{align}

(\autoref{appendix:fig:design:zsa_norm}) \textbf{Normalization 3:} Normalization used on $z^{sa}$. Since $z^{sa}$ is trained to predict the next state embedding~$z^{s'}$, which is normalized, normalization may not be needed for the output of the state-action encoder $g$. We try also applying normalization to the output of $g$. This makes the input to the value function and policy: 
\begin{align} 
    &Q(\text{AvgL1Norm}(g(z^s, a)), \text{AvgL1Norm}(f(s)), \text{AvgL1Norm}(\phi^{sa})), \\
    &\pi(\text{AvgL1Norm}(f(s)), \text{AvgL1Norm}(\phi^{s})).
\end{align}

(\autoref{appendix:fig:design:alternate_norm}) \textbf{Normalization 4:} BatchNorm. BatchNorm is used instead of AvgL1Norm. This makes the input to the value function and policy: 
\begin{align} 
    &Q(z^{sa}, \text{BatchNorm}(f(s)), \text{BatchNorm}(\phi^{sa})), \\
    &\pi(\text{BatchNorm}(f(s)), \text{BatchNorm}(\phi^s)),
\end{align}

(\autoref{appendix:fig:design:alternate_norm}) \textbf{Normalization 5:} LayerNorm. LayerNorm is used instead of AvgL1Norm. This makes the input to the value function and policy: 
\begin{align} 
    &Q(z^{sa}, \text{LayerNorm}(f(s)), \text{LayerNorm}(\phi^{sa})), \\
    &\pi(\text{LayerNorm}(f(s)), \text{LayerNorm}(\phi^s)),
\end{align}

(\autoref{appendix:fig:design:SPR_norm}) \textbf{Normalization 3:} Cosine similarity loss. This is inspired by SPR~\citep{schwarzer2020data}, where normalization is not used and a cosine similarity loss is used instead of MSE for updating the encoder. This means the encoder loss function, which is originally:
\begin{align} %
\Loss(f,g) :=&~\Bigl( g(f(s),a) - |f(s')|_\times \Bigr)^2 = \lp z^{sa} - |z^{s'}|_\times \rp^2,
\end{align}
now becomes
\begin{align} %
\Loss(f,g) &:= \text{Cosine} \lp z^{sa}, |z^{s'}|_\times \rp, \\
\text{Cosine}\lp z^{sa}, |z^{s'}|_\times \rp &:= \lp \frac{z^{sa}}{\lv z^{sa} \rv_2}  \rp^\top \lp \frac{|z^{s'}|_\times}{\lv |z^{s'}|_\times \rv_2} \rp.
\end{align}
No normalization is used elsewhere. 

\clearpage

\begin{table*}[ht]
	\centering
	\small
	\setlength{\tabcolsep}{4pt}
	\newcolumntype{Y}{>{\centering\arraybackslash}X} %
	\caption{Average performance on the MuJoCo benchmark at 1M time steps. $\pm$~captures a $95\%$ confidence interval around the average performance. Results are over $10$ seeds.} %
	\begin{tabularx}{\textwidth}{lYYYYY}
		\toprule
		Algorithm & HalfCheetah & Hopper & Walker2d & Ant & Humanoid \\ 
		\midrule
		TD7 (no checkpoints) & 17123 \textcolor{gray}{$\pm$ 296\po} & \po3361 \textcolor{gray}{$\pm$ 429\po} & \po5718 \textcolor{gray}{$\pm$ 308\po} & \po8605 \textcolor{gray}{$\pm$ 1008} & \po7381 \textcolor{gray}{$\pm$ 172\po} \\
		TD3 & 10574 \textcolor{gray}{$\pm$ 897\po} & \po3226 \textcolor{gray}{$\pm$ 315\po} & \po3946 \textcolor{gray}{$\pm$ 292\po} & \po3942 \textcolor{gray}{$\pm$ 1030} & \po5165 \textcolor{gray}{$\pm$ 145\po} \\
		\midrule
		No normalization on $\phi$ & 17231 \textcolor{gray}{$\pm$ 246\po} & \po2647 \textcolor{gray}{$\pm$ 466\po} & \po5639 \textcolor{gray}{$\pm$ 1248} & \po8191 \textcolor{gray}{$\pm$ 846\po} & \po5862 \textcolor{gray}{$\pm$ 1471} \\
		No normalization & 17275 \textcolor{gray}{$\pm$ 288\po} & \po3359 \textcolor{gray}{$\pm$ 479\po} & \po6168 \textcolor{gray}{$\pm$ 164\po} & \po7274 \textcolor{gray}{$\pm$ 662\po} & \po4803 \textcolor{gray}{$\pm$ 1706} \\
		Normalization on $z^{sa}$ & 16947 \textcolor{gray}{$\pm$ 284\po} & \po3383 \textcolor{gray}{$\pm$ 262\po} & \po5502 \textcolor{gray}{$\pm$ 1142} & \po8049 \textcolor{gray}{$\pm$ 538\po} & \po6418 \textcolor{gray}{$\pm$ 302\po} \\
		BatchNorm & 17299 \textcolor{gray}{$\pm$ 218\po} & \po2318 \textcolor{gray}{$\pm$ 473\po} & \po4839 \textcolor{gray}{$\pm$ 987\po} & \po7636 \textcolor{gray}{$\pm$ 769\po} & \po6979 \textcolor{gray}{$\pm$ 325\po} \\
		LayerNorm & 17132 \textcolor{gray}{$\pm$ 360\po} & \po2451 \textcolor{gray}{$\pm$ 902\po} & \po4470 \textcolor{gray}{$\pm$ 1096} & \po6331 \textcolor{gray}{$\pm$ 1024} & \po6712 \textcolor{gray}{$\pm$ 1297} \\
		Cosine similarity loss & 16897 \textcolor{gray}{$\pm$ 372\po} & \po3324 \textcolor{gray}{$\pm$ 249\po} & \po5566 \textcolor{gray}{$\pm$ 353\po} & \po7873 \textcolor{gray}{$\pm$ 511\po} & \po4370 \textcolor{gray}{$\pm$ 1573} \\
		\bottomrule
	\end{tabularx}
\end{table*}

\begin{figure}[ht]
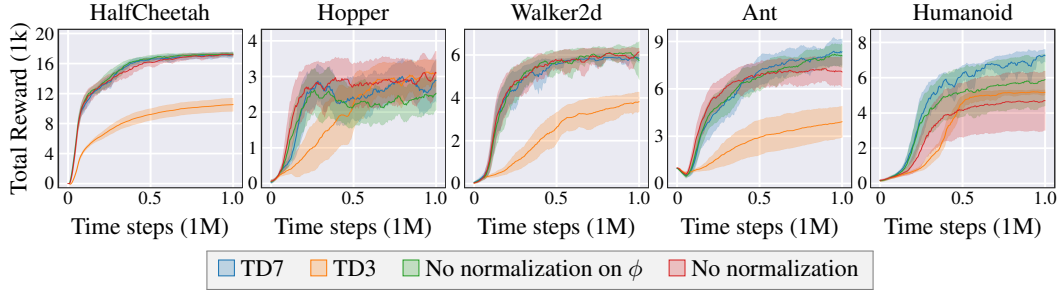

\appendixfigure{design_6}
\fcolorbox{gray}{gray!10}{
\small
\cblock{sb_blue}~TD7 \quad \cblock{sb_orange}~TD3 \quad \cblock{sb_green}~No normalization on $\phi$ \quad \cblock{sb_red}~No normalization
} 
\caption{\textbf{Removing normalization.} Learning curves on the MuJoCo benchmark, where normalization is removed on certain components of the input to the value function and policy. Results are averaged over $10$ seeds. The shaded area captures a $95\%$ confidence interval around the average performance.} \label{appendix:fig:design:remove_norm}
\end{figure}

\begin{figure}[ht]
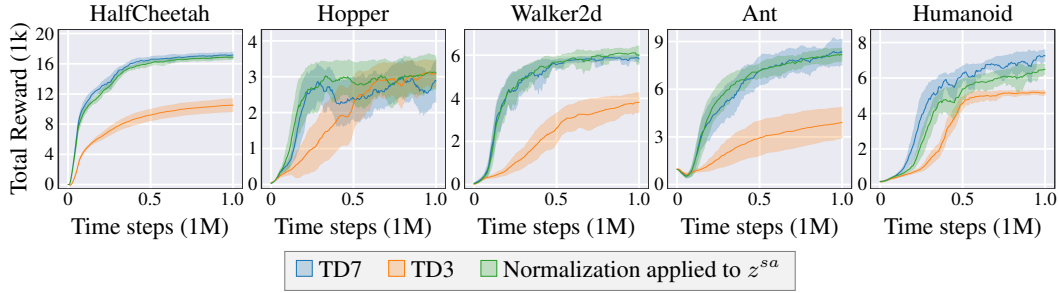

\appendixfigure{design_7}
\fcolorbox{gray}{gray!10}{
\small
\cblock{sb_blue}~TD7 \quad \cblock{sb_orange}~TD3 \quad \cblock{sb_green}~Normalization applied to $z^{sa}$
} 
\caption{\textbf{Normalization on $z^{sa}$.} Learning curves on the MuJoCo benchmark, where normalization is also applied to the state-action embedding~$z^{sa}$. Results are averaged over $10$ seeds. The shaded area captures a $95\%$ confidence interval around the average performance.} \label{appendix:fig:design:zsa_norm}
\end{figure}

\clearpage

\begin{figure}[ht]
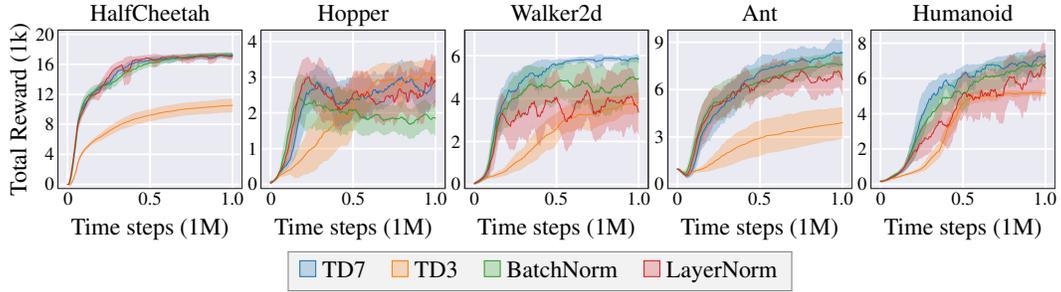

\appendixfigure{design_8}
\fcolorbox{gray}{gray!10}{
\small
\cblock{sb_blue}~TD7 \quad \cblock{sb_orange}~TD3 \quad \cblock{sb_green}~BatchNorm \quad \cblock{sb_red}~LayerNorm
} 
\caption{\textbf{Alternate normalization.} Learning curves on the MuJoCo benchmark, where BatchNorm or LayerNorm is used instead of AvgL1Norm. Results are averaged over $10$ seeds. The shaded area captures a $95\%$ confidence interval around the average performance.} \label{appendix:fig:design:alternate_norm}
\end{figure}

\begin{figure}[ht]
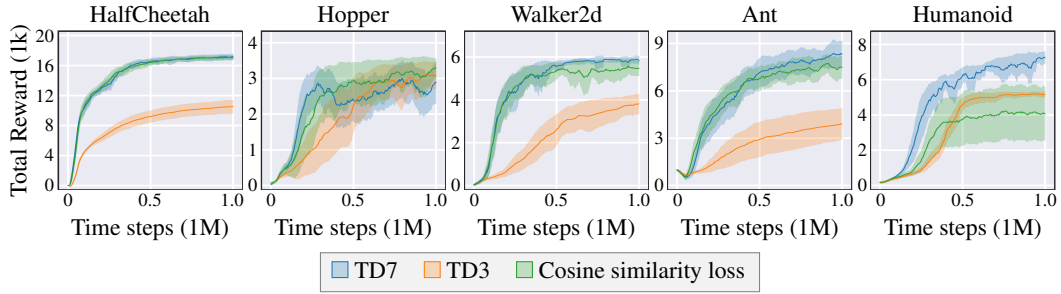

\appendixfigure{design_9}
\fcolorbox{gray}{gray!10}{
\small
\cblock{sb_blue}~TD7 \quad \cblock{sb_orange}~TD3 \quad \cblock{sb_green}~Cosine similarity loss
} 
\caption{\textbf{Cosine similarity instead of normalization.} Learning curves on the MuJoCo benchmark, where normalization is removed and replaced with a cosine similarity loss, as inspired by SPR~\citep{schwarzer2020data}. Results are averaged over $10$ seeds. The shaded area captures a $95\%$ confidence interval around the average performance.} \label{appendix:fig:design:SPR_norm}
\end{figure}

\clearpage

\subsection{End-to-end} \label{appendix:subsec:endtoend}

The embeddings used in SALE are decoupled, meaning that the encoders used to output the embeddings are trained independently from the value function or policy. Instead, we might consider training the encoders end-to-end with the value function. Doing so requires training the current embedding with the value function loss (rather than using the fixed embedding). This means that the encoders are trained with the following loss: 
\begin{align} 
\Loss(f_{t+1}, g_{t+1}) :=&~\text{Huber}\Bigl( \texttt{target} - Q_{t+1}(z_{t+1}^{sa}, z_{t+1}^s, s, a) \Bigr) + \beta \lp z_{t+1}^{sa} - |z_{t+1}^{s'}|_\times \rp^2,\\
\texttt{target} :=&~r + \y \text{ clip} \bigl( \min \lp Q_{t,1} (x), Q_{t,2} (x) \rp, Q_\text{min}, Q_\text{max} \bigr), \\
x :=&~[ z_{t}^{s'a'}, z_{t}^{s'}, s', a' ],\\
a' :=&~\pi_{t}(z^{s'}_{t}, s') + \e, \\
\e \sim&~\text{clip}(\N(0,\sigma^2), -c, c). 
\end{align}
In \autoref{appendix:table:endtoend} and \autoref{appendix:fig:design:endtoend} we display the performance when varying the hyperparameter~$\beta$.

\begin{table*}[ht]
	\centering
	\small
	\setlength{\tabcolsep}{4pt}
	\newcolumntype{Y}{>{\centering\arraybackslash}X} %
	\caption{Average performance on the MuJoCo benchmark at 1M time steps. $\pm$~captures a $95\%$ confidence interval around the average performance. Results are over $10$ seeds.} \label{appendix:table:endtoend}
	\begin{tabularx}{\textwidth}{lYYYYY}
		\toprule
		Algorithm & HalfCheetah & Hopper & Walker2d & Ant & Humanoid \\ 
		\midrule
		TD7 (no checkpoints) & 17123 \textcolor{gray}{$\pm$ 296\po} & \po3361 \textcolor{gray}{$\pm$ 429\po} & \po5718 \textcolor{gray}{$\pm$ 308\po} & \po8605 \textcolor{gray}{$\pm$ 1008} & \po7381 \textcolor{gray}{$\pm$ 172\po} \\
		TD3 & 10574 \textcolor{gray}{$\pm$ 897\po} & \po3226 \textcolor{gray}{$\pm$ 315\po} & \po3946 \textcolor{gray}{$\pm$ 292\po} & \po3942 \textcolor{gray}{$\pm$ 1030} & \po5165 \textcolor{gray}{$\pm$ 145\po} \\
		\midrule
		End-to-end, $0.1$ & 16186 \textcolor{gray}{$\pm$ 360\po} & \po1820 \textcolor{gray}{$\pm$ 674\po} & \po5013 \textcolor{gray}{$\pm$ 729\po} & \po6601 \textcolor{gray}{$\pm$ 1251} & \po5076 \textcolor{gray}{$\pm$ 897\po} \\
		End-to-end, $1$ & 15775 \textcolor{gray}{$\pm$ 658\po} & \po1779 \textcolor{gray}{$\pm$ 537\po} & \po4882 \textcolor{gray}{$\pm$ 710\po} & \po6604 \textcolor{gray}{$\pm$ 1135} & \po5880 \textcolor{gray}{$\pm$ 703\po} \\
		End-to-end, $10$ & 16472 \textcolor{gray}{$\pm$ 365\po} & \po1534 \textcolor{gray}{$\pm$ 341\po} & \po4900 \textcolor{gray}{$\pm$ 760\po} & \po6626 \textcolor{gray}{$\pm$ 1253} & \po5279 \textcolor{gray}{$\pm$ 1095} \\
		\bottomrule
	\end{tabularx}
\end{table*}

\begin{figure}[ht]
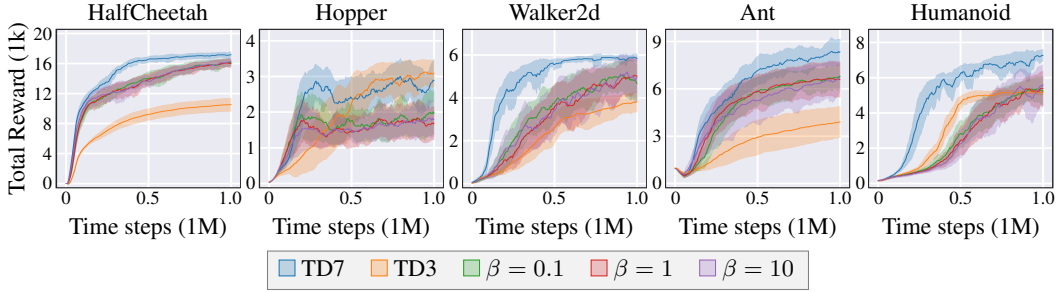

\appendixfigure{design_10}
\fcolorbox{gray}{gray!10}{
\small
\cblock{sb_blue}~TD7 \quad \cblock{sb_orange}~TD3 \quad \cblock{sb_green}~$\beta=0.1$ \quad \cblock{sb_red}~$\beta=1$ \quad \cblock{sb_purple}~$\beta=10$
} 
\caption{\textbf{End-to-end learning.} Learning curves on the MuJoCo benchmark, when training the embeddings end-to-end with the value function, where the encoder loss is weighted with respect to $\beta$. Results are averaged over $10$ seeds. The shaded area captures a $95\%$ confidence interval around the average performance.} \label{appendix:fig:design:endtoend}
\end{figure}

\clearpage

\section{Extrapolation Error Study}\label{appendix:sec:extrapolation}

\subsection{Extrapolation Error}

Extrapolation error is an issue with deep RL where the value function poorly estimates the value of unseen or rare actions~\citep{fujimoto2019off}. If we assume that $(s,a)$ is not contained in the dataset, then $Q(s,a)$ is effectively a guess generated by the value function~$Q$. During training, $s$ is always sampled from the dataset, but $a$ is often generated by the policy~($a \sim \pi(s))$, resulting in a potentially poor estimate.

Extrapolation error is typically considered a problem for offline RL, where the agent is given a fixed dataset and cannot interact further with the environment, as actions sampled from the policy may not be contained in the dataset. Extrapolation error is not considered a problem in online RL since the policy interacts with the environment, collecting data for the corresponding actions it generates. %

In our empirical analysis (\autoref{appendix:subsec:extrapolation_experiments}) we observe the presence of extrapolation error when using SALE. Our hypothesis is that by significantly expanding the input dimension dependent on the action, the network becomes more prone to erroneous extrapolations (for example, for the Ant environment, the original action dimension size is $8$, but the state-action embedding $z^{sa}$ has a dimension size of $256$). This is because unseen actions can appear significantly more distinct from seen actions, due to the dramatic increase in the number of dimensions used to represent the action-dependent input.

\subsection{Empirical Analysis} \label{appendix:subsec:extrapolation_experiments}

In this section we vary the input to the value function in TD7 to understand the role of the input dimension size and extrapolation error. The default input to the value function in TD7 is as follows:
\begin{align}
	&Q(z^{sa}, z^{s}, \phi), \\
	&\phi := \text{AvgL1Norm}(\text{Linear}(s,a)). 
\end{align}
Throughout all experiments, we assume no value clipping (as this is introduced in response to the analysis from this section). This makes the value function loss~(originally \autoref{appendix:eqn:critic_loss}):
\begin{align}
	\Loss(Q_{t+1}) :=&~\text{Huber}\Bigl( \texttt{target} - Q_{t+1}(z_{t}^{sa}, z_{t}^s, s, a) \Bigr),\\
	\texttt{target} :=&~r + \y \min \lp Q_{t,1} (x), Q_{t,2} (x) \rp, \\
	x :=&~[z_{t-1}^{s'a'}, z_{t-1}^{s'}, s', a'],\\
	a' :=&~\pi_{t}(z^{s'}_{t-1}, s') + \e, \\
	\e \sim&~\text{clip}(\N(0,\sigma^2), -c, c). 
\end{align}
Policy checkpoints are not used and the policy is not modified. Since extrapolation error is closely linked to available data, we also vary the maximum size of the replay buffer (default: 1M).%

(\autoref{appendix:fig:extrapolation:clipping}) \textbf{Extrapolation 1:} No clipping. This is TD7 without value clipping in the value function loss as discussed above.

(\autoref{appendix:fig:extrapolation:nozsa}) \textbf{Extrapolation 2:} No $z^{sa}$. We remove $z^{sa}$ from the value function input. This makes the input: $Q_{t+1}(z^s_t, \phi)$. 

(\autoref{appendix:fig:extrapolation:smallphi}) \textbf{Extrapolation 3:} Small $\phi$. We reduce the number of dimensions of $\phi$ from $256$ to $16$. 

(\autoref{appendix:fig:extrapolation:nozsa_smallphi}) \textbf{Extrapolation 4:} No $z^{sa}$ and small $\phi$. We remove $z^{sa}$ from the value function input and reduce the number of dimensions of $\phi$ from $256$ to $16$.

(\autoref{appendix:fig:extrapolation:frozen}) \textbf{Extrapolation 5:} Frozen embeddings. The encoders are left unchanged throughout training, by leaving them untrained. This means the input of the value function is $Q_{t+1}(z^s_0, \phi)$. 

(\autoref{appendix:fig:extrapolation:TD7}) \textbf{Extrapolation 6:} TD7. The full TD7 method (without checkpoints).  

\clearpage

\begin{figure}[ht]
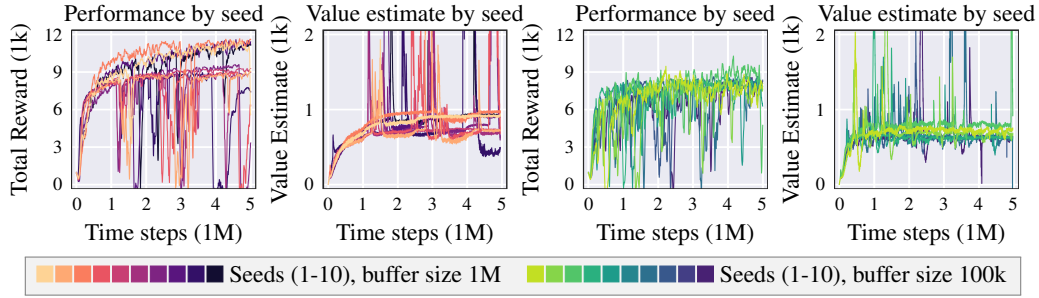

\appendixextrapolation{0}
\fcolorbox{gray}{gray!10}{
\small
\cblockfull{m_1}~\cblockfull{m_2}~\cblockfull{m_3}~\cblockfull{m_4}~\cblockfull{m_5}~\cblockfull{m_6}~\cblockfull{m_7}~\cblockfull{m_8}~\cblockfull{m_9}~\cblockfull{m_10}~Seeds (1-10), buffer size 1M \quad 
\cblockfull{v_1}~\cblockfull{v_2}~\cblockfull{v_3}~\cblockfull{v_4}~\cblockfull{v_5}~\cblockfull{v_6}~\cblockfull{v_7}~\cblockfull{v_8}~\cblockfull{v_9}~\cblockfull{v_10}~Seeds (1-10), buffer size 100k
} 
\caption{\textbf{No clipping.} The performance curve and corresponding value estimate made by the value function, where no value clipping is used. The results for each individual seed are presented for a replay buffer size of either 1M (left) or 100k (right). The environment used is Ant.} \label{appendix:fig:extrapolation:clipping}
\end{figure}

\begin{figure}[ht]
	\appendixextrapolation{2}
	\fcolorbox{gray}{gray!10}{
		\small
		\cblockfull{m_1}~\cblockfull{m_2}~\cblockfull{m_3}~\cblockfull{m_4}~\cblockfull{m_5}~\cblockfull{m_6}~\cblockfull{m_7}~\cblockfull{m_8}~\cblockfull{m_9}~\cblockfull{m_10}~Seeds (1-10), buffer size 1M \quad 
		\cblockfull{v_1}~\cblockfull{v_2}~\cblockfull{v_3}~\cblockfull{v_4}~\cblockfull{v_5}~\cblockfull{v_6}~\cblockfull{v_7}~\cblockfull{v_8}~\cblockfull{v_9}~\cblockfull{v_10}~Seeds (1-10), buffer size 100k
	} 
	\caption{\textbf{No $z^{sa}$.} The performance curve and corresponding value estimate made by the value function, where $z^{sa}$ is not included in the input. The results for each individual seed are presented for a replay buffer size of either 1M (left) or 100k (right). The environment used is Ant.} \label{appendix:fig:extrapolation:nozsa}
\end{figure}

\begin{figure}[ht]
	\appendixextrapolation{3}
	\fcolorbox{gray}{gray!10}{
		\small
		\cblockfull{m_1}~\cblockfull{m_2}~\cblockfull{m_3}~\cblockfull{m_4}~\cblockfull{m_5}~\cblockfull{m_6}~\cblockfull{m_7}~\cblockfull{m_8}~\cblockfull{m_9}~\cblockfull{m_10}~Seeds (1-10), buffer size 1M \quad 
		\cblockfull{v_1}~\cblockfull{v_2}~\cblockfull{v_3}~\cblockfull{v_4}~\cblockfull{v_5}~\cblockfull{v_6}~\cblockfull{v_7}~\cblockfull{v_8}~\cblockfull{v_9}~\cblockfull{v_10}~Seeds (1-10), buffer size 100k
	} 
	\caption{\textbf{Small $\phi$.} The performance curve and corresponding value estimate made by the value function, where the dimension size of $\phi$ is reduced. The results for each individual seed are presented for a replay buffer size of either 1M (left) or 100k (right). The environment used is Ant.} \label{appendix:fig:extrapolation:smallphi}
\end{figure}

\clearpage

\begin{figure}[ht]
	\appendixextrapolation{4}
	\fcolorbox{gray}{gray!10}{
		\small
		\cblockfull{m_1}~\cblockfull{m_2}~\cblockfull{m_3}~\cblockfull{m_4}~\cblockfull{m_5}~\cblockfull{m_6}~\cblockfull{m_7}~\cblockfull{m_8}~\cblockfull{m_9}~\cblockfull{m_10}~Seeds (1-10), buffer size 1M \quad 
		\cblockfull{v_1}~\cblockfull{v_2}~\cblockfull{v_3}~\cblockfull{v_4}~\cblockfull{v_5}~\cblockfull{v_6}~\cblockfull{v_7}~\cblockfull{v_8}~\cblockfull{v_9}~\cblockfull{v_10}~Seeds (1-10), buffer size 100k
	} 
	\caption{\textbf{No $z^{sa}$, small $\phi$.} The performance curve and corresponding value estimate made by the value function, where $z^{sa}$ is not included in the input and the dimension size of $\phi$ is reduced. The results for each individual seed are presented for a replay buffer size of either 1M (left) or 100k (right). The environment used is Ant.} \label{appendix:fig:extrapolation:nozsa_smallphi}
\end{figure}

\begin{figure}[ht]
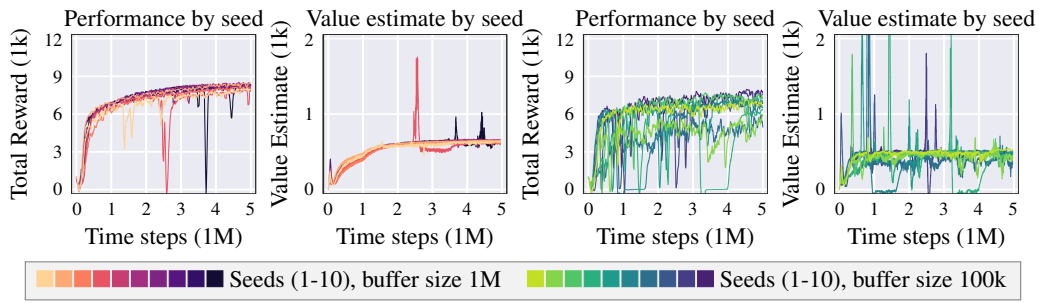

	\appendixextrapolation{5}
	\fcolorbox{gray}{gray!10}{
		\small
		\cblockfull{m_1}~\cblockfull{m_2}~\cblockfull{m_3}~\cblockfull{m_4}~\cblockfull{m_5}~\cblockfull{m_6}~\cblockfull{m_7}~\cblockfull{m_8}~\cblockfull{m_9}~\cblockfull{m_10}~Seeds (1-10), buffer size 1M \quad 
		\cblockfull{v_1}~\cblockfull{v_2}~\cblockfull{v_3}~\cblockfull{v_4}~\cblockfull{v_5}~\cblockfull{v_6}~\cblockfull{v_7}~\cblockfull{v_8}~\cblockfull{v_9}~\cblockfull{v_10}~Seeds (1-10), buffer size 100k
	} 
	\caption{\textbf{Frozen embeddings.} The performance curve and corresponding value estimate made by the value function, where the encoders are not trained. The results for each individual seed are presented for a replay buffer size of either 1M (left) or 100k (right). The environment used is Ant.} \label{appendix:fig:extrapolation:frozen}
\end{figure}

\begin{figure}[ht]
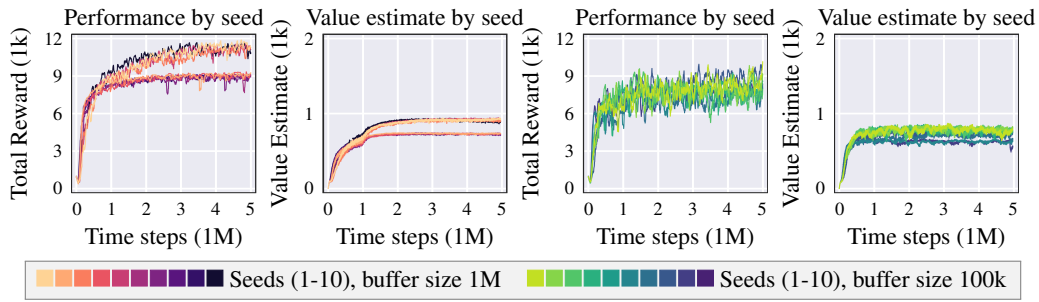

	\appendixextrapolation{1}
	\fcolorbox{gray}{gray!10}{
		\small
		\cblockfull{m_1}~\cblockfull{m_2}~\cblockfull{m_3}~\cblockfull{m_4}~\cblockfull{m_5}~\cblockfull{m_6}~\cblockfull{m_7}~\cblockfull{m_8}~\cblockfull{m_9}~\cblockfull{m_10}~Seeds (1-10), buffer size 1M \quad 
		\cblockfull{v_1}~\cblockfull{v_2}~\cblockfull{v_3}~\cblockfull{v_4}~\cblockfull{v_5}~\cblockfull{v_6}~\cblockfull{v_7}~\cblockfull{v_8}~\cblockfull{v_9}~\cblockfull{v_10}~Seeds (1-10), buffer size 100k
	} 
	\caption{\textbf{TD7.} The performance curve and corresponding value estimate made by the value function of TD7 (using value clipping). The results for each individual seed are presented for a replay buffer size of either 1M (left) or 100k (right). The environment used is Ant.} \label{appendix:fig:extrapolation:TD7}
\end{figure}

\clearpage

\section{Policy Checkpoints}\label{appendix:sec:checkpoints}

\subsection{Motivation}

Deep RL algorithms are widely known for their inherent instability, which often results in substantial variance in performance during training. Instability in deep RL can occur on a micro timescale (performance can shift dramatically between episodes)~\citep{henderson2017deep, fujimoto2021minimalist} and a macro timescale (the algorithm can diverge or collapse with too much training)~\citep{kumar2020implicit, lyle2021understanding}. In Figure \ref{fig:presentation} \& \ref{fig:unstable} we show the instability of RL methods by (a) showing how presenting the average performance can hide the instability in a single trial of RL (\autoref{fig:presentation}), and (b) measuring how much the performance can drop between nearby evaluations (\autoref{fig:unstable}). 

\begin{figure}[ht]
\begin{minipage}{0.2\textwidth}
\hspace{-8pt}
\begin{tikzpicture}[trim axis right]
\begin{axis}[
    width=\textwidth,
    title={Hopper},
    ylabel style={yshift=-2pt},
    xlabel={Time steps (1M)},
    ylabel={Total Reward (1k)},
    xtick={0, 1, 2, 3, 4, 5},
    xticklabels={0, 1, 2, 3, 4, 5},
    ytick={0, 1, 2, 3, 4},
    yticklabels={\po\po0, \po\po1, \po\po2, \po\po3, \po\po4},
]
\addplot graphics [
ymin=-0.12, ymax=4.12,
xmin=-0.15, xmax=5.15,
]{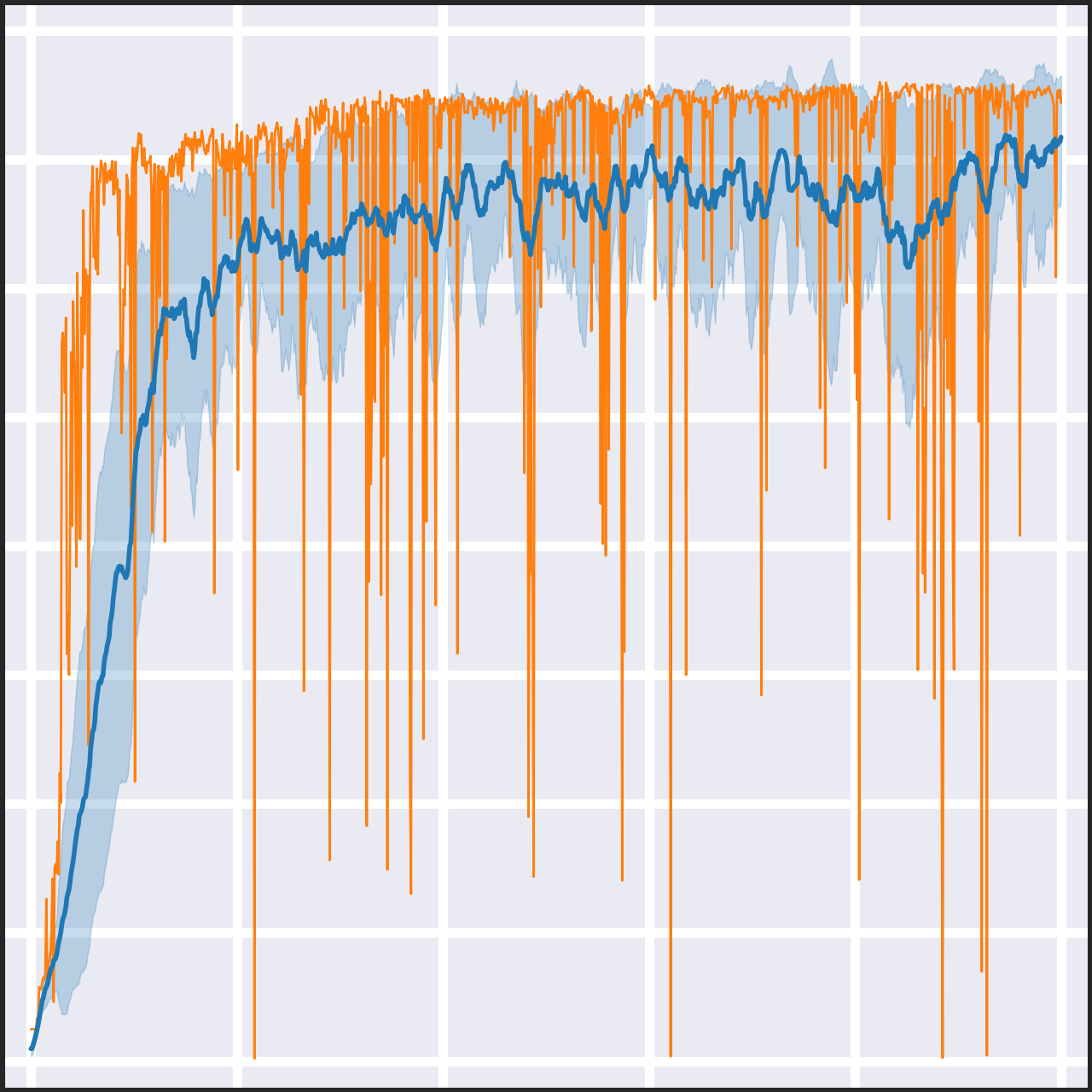};
\end{axis}
\end{tikzpicture}
\end{minipage}
\hfill
\begin{minipage}{0.71\textwidth}
\captionof{figure}[]{\quad \cblock{sb_blue}~Standard presentation \quad \cblockfull{sb_orange}~Single seed

The training curve as commonly presented in RL papers~~\cblock{sb_blue} and the training curve of a single seed~~\cblockfull{sb_orange}. Both curves are from the TD3 algorithm, trained for 5M time steps. 
The standard presentation is to evaluate every $N_\text{freq}$ steps, average scores over $N_\text{episodes}$ evaluation episodes and $N_\text{seeds}$ seeds, then to smooth the curve by averaging over a window of $N_\text{window}$ evaluations. (In our case this corresponds to $N_\text{freq}=5000, N_\text{episodes}=10, N_\text{seeds}=10, N_\text{window}=10$). %
The learning curve of a single seed has no smoothing over seeds or evaluations ($N_\text{seeds}=1$, $N_\text{window}=1$). 
By averaging over many seeds and evaluations, the training curves in RL can appear deceptively smooth and stable. 
}\label{fig:presentation}
\end{minipage}

\begin{minipage}{0.2\textwidth}
\hspace{-8pt}
\begin{tikzpicture}[trim axis right]
\begin{axis}[
    width=\textwidth,
    title={Hopper},
    ylabel style={yshift=-2pt},
    xlabel={> \% Decrease},
    ylabel={\phantom{(}\% of Evaluations\phantom{)}},
    xtick={0, 20, 40, 60, 80, 100},
    xticklabels={0, 20, 40, 60, 80, 100},
    ytick={0, 20, 40, 60, 80, 100},
    yticklabels={0, 20, 40, 60, 80, 100},
]
\addplot graphics [
ymin=0, ymax=103,
xmin=-3, xmax=106,
]{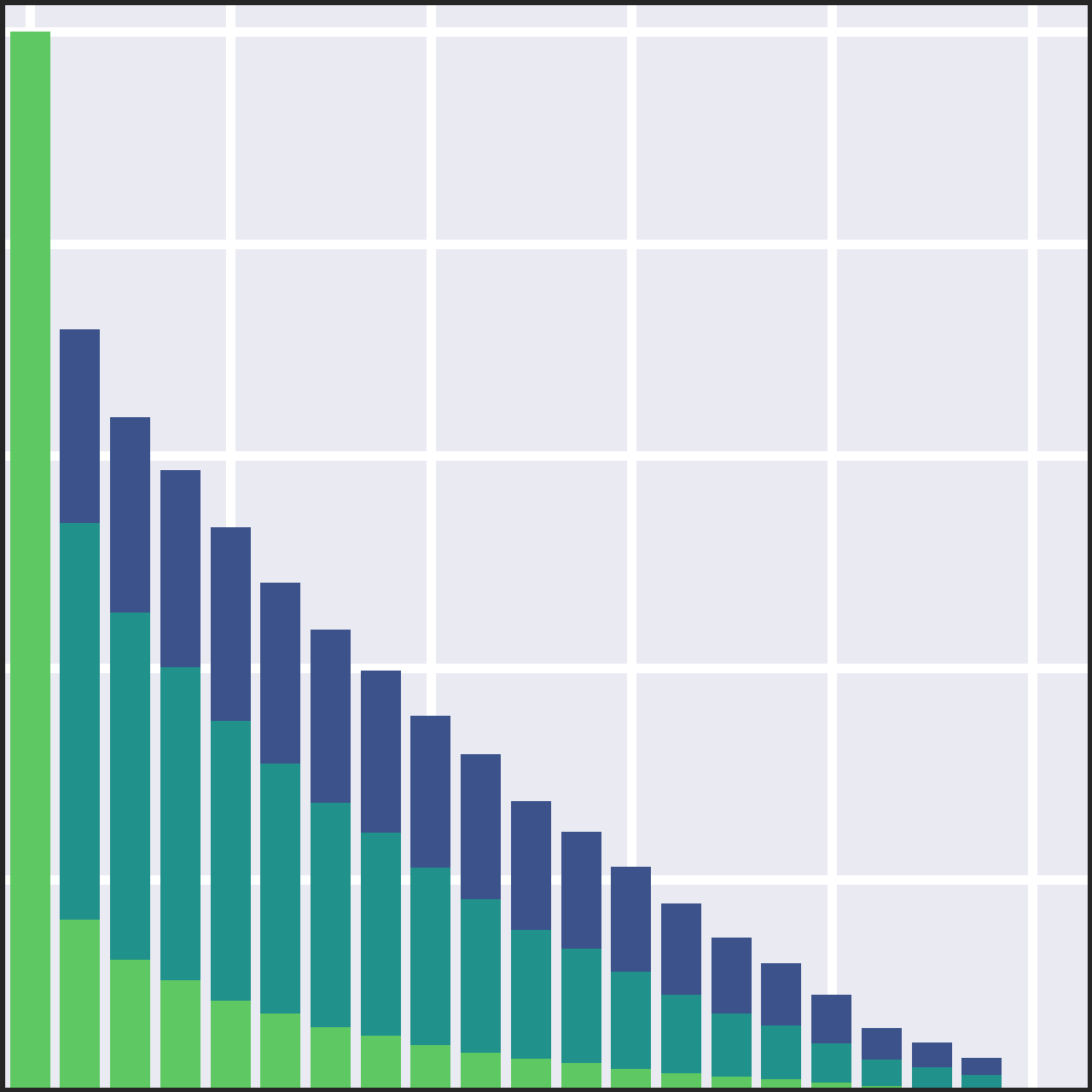};
\end{axis}
\end{tikzpicture}
\end{minipage}
\hfill
\begin{minipage}{0.71\textwidth}
\captionof{figure}[]{
\quad 
\cblockfull{e_3}~$\texttt{window}=1$ \quad \cblockfull{e_2}~$\texttt{window}=5$ \quad \cblockfull{e_1}~$\texttt{window}=10$ 

The \% of evaluations which suffer from a performance drop of at least $x$\% in the next $\texttt{window}$ evaluations for the TD3 algorithm, trained for 5M time steps. 
For example, for any given evaluation, the likelihood of one of the next $10$ evaluations performing at least $50$\% worse is around $30$\%.  Only evaluations from the last $1$M training steps are used, so that early training instability is not accounted for. This shows that large drops in performance are common.} \label{fig:unstable}
\end{minipage}
\end{figure}

\subsection{Method}

The overview for our approach for using checkpoints in RL:
\begin{itemize}[nosep]
	\item Assess the current policy (using training episodes).
	\item Train the current policy (with a number of time steps equal (or proportional) to the number of time steps viewed during assessment). 
	\item If the current policy outperforms the checkpoint policy, then update the checkpoint policy.
\end{itemize}
Note that we aim to use training episodes to evaluate the checkpoint policy, as any evaluation or test time episodes are considered entirely separate from the training process. For clarity, we will use the term \textit{assessment} (rather than evaluation) to denote any measure of performance which occurs during training. In \autoref{appendix:alg:basic_checkpoint} we outline the basic policy checkpoint strategy.

\tikzexternaldisable
\begin{algorithm}[ht]
\small
   \caption{Basic Policy Checkpoints} \label{appendix:alg:basic_checkpoint}
\begin{algorithmic}[1]
    \BeginBox[fill=gray!10]
    \For{$\texttt{episode}=1$ {\bfseries to} $\texttt{assessment\_episodes}$}
    \Comment{Assessment}
    \State Follow the current policy $\pi_{t+1}$ and determine \texttt{episode\_reward}.
    \State Update the performance measure of the current policy.
    \State Increment \texttt{timesteps\_since\_training} by the length of the episode.
    \EndFor
    \EndBox
    \BeginBox[fill=gray!20]
    \If{actor $\pi_{t+1}$ \texttt{outperforms} checkpoint policy $\pi_c$} 
    \Comment{Checkpointing}
    \State Update checkpoint networks $\pi_c \leftarrow \pi_{t+1}$, $f_c \leftarrow f_{t}$. 
    \State Update checkpoint performance. 
    \EndIf 
    \EndBox
    \BeginBox[fill=gray!10]
    \For{$i=1$ {\bfseries to} $\texttt{timesteps\_since\_training}$}
    \Comment{Training} 
    \State Train RL agent.
    \State Reset the performance measure of the current policy. 
    \EndFor
    \EndBox
\end{algorithmic}
\end{algorithm}
\tikzexternalenable

\textbf{Minimum performance criteria.} An interesting choice that remains is how to determine whether the current policy outperforms the checkpoint policy. %
Instead of using the average performance, we consider the minimum performance over a small number of assessment episodes. Using the minimum provides a criteria which is more sensitive to instability, favoring policies which achieve a consistent performance. 

Additionally, using the minimum allows us to end the assessment phase early, if the minimum performance of the current policy drops below the minimum performance of the checkpoint policy. This allows us to use a higher maximum number of assessment episodes, while not wasting valuable training episodes assessing suboptimal policies. In \autoref{appendix:alg:checkpoint} we outline policy checkpoints where the minimum performance is used, and the assessment phase is variable length due to early termination. 

\tikzexternaldisable
\begin{algorithm}[ht]
\small
   \caption{Policy Checkpoints with Minimum Performance and Early Termination} \label{appendix:alg:checkpoint}
\begin{algorithmic}[1]
    \BeginBox[fill=gray!10]
    \For{$\texttt{episode}=1$ {\bfseries to} $\texttt{assessment\_episodes}$}
    \Comment{Assessment}
    \State Follow the current policy $\pi_{t+1}$ and determine \texttt{episode\_reward}.
    \State $\texttt{min\_performance} \leftarrow \min(\texttt{min\_performance}, \texttt{episode\_reward} )$.
	\State Increment \texttt{timesteps\_since\_training} by the length of the episode.
    \EndBox
    \BeginBox[fill=gray!20]
    \If{$\texttt{min\_performance} \leq \texttt{checkpoint\_performance}$} 
    \Comment{Early termination}
    \State End current assessment. 
    \EndIf
    \EndFor
    \EndBox
    \BeginBox[fill=gray!10]
    \If{$\texttt{min\_performance} \geq \texttt{checkpoint\_performance}$} 
    \Comment{Checkpointing}
    \State Update checkpoint networks $\pi_c \leftarrow \pi_{t+1}$, $f_c \leftarrow f_{t}$. 
    \State $\texttt{checkpoint\_performance} \leftarrow \texttt{min\_performance}$
    \EndIf
    \EndBox
    \BeginBox[fill=gray!20]
    \For{$i=1$ {\bfseries to} $\texttt{timesteps\_since\_training}$}
    \Comment{Training} 
    \State Train RL agent.
    \State Reset \texttt{min\_performance}.
    \EndFor
    \EndBox
\end{algorithmic}
\end{algorithm}
\tikzexternalenable

\textbf{Early learning.} In our empirical analysis we observe that long assessment phases had a negative impact on early learning (\autoref{appendix:table:checkpoint_immediate}). A final adjustment to policy checkpoints is to keep the number of assessment episodes fixed to 1 during the early stages of learning, and then increase it to a much larger value later in learning. Since the policy or environment may be stochastic, when increasing the number of assessment episodes, we reduce the performance of the checkpoint policy (by a factor of $0.9$), since the checkpoint performance may be overfit to the single episode used to assess its performance.

In Algorithm 1 in the main text, several variables are referenced. These are as follows:
\begin{itemize}[nosep, itemsep=3pt]
    \item \texttt{checkpoint\_condition}: The checkpoint condition refers to either (a) the maximum number of assessment episodes being reached (20) or the current minimum performance dropping below the minimum performance of the checkpoint policy.
    \item \texttt{outperforms}: The maximum number of assessment episodes were reached and the current minimum performance is higher than the minimum performance of the checkpoint policy. 
\end{itemize}

\textbf{Variable descriptions.} In \autoref{table:hyperparameters} (in \autoref{appendix:sec:TD7_hyperparameters}), which describes the hyperparameters of TD7, several variables are referenced. These are as follows:
\begin{itemize}[nosep, itemsep=3pt]
	\item Checkpoint criteria: the measure used to evaluate the performance of the current and checkpoint policy (default: minimum performance over the assessment episodes). 
	\item Early assessment episodes: the maximum number of assessment episodes during the early stage of learning (default: 1).
	\item Late assessment episodes: the maximum number of assessment episodes during the late stage of learning (default: 20).
	\item Early time steps: the duration of the early learning stage in time steps. We consider early learning to be the first 750k time steps.
	\item Criteria reset weight: The multiplier on the current performance of the checkpoint policy that is applied once, after the early learning stage ends (default: $0.9$).
\end{itemize}

\clearpage

\subsection{Empirical Analysis}

TD7 uses a checkpointing scheme which:
\begin{itemize}[nosep, itemsep=3pt]
	\item Measures the performance of the current and checkpoint policy by the minimum performance over 20 assessment episodes.
	\item Uses a variable length assessment phase, where the assessment phase terminates early if the minimum performance of the current policy is less than the minimum performance of the checkpoint policy.
	\item Uses an early learning stage where the maximum number of assessment episodes starts at 1 and is increased to 20 after 750k time steps. 
\end{itemize}

In this section, we vary some of the choices made when using policy checkpoints and analyze the stability benefits of using checkpoints. 

(\autoref{appendix:fig:checkpoint:assessment_episodes}) \textbf{Checkpoint 1:} Maximum number of assessment episodes. We vary the maximum number episodes used to assess the performance of the current policy (default 20). Early termination of the assessment phase is still used. 

(\autoref{appendix:fig:checkpoint:mean}) \textbf{Checkpoint 2:} Mean, fixed assessment length. We use the mean performance instead of the minimum (default) to determine if the current policy outperforms the checkpoint policy. In this case, early termination of the assessment phase is not used. The number of assessment episodes is fixed at 20.

(\autoref{appendix:fig:checkpoint:mean}) \textbf{Checkpoint 3:} Mean, variable assessment length. The mean performance is used to determine if the current policy outperforms the checkpoint policy. Early termination of the assessment phase is used. After each episode, we terminate the assessment phase if the current mean performance is below the checkpoint performance. This means the number of assessment episodes is variable, up to a maximum of 20.

(\autoref{appendix:fig:checkpoint:fixed}) \textbf{Checkpoint 4:} Fixed assessment length. The same number of assessment episodes is used during each assessment phase, without early termination. %

(\autoref{appendix:fig:checkpoint:early}) \textbf{Checkpoint 5:} Immediate. The number of assessment episodes starts at the corresponding number (as opposed to starting at 1 and increasing to a higher number after the early learning state completes at 750k time steps). 

In \autoref{appendix:table:checkpoint_main} we present the main set of results at 5M time steps. To measure early learning performance, in \autoref{appendix:table:checkpoint_immediate} we present the performance of using immediate checkpoints at 300k time steps. To observe stability benefits, in \autoref{appendix:fig:singleseed} we display the performance of five individual seeds, with and without checkpoints.

\begin{table*}[ht]
\centering
\small
\setlength{\tabcolsep}{4pt}
\newcolumntype{Y}{>{\centering\arraybackslash}X} %
\caption{Average performance on the MuJoCo benchmark at 5M time steps. $\pm$~captures a $95\%$ confidence interval around the average performance. Results are over $10$ seeds.} \label{appendix:table:checkpoint_main}
\begin{tabularx}{\textwidth}{lYYYYY}
\toprule
Algorithm & HalfCheetah & Hopper & Walker2d & Ant & Humanoid \\ 
\midrule
TD7 & 18165 \textcolor{gray}{$\pm$ 255\po} & \po4075 \textcolor{gray}{$\pm$ 225\po} & \po7397 \textcolor{gray}{$\pm$ 454\po} & 10133 \textcolor{gray}{$\pm$ 966\po} & 10281 \textcolor{gray}{$\pm$ 588\po} \\
TD7 (no checkpoints) & 18328 \textcolor{gray}{$\pm$ 331\po} & \po3851 \textcolor{gray}{$\pm$ 372\po} & \po6519 \textcolor{gray}{$\pm$ 209\po} & 10388 \textcolor{gray}{$\pm$ 1024} & \po9521 \textcolor{gray}{$\pm$ 820\po} \\
\midrule
Max 10 episodes & 18257 \textcolor{gray}{$\pm$ 338\po} & \po4208 \textcolor{gray}{$\pm$ 52\po\po} & \po6856 \textcolor{gray}{$\pm$ 273\po} & \po9890 \textcolor{gray}{$\pm$ 661\po} & 10689 \textcolor{gray}{$\pm$ 576\po} \\
Max 50 episodes & 17875 \textcolor{gray}{$\pm$ 192\po} & \po4110 \textcolor{gray}{$\pm$ 83\po\po} & \po7104 \textcolor{gray}{$\pm$ 559\po} & \po8226 \textcolor{gray}{$\pm$ 743\po} & 10239 \textcolor{gray}{$\pm$ 273\po} \\
\midrule
Mean, fixed & 18137 \textcolor{gray}{$\pm$ 417\po} & \po4061 \textcolor{gray}{$\pm$ 160\po} & \po7347 \textcolor{gray}{$\pm$ 459\po} & \po9866 \textcolor{gray}{$\pm$ 804\po} & \po9850 \textcolor{gray}{$\pm$ 497\po} \\
Mean, variable & 17943 \textcolor{gray}{$\pm$ 291\po} & \po4217 \textcolor{gray}{$\pm$ 9\po\po\po} & \po6845 \textcolor{gray}{$\pm$ 476\po} & \po9614 \textcolor{gray}{$\pm$ 849\po} & 10784 \textcolor{gray}{$\pm$ 188\po} \\
\midrule
Immediate, 10 episodes & 17767 \textcolor{gray}{$\pm$ 374\po} & \po4218 \textcolor{gray}{$\pm$ 17\po\po} & \po7416 \textcolor{gray}{$\pm$ 451\po} & 10090 \textcolor{gray}{$\pm$ 961\po} & \po9573 \textcolor{gray}{$\pm$ 255\po} \\
Immediate, 20 episodes & 17900 \textcolor{gray}{$\pm$ 127\po} & \po4078 \textcolor{gray}{$\pm$ 53\po\po} & \po7696 \textcolor{gray}{$\pm$ 616\po} & \po9576 \textcolor{gray}{$\pm$ 935\po} & 10219 \textcolor{gray}{$\pm$ 551\po} \\
Immediate, 50 episodes & 17619 \textcolor{gray}{$\pm$ 275\po} & \po4033 \textcolor{gray}{$\pm$ 154\po} & \po6712 \textcolor{gray}{$\pm$ 313\po} & \po9716 \textcolor{gray}{$\pm$ 1091} & \po9687 \textcolor{gray}{$\pm$ 584\po} \\
\midrule
Fixed, 10 episodes & 18209 \textcolor{gray}{$\pm$ 332\po} & \po4130 \textcolor{gray}{$\pm$ 56\po\po} & \po6890 \textcolor{gray}{$\pm$ 410\po} & 10357 \textcolor{gray}{$\pm$ 893\po} & \po9682 \textcolor{gray}{$\pm$ 1408} \\
Fixed, 20 episodes & 17925 \textcolor{gray}{$\pm$ 334\po} & \po3777 \textcolor{gray}{$\pm$ 337\po} & \po6882 \textcolor{gray}{$\pm$ 452\po} & \po9716 \textcolor{gray}{$\pm$ 772\po} & 10471 \textcolor{gray}{$\pm$ 560\po} \\
Fixed, 50 episodes & 17807 \textcolor{gray}{$\pm$ 183\po} & \po3759 \textcolor{gray}{$\pm$ 142\po} & \po6800 \textcolor{gray}{$\pm$ 422\po} & \po8881 \textcolor{gray}{$\pm$ 408\po} & \po8830 \textcolor{gray}{$\pm$ 655\po} \\
\bottomrule
\end{tabularx}
\end{table*}

\begin{table*}[ht]
	\centering
	\small
	\setlength{\tabcolsep}{4pt}
	\newcolumntype{Y}{>{\centering\arraybackslash}X} %
	\caption{Average performance on the MuJoCo benchmark at 300k time steps. $\pm$~captures a $95\%$ confidence interval around the average performance. Results are over $10$ seeds.} \label{appendix:table:checkpoint_immediate}
	\begin{tabularx}{\textwidth}{lYYYYY}
		\toprule
		Algorithm & HalfCheetah & Hopper & Walker2d & Ant & Humanoid \\ 
		\midrule
		TD7 & 15031 \textcolor{gray}{$\pm$ 401\po} & \po2948 \textcolor{gray}{$\pm$ 464\po} & \po5379 \textcolor{gray}{$\pm$ 328\po} & \po6171 \textcolor{gray}{$\pm$ 831\po} & \po5332 \textcolor{gray}{$\pm$ 714\po} \\
		\midrule
		Immediate, 10 episodes & 13778 \textcolor{gray}{$\pm$ 856\po} & \po3306 \textcolor{gray}{$\pm$ 34\po\po} & \po4595 \textcolor{gray}{$\pm$ 343\po} & \po5977 \textcolor{gray}{$\pm$ 764\po} & \po4660 \textcolor{gray}{$\pm$ 735\po} \\
		Immediate, 20 episodes & 12289 \textcolor{gray}{$\pm$ 507\po} & \po3052 \textcolor{gray}{$\pm$ 195\po} & \po4518 \textcolor{gray}{$\pm$ 321\po} & \po5504 \textcolor{gray}{$\pm$ 889\po} & \po4531 \textcolor{gray}{$\pm$ 738\po} \\
		Immediate, 50 episodes & \po6658 \textcolor{gray}{$\pm$ 1309} & \po2668 \textcolor{gray}{$\pm$ 247\po} & \po2897 \textcolor{gray}{$\pm$ 816\po} & \po3709 \textcolor{gray}{$\pm$ 1848} & \po2060 \textcolor{gray}{$\pm$ 620\po} \\
		\bottomrule
	\end{tabularx}
\end{table*}

\begin{figure}[ht]
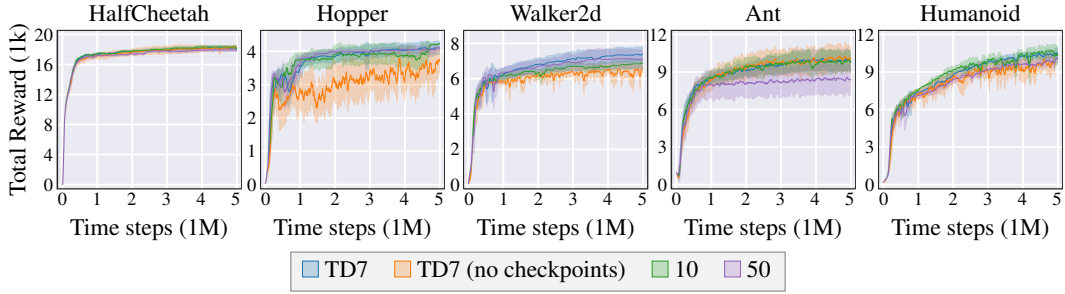

\appendixfigurecheck{checkpoint_0}
\fcolorbox{gray}{gray!10}{
\small
\cblock{sb_blue}~TD7 \quad \cblock{sb_orange}~TD7 (no checkpoints) \quad \cblock{sb_green}~10 \quad \cblock{sb_purple}~50
} 
\caption{\textbf{Maximum number of assessment episodes.} Learning curves on the MuJoCo benchmark, varying the maximum number episodes that the policy is fixed for. Results are averaged over $10$ seeds. The shaded area captures a $95\%$ confidence interval around the average performance.} \label{appendix:fig:checkpoint:assessment_episodes}
\end{figure}

\begin{figure}[ht]
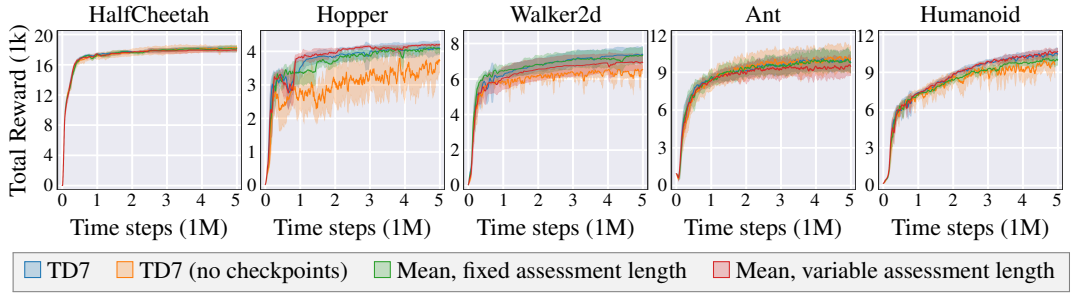

\appendixfigurecheck{checkpoint_1}
\fcolorbox{gray}{gray!10}{
\small
\cblock{sb_blue}~TD7 \quad \cblock{sb_orange}~TD7 (no checkpoints) \quad \cblock{sb_green}~Mean, fixed assessment length \quad \cblock{sb_red}~Mean, variable assessment length
} 
\caption{\textbf{Checkpoint condition.} Learning curves on the MuJoCo benchmark, using the mean performance of assessment episodes rather than the minimum. The number of assessment episodes is kept at $20$. Results are averaged over $10$ seeds. The shaded area captures a $95\%$ confidence interval around the average performance.} \label{appendix:fig:checkpoint:mean}
\end{figure}

\begin{figure}[ht]
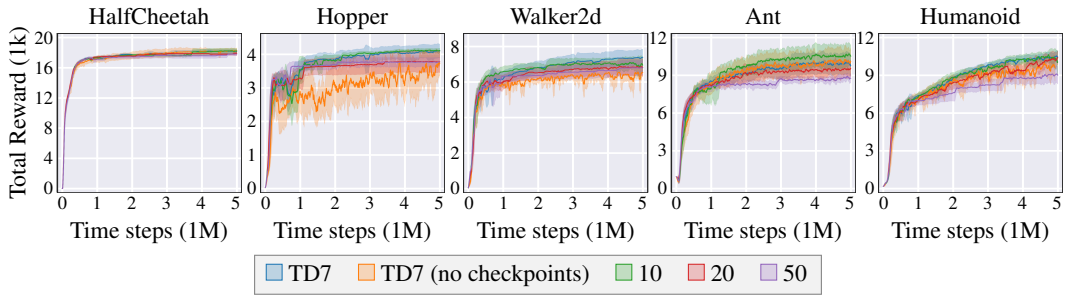

\appendixfigurecheck{checkpoint_3}
\fcolorbox{gray}{gray!10}{
\small
\cblock{sb_blue}~TD7 \quad \cblock{sb_orange}~TD7 (no checkpoints) \quad \cblock{sb_green}~10 \quad \cblock{sb_red}~20 \quad \cblock{sb_purple}~50
} 
\caption{\textbf{Fixed assessment length.} Learning curves on the MuJoCo benchmark, where there is no early termination of the assessment phase (i.e\ the number of assessment episodes is fixed), and the number of assessment episodes is varied. Results are averaged over $10$ seeds. The shaded area captures a $95\%$ confidence interval around the average performance.} \label{appendix:fig:checkpoint:fixed}
\end{figure}

\clearpage

\begin{figure}[ht]
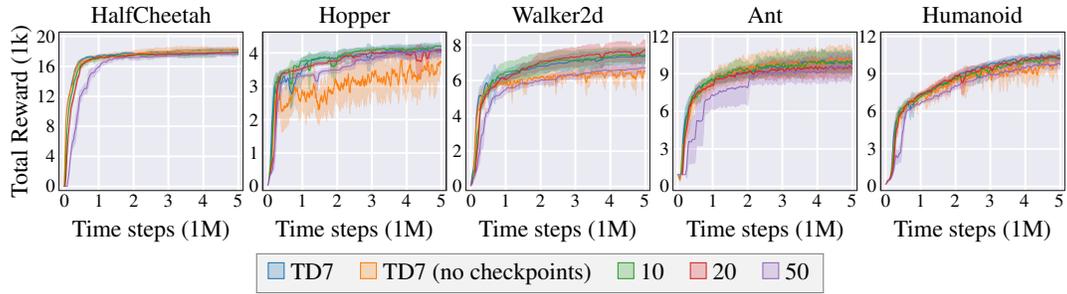

\appendixfigurecheck{checkpoint_2}
\fcolorbox{gray}{gray!10}{
\small
\cblock{sb_blue}~TD7 \quad \cblock{sb_orange}~TD7 (no checkpoints) \quad \cblock{sb_green}~10 \quad \cblock{sb_red}~20 \quad \cblock{sb_purple}~50
} 
\caption{\textbf{Immediate.} Learning curves on the MuJoCo benchmark, where the maximum number of assessment episodes does not start at $1$ (i.e.\ there is no early learning phase) and is varied. Results are averaged over $10$ seeds. The shaded area captures a $95\%$ confidence interval around the average performance.} \label{appendix:fig:checkpoint:early}
\end{figure}

\begin{figure}[ht]
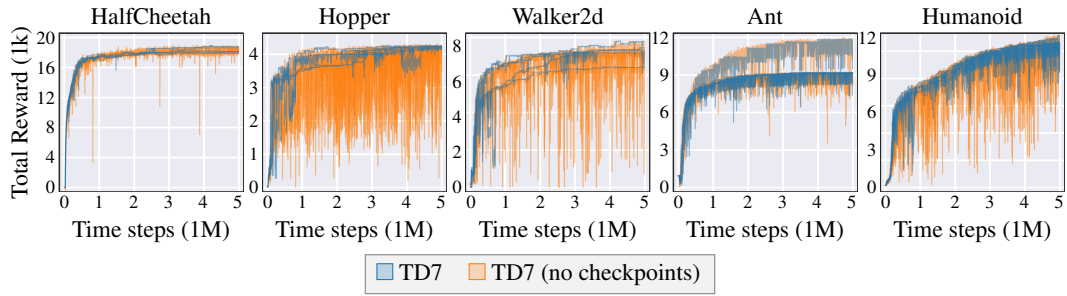

	\appendixfigurecheck{singleseed_TD7}
	\fcolorbox{gray}{gray!10}{
		\small
		\cblock{sb_blue}~TD7 \quad \cblock{sb_orange}~TD7 (no checkpoints)
	} 
	\caption{\textbf{Performance of individual seeds with and without checkpoints.} Learning curves of five individual seeds, with and without checkpoints, on the MuJoCo benchmark. The shaded area captures a $95\%$ confidence interval around the average performance.} \label{appendix:fig:singleseed}
\end{figure}

\clearpage

\section{Ablation Study} \label{appendix:sec:ablation}

In this section we perform an ablation study on the components of TD7. 

\subsection{Main Components}

(\autoref{appendix:fig:main_ablation}) \textbf{Ablation 1:} No SALE. SALE is entirely removed from TD7. This means the input to the value function and policy return to the original $Q(s,a)$ and $\pi(s)$, respectively. TD7 keeps LAP, policy checkpoints, and any implementation differences over TD3 (discussed in \autoref{appendix:sec:algorithm}).

(\autoref{appendix:fig:main_ablation}) \textbf{Ablation 2:} No checkpoints. Similar to TD3 and other off-policy deep RL algorithms, TD7 is trained at every time step, rather than after a batch of episodes. The current policy is used at test time.

(\autoref{appendix:fig:main_ablation}) \textbf{Ablation 3:} No LAP. TD7 uses the standard replay buffer where transitions are sampled with uniform probability. The value function loss~(\autoref{appendix:eqn:critic_loss}) uses the mean-squared error (MSE) rather than the Huber loss.

\begin{table*}[ht]
\centering
\small
\setlength{\tabcolsep}{4pt}
\newcolumntype{Y}{>{\centering\arraybackslash}X} %
\caption{Average performance on the MuJoCo benchmark at 5M time steps. $\pm$~captures a $95\%$ confidence interval around the average performance. Results are over $10$ seeds.} %
\begin{tabularx}{\textwidth}{lYYYYY}
\toprule
Algorithm & HalfCheetah & Hopper & Walker2d & Ant & Humanoid \\ 
\midrule
TD7 & 18165 \textcolor{gray}{$\pm$ 255\po}  & \po4075 \textcolor{gray}{$\pm$ 225\po}  & \po7397 \textcolor{gray}{$\pm$ 454\po}  & 10133 \textcolor{gray}{$\pm$ 966\po}  & 10281 \textcolor{gray}{$\pm$ 588\po} \\
TD3 & 14337 \textcolor{gray}{$\pm$ 1491}  & \po3682 \textcolor{gray}{$\pm$ 83\po\po}  & \po5078 \textcolor{gray}{$\pm$ 343\po}  & \po5589 \textcolor{gray}{$\pm$ 758\po}  & \po5433 \textcolor{gray}{$\pm$ 245\po} \\
\midrule
No SALE & 17099 \textcolor{gray}{$\pm$ 335\po}  & \po4018 \textcolor{gray}{$\pm$ 170\po}  & \po6418 \textcolor{gray}{$\pm$ 261\po}  & \po7861 \textcolor{gray}{$\pm$ 253\po}  & \po7275 \textcolor{gray}{$\pm$ 608\po} \\
No checkpoints & 18328 \textcolor{gray}{$\pm$ 331\po}  & \po3851 \textcolor{gray}{$\pm$ 372\po}  & \po6519 \textcolor{gray}{$\pm$ 209\po}  & 10388 \textcolor{gray}{$\pm$ 1024}  & \po9521 \textcolor{gray}{$\pm$ 820\po} \\
No LAP & 18104 \textcolor{gray}{$\pm$ 315\po}  & \po4188 \textcolor{gray}{$\pm$ 22\po\po}  & \po7233 \textcolor{gray}{$\pm$ 251\po}  & \po6940 \textcolor{gray}{$\pm$ 1044}  & 10155 \textcolor{gray}{$\pm$ 522\po} \\
\bottomrule
\end{tabularx}
\end{table*}

\begin{figure}[ht]
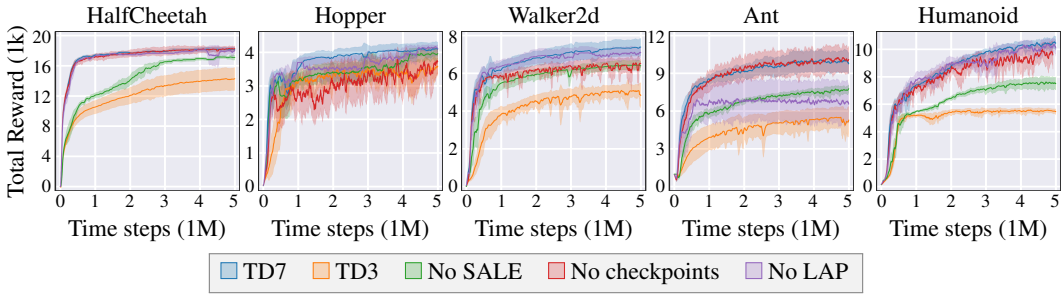

\appendixfigurefive{5M_ablation}
\fcolorbox{gray}{gray!10}{
\small
\cblock{sb_blue}~TD7 \quad \cblock{sb_orange}~TD3 \quad \cblock{sb_green}~No SALE \quad \cblock{sb_red} No checkpoints \quad \cblock{sb_purple} No LAP
} 
\caption{\textbf{Main ablation.} Learning curves on the MuJoCo benchmark, removing the main components of TD7. Results are averaged over $10$ seeds. The shaded area captures a $95\%$ confidence interval around the average performance.} \label{appendix:fig:main_ablation}
\end{figure}

\clearpage

\subsection{Subcomponents}

(\autoref{appendix:fig:main_ablation}) \textbf{Ablation 4:} No SALE, with encoder. The encoder is kept but trained end-to-end with the value function, rather than via the SALE loss function. This means the encoder can be simply treated as an extension of the architecture of the value function. Since the embeddings trained with the value function may be conflicting with the policy, the policy does not use the state embedding, and only takes its original input~$\pi(s)$.

(\autoref{appendix:fig:main_ablation}) \textbf{Ablation 5:} TD3 with encoder. We add the encoder from SALE to TD3 but train it end-to-end with the value function. The encoder is not applied to the policy. 

(\autoref{appendix:fig:ablation:checkpoints}) \textbf{Ablation 6:} Current policy. TD7 is trained in an identical fashion, but uses the current policy at test time, rather than the checkpoint policy. 

(\autoref{appendix:fig:ablation:checkpoints}) \textbf{Ablation 7:} TD3 with checkpoints. We add policy checkpoints to TD3, using the checkpoint policy at test time.

(\autoref{appendix:fig:ablation:LAP}) \textbf{Ablation 8:} TD3 with LAP. We add LAP to TD3 (identical to the TD3 + LAP~\citep{fujimoto2020equivalence}). 

(\autoref{appendix:fig:ablation:clipping}) \textbf{Ablation 9:} No clipping. We remove our proposed value clipping for mitigating extrapolation error. This makes the value function loss~(originally \autoref{appendix:eqn:critic_loss}):
\begin{align}
\Loss(Q_{t+1}) :=&~\text{Huber}\Bigl( \texttt{target} - Q_{t+1}(z_{t}^{sa}, z_{t}^s, s, a) \Bigr),\\
\texttt{target} :=&~r + \y \min \lp Q_{t,1} (x), Q_{t,2} (x) \rp, \\
x :=&~[z_{t-1}^{s'a'}, z_{t-1}^{s'}, s', a'],\\
a' :=&~\pi_{t}(z^{s'}_{t-1}, s') + \e, \\
\e \sim&~\text{clip}(\N(0,\sigma^2), -c, c). 
\end{align}

(\autoref{appendix:fig:ablation:clipping}) \textbf{Ablation 10:} TD3 with clipping. We add value clipping to the loss function of TD3. 

(\autoref{appendix:fig:ablation:SALE_improvements}) \textbf{Ablation 11:} No normalization. We remove the use of AvgL1Norm from TD7. This is identical to no normalization in the design study in \autoref{appendix:subsec:normalization}, except this version includes policy checkpoints (the design study does not use checkpoints).

(\autoref{appendix:fig:ablation:SALE_improvements}) \textbf{Ablation 12:} No fixed encoder. We remove the fixed embeddings from the value function and the policy, this means the networks use embeddings from the current encoder, rather than the fixed encoder from the previous iteration. This makes the input: $Q_{t+1}(z_{t+1}^{sa}, z^s_{t+1}, s, a)$ and $\pi_{t+1}(z^s_{t+1}, s)$. This is identical to no normalization in the design study in \autoref{appendix:subsec:normalization}, except this version includes policy checkpoints (the design study does not use checkpoints). 

(\autoref{appendix:fig:ablation:implementation}) \textbf{Ablation 13:} No implementation. We remove the implementation details differences discussed in \autoref{appendix:sec:algorithm}, namely using both value functions when updating the policy, ELU activation functions in the value function network. The target network update is not changed as this may influence the use of fixed encoders. 

(\autoref{appendix:fig:ablation:implementation}) \textbf{Ablation 14:} Our TD3. TD3 with the implementation detail differences discussed in \autoref{appendix:sec:algorithm}, namely using both value functions when updating the policy, ELU activation functions in the value function network, and updating the target network every $250$ time steps, rather than using an exponential moving average. 

\clearpage

\begin{table*}[ht]
\centering
\small
\setlength{\tabcolsep}{4pt}
\newcolumntype{Y}{>{\centering\arraybackslash}X} %
\caption{Average performance on the MuJoCo benchmark at 1M time steps. $\pm$~captures a $95\%$ confidence interval around the average performance. Results are over $10$ seeds.} %
\begin{tabularx}{\textwidth}{lYYYYY}
\toprule
Algorithm & HalfCheetah & Hopper & Walker2d & Ant & Humanoid \\ 
\midrule
TD7 & 17434 \textcolor{gray}{$\pm$ 155\po}  & \po3512 \textcolor{gray}{$\pm$ 315\po}  & \po6097 \textcolor{gray}{$\pm$ 570\po}  & \po8509 \textcolor{gray}{$\pm$ 422\po}  & \po7429 \textcolor{gray}{$\pm$ 153\po} \\
TD3 & 10574 \textcolor{gray}{$\pm$ 897\po}  & \po3226 \textcolor{gray}{$\pm$ 315\po}  & \po3946 \textcolor{gray}{$\pm$ 292\po}  & \po3942 \textcolor{gray}{$\pm$ 1030}  & \po5165 \textcolor{gray}{$\pm$ 145\po} \\
\midrule
No SALE & 12981 \textcolor{gray}{$\pm$ 261\po}  & \po3536 \textcolor{gray}{$\pm$ 65\po\po}  & \po5237 \textcolor{gray}{$\pm$ 376\po}  & \po5296 \textcolor{gray}{$\pm$ 1336}  & \po6263 \textcolor{gray}{$\pm$ 289\po} \\
No SALE, with encoder & 15639 \textcolor{gray}{$\pm$ 548\po}  & \po3544 \textcolor{gray}{$\pm$ 371\po}  & \po5350 \textcolor{gray}{$\pm$ 427\po}  & \po6003 \textcolor{gray}{$\pm$ 920\po}  & \po5480 \textcolor{gray}{$\pm$ 83\po\po} \\
TD3 with encoder & 12495 \textcolor{gray}{$\pm$ 813\po}  & \po1750 \textcolor{gray}{$\pm$ 302\po}  & \po4226 \textcolor{gray}{$\pm$ 491\po}  & \po5255 \textcolor{gray}{$\pm$ 579\po}  & \po5082 \textcolor{gray}{$\pm$ 317\po} \\
\midrule
No checkpoints & 17123 \textcolor{gray}{$\pm$ 296\po}  & \po3361 \textcolor{gray}{$\pm$ 429\po}  & \po5718 \textcolor{gray}{$\pm$ 308\po}  & \po8605 \textcolor{gray}{$\pm$ 1008}  & \po7381 \textcolor{gray}{$\pm$ 172\po} \\
Current policy & 17420 \textcolor{gray}{$\pm$ 273\po}  & \po2940 \textcolor{gray}{$\pm$ 636\po}  & \po5765 \textcolor{gray}{$\pm$ 800\po}  & \po8748 \textcolor{gray}{$\pm$ 397\po}  & \po7162 \textcolor{gray}{$\pm$ 274\po} \\
TD3 with checkpoints & 10255 \textcolor{gray}{$\pm$ 656\po}  & \po3414 \textcolor{gray}{$\pm$ 77\po\po}  & \po3266 \textcolor{gray}{$\pm$ 474\po}  & \po3843 \textcolor{gray}{$\pm$ 749\po}  & \po5349 \textcolor{gray}{$\pm$ 72\po\po} \\
\midrule
No LAP & 17347 \textcolor{gray}{$\pm$ 207\po}  & \po3697 \textcolor{gray}{$\pm$ 144\po}  & \po6382 \textcolor{gray}{$\pm$ 339\po}  & \po6571 \textcolor{gray}{$\pm$ 1504}  & \po8082 \textcolor{gray}{$\pm$ 260\po} \\
TD3 with LAP & 10324 \textcolor{gray}{$\pm$ 1159}  & \po3117 \textcolor{gray}{$\pm$ 554\po}  & \po4127 \textcolor{gray}{$\pm$ 330\po}  & \po4310 \textcolor{gray}{$\pm$ 1150}  & \po5090 \textcolor{gray}{$\pm$ 190\po} \\
\midrule
No clipping & 17378 \textcolor{gray}{$\pm$ 100\po}  & \po3762 \textcolor{gray}{$\pm$ 118\po}  & \po6198 \textcolor{gray}{$\pm$ 289\po}  & \po7695 \textcolor{gray}{$\pm$ 497\po}  & \po7251 \textcolor{gray}{$\pm$ 274\po} \\
TD3 with clipping & 10283 \textcolor{gray}{$\pm$ 422\po}  & \po2969 \textcolor{gray}{$\pm$ 682\po}  & \po3990 \textcolor{gray}{$\pm$ 258\po}  & \po3711 \textcolor{gray}{$\pm$ 799\po}  & \po5254 \textcolor{gray}{$\pm$ 203\po} \\
\midrule
No normalization & 17391 \textcolor{gray}{$\pm$ 275\po}  & \po3640 \textcolor{gray}{$\pm$ 95\po\po}  & \po6256 \textcolor{gray}{$\pm$ 317\po}  & \po7807 \textcolor{gray}{$\pm$ 266\po}  & \po4829 \textcolor{gray}{$\pm$ 1809} \\
No fixed encoder & 17145 \textcolor{gray}{$\pm$ 138\po}  & \po3710 \textcolor{gray}{$\pm$ 120\po}  & \po5869 \textcolor{gray}{$\pm$ 531\po}  & \po8287 \textcolor{gray}{$\pm$ 379\po}  & \po7412 \textcolor{gray}{$\pm$ 337\po} \\
\midrule
No implementation & 17334 \textcolor{gray}{$\pm$ 99\po\po}  & \po3750 \textcolor{gray}{$\pm$ 136\po}  & \po5819 \textcolor{gray}{$\pm$ 71\po\po}  & \po7756 \textcolor{gray}{$\pm$ 704\po}  & \po7042 \textcolor{gray}{$\pm$ 259\po} \\
Our TD3 & 11068 \textcolor{gray}{$\pm$ 1399}  & \po2791 \textcolor{gray}{$\pm$ 632\po}  & \po4179 \textcolor{gray}{$\pm$ 297\po}  & \po5489 \textcolor{gray}{$\pm$ 448\po}  & \po5186 \textcolor{gray}{$\pm$ 108\po} \\
\bottomrule
\end{tabularx}
\end{table*}

\begin{figure}[ht]
	\appendixfigure{ablation_0}
	\fcolorbox{gray}{gray!10}{
		\small
		\cblock{sb_blue}~TD7 \quad \cblock{sb_orange}~TD3 \quad \cblock{sb_green}~No SALE \quad \cblock{sb_red} No SALE, with encoder \quad \cblock{sb_purple}~TD3 with encoder
	} 
	\caption{\textbf{SALE.} Learning curves on the MuJoCo benchmark, varying the usage of SALE. Results are averaged over $10$ seeds. The shaded area captures a $95\%$ confidence interval around the average performance.}  \label{appendix:fig:ablation:SALE}
\end{figure}

\begin{figure}[ht]
\appendixfigure{ablation_1}
\fcolorbox{gray}{gray!10}{
\small
\cblock{sb_blue}~TD7 \quad \cblock{sb_orange}~TD3 \quad \cblock{sb_green}~No checkpoints \quad \cblock{sb_red} Current policy \quad \cblock{sb_purple}~TD3 with checkpoints
} 
\caption{\textbf{Checkpoints.} Learning curves on the MuJoCo benchmark, varying the usage of policy checkpoints. Results are averaged over $10$ seeds. The shaded area captures a $95\%$ confidence interval around the average performance.}  \label{appendix:fig:ablation:checkpoints}
\vspace{-8pt}
\end{figure}

\clearpage

\begin{figure}[ht]
	\appendixfigure{ablation_2}
	\fcolorbox{gray}{gray!10}{
		\small
		\cblock{sb_blue}~TD7 \quad \cblock{sb_orange}~TD3 \quad \cblock{sb_green}~No LAP \quad \cblock{sb_red}~TD3 with LAP
	} 
	\captionof{figure}{\textbf{LAP.} Learning curves on the MuJoCo benchmark, varying the usage of LAP. Results are averaged over $10$ seeds. The shaded area captures a $95\%$ confidence interval around the average performance.}  \label{appendix:fig:ablation:LAP}
	
	\appendixfigure{ablation_3}
	\fcolorbox{gray}{gray!10}{
		\small
		\cblock{sb_blue}~TD7 \quad \cblock{sb_orange}~TD3 \quad \cblock{sb_green}~No clipping \quad \cblock{sb_red}~TD3 with clipping
	} 
	\captionof{figure}{\textbf{Clipping.} Learning curves on the MuJoCo benchmark, varying the usage of our proposed value clipping for mitigating extrapolation error. Results are averaged over $10$ seeds. The shaded area captures a $95\%$ confidence interval around the average performance.} \label{appendix:fig:ablation:clipping}
	
	\appendixfigure{ablation_4}
	\fcolorbox{gray}{gray!10}{
		\small
		\cblock{sb_blue}~TD7 \quad \cblock{sb_orange}~TD3 \quad \cblock{sb_green}~No normalization \quad \cblock{sb_red}~No fixed encoder
	} 
	\captionof{figure}{\textbf{SALE components.} Learning curves on the MuJoCo benchmark, removing components of SALE. Results are averaged over $10$ seeds. The shaded area captures a $95\%$ confidence interval around the average performance.} \label{appendix:fig:ablation:SALE_improvements}
	
	\appendixfigure{ablation_5}
	\fcolorbox{gray}{gray!10}{
		\small
		\cblock{sb_blue}~TD7 \quad \cblock{sb_orange}~TD3 \quad \cblock{sb_green}~No implementation \quad \cblock{sb_red}~Our TD3
	} 
	\captionof{figure}{\textbf{Implementation differences.} Learning curves on the MuJoCo benchmark, varying the minor implementation details between TD3 and TD7. Results are averaged over $10$ seeds. The shaded area captures a $95\%$ confidence interval around the average performance.}  \label{appendix:fig:ablation:implementation}
\end{figure}

\clearpage

\section{Offline RL Learning Curves}

\begin{figure}[ht]
\centering
\hspace{48pt}
\begin{tikzpicture}[trim axis right, trim axis left]
\begin{axis}[
    width=0.175\textwidth,
    x tick label style={yshift=2pt},
    y tick label style={xshift=2pt},
    ylabel style={yshift=-3pt},
    title={\shortstack{\vphantom{p}HalfCheetah\\Medium\vphantom{p}}},
    ylabel={Normalized Score},
    xlabel={Time steps (1M)},
    xtick={0, 0.5, 1.0},
    xticklabels={0, 0.5, 1.0},
    ytick={0, 20, 40, 60, 80, 100, 120},
    yticklabels={0, 20, 40, 60, 80, 100, 120},
]
\addplot graphics [
ymin=-3.6, ymax=123.6,
xmin=-0.06, xmax=1.06,
]{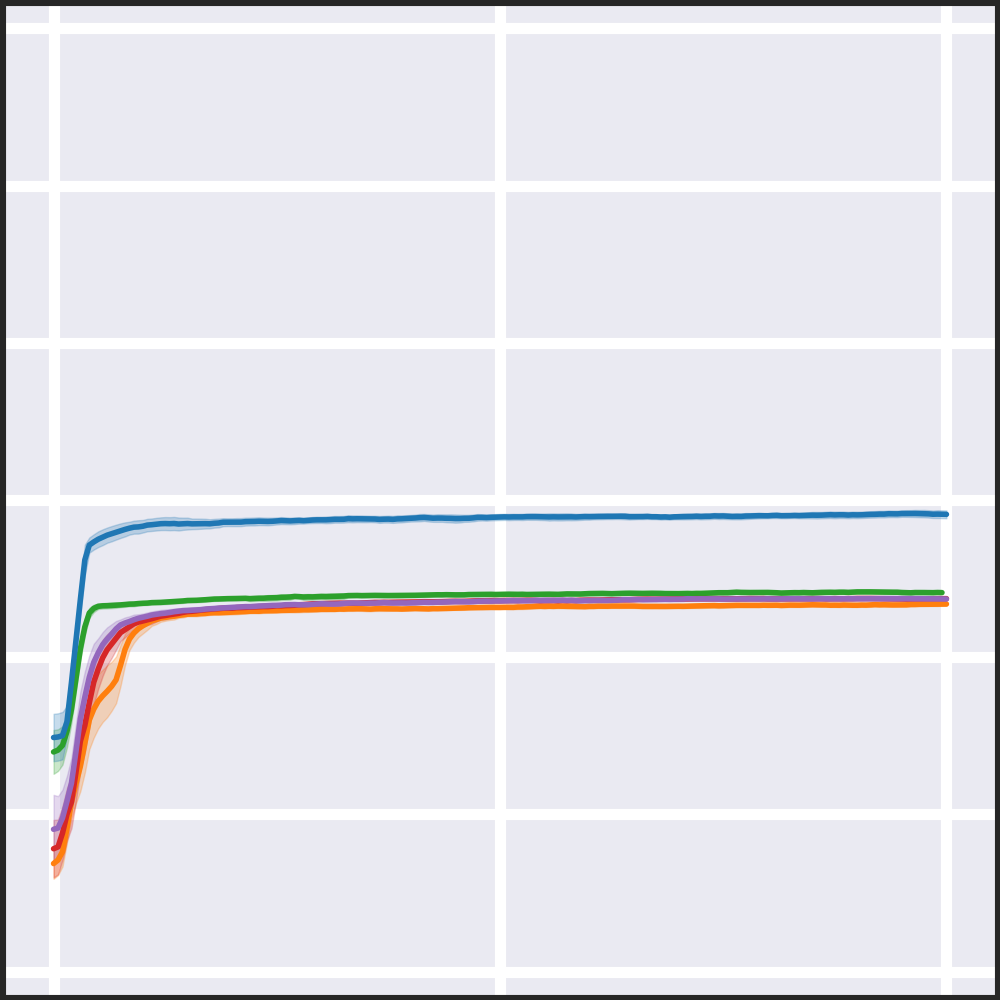};
\end{axis}
\end{tikzpicture}
\hfill
\begin{tikzpicture}[trim axis right]
\begin{axis}[
    width=0.175\textwidth,
    x tick label style={yshift=2pt},
    y tick label style={xshift=2pt},
    title={\shortstack{\vphantom{p}HalfCheetah\\Medium-Replay\vphantom{p}}},
    xlabel={Time steps (1M)},
    xtick={0, 0.5, 1.0},
    xticklabels={0, 0.5, 1.0},
    ytick={0, 20, 40, 60, 80, 100, 120},
    yticklabels={0, 20, 40, 60, 80, 100, 120},
]
\addplot graphics [
ymin=-3.6, ymax=123.6,
xmin=-0.06, xmax=1.06,
]{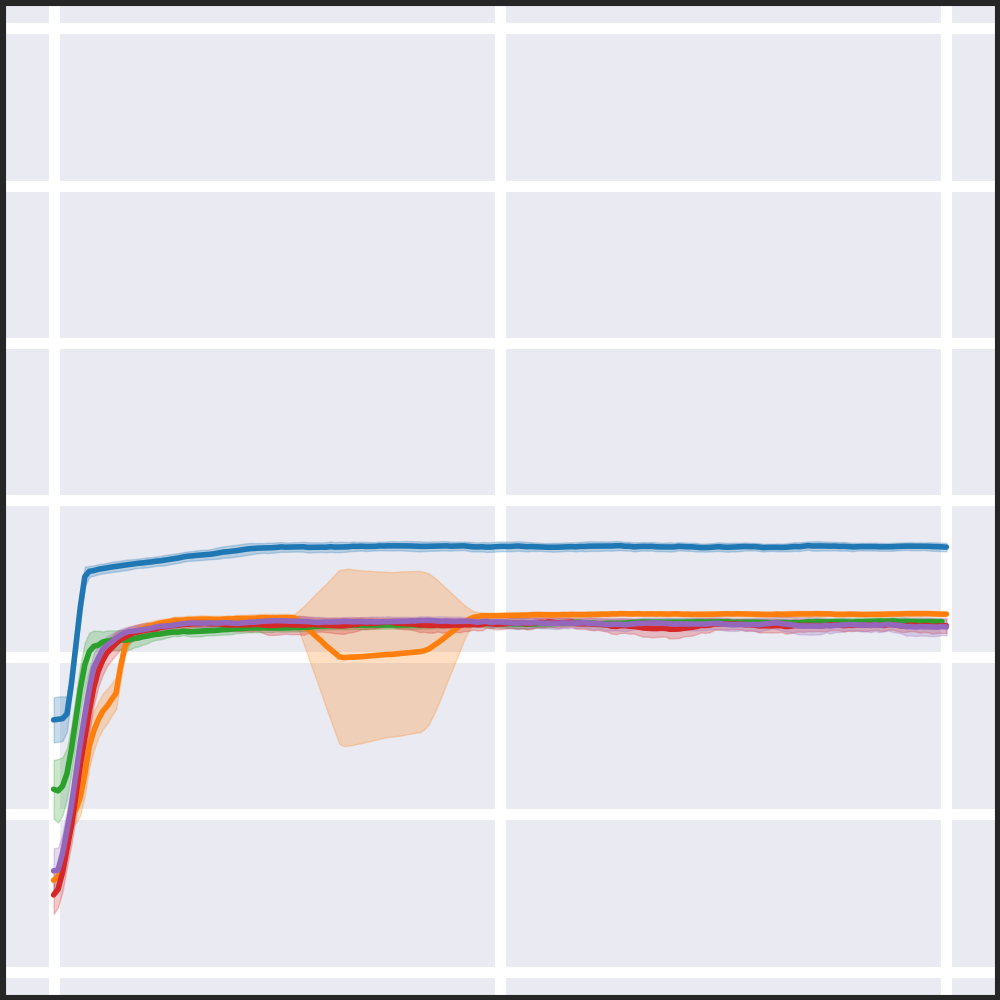};
\end{axis}
\end{tikzpicture}
\hfill
\begin{tikzpicture}[trim axis right]
\begin{axis}[
    width=0.175\textwidth,
    x tick label style={yshift=2pt},
    y tick label style={xshift=2pt},    
    title={\shortstack{\vphantom{p}HalfCheetah\\Medium-Expert\vphantom{p}}},
    xlabel={Time steps (1M)},
    xtick={0, 0.5, 1.0},
    xticklabels={0, 0.5, 1.0},
    ytick={0, 20, 40, 60, 80, 100, 120},
    yticklabels={0, 20, 40, 60, 80, 100, 120},
]
\addplot graphics [
ymin=-3.6, ymax=123.6,
xmin=-0.06, xmax=1.06,
]{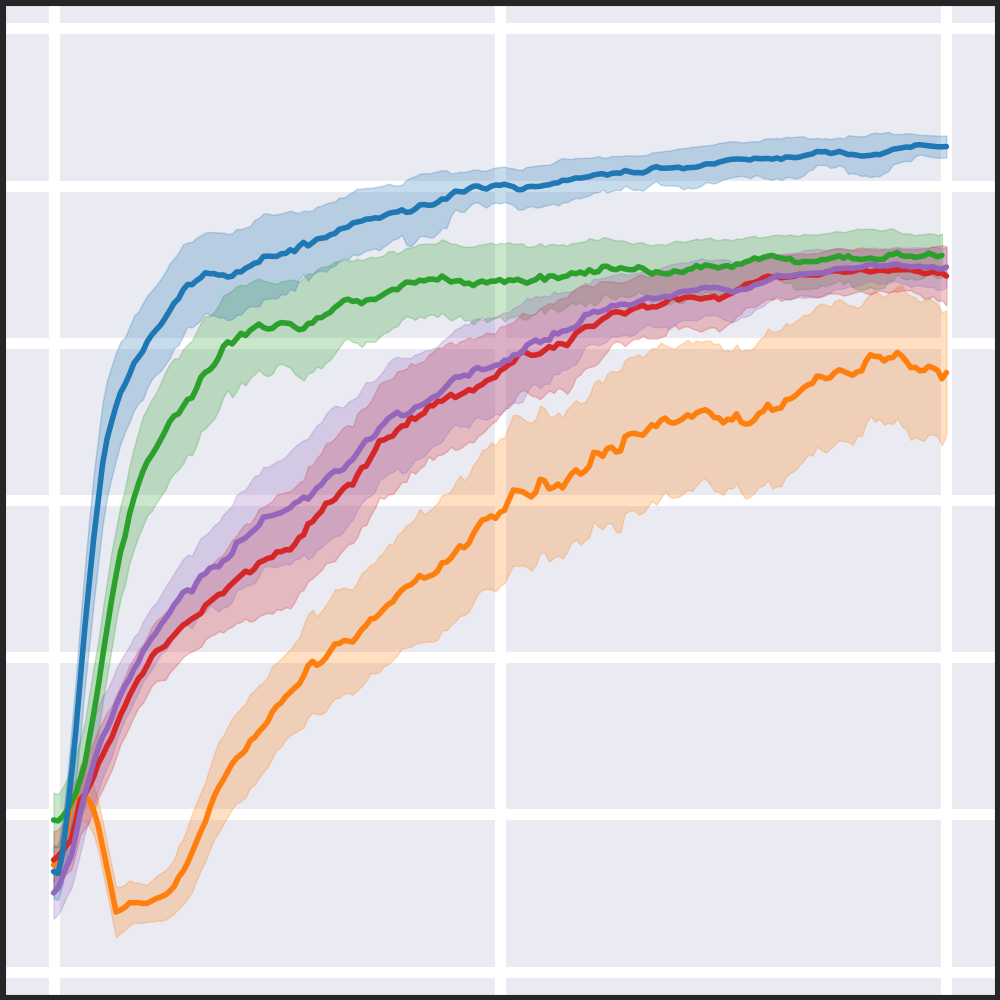};
\end{axis}
\end{tikzpicture}
\hspace{48pt}

\hspace{48pt}
\begin{tikzpicture}[trim axis right, trim axis left]
\begin{axis}[
    width=0.175\textwidth,
    x tick label style={yshift=2pt},
    y tick label style={xshift=2pt},
    ylabel style={yshift=-3pt},
    title={\shortstack{\vphantom{p}Hopper\\Medium\vphantom{p}}},
    ylabel={Normalized Score},
    xlabel={Time steps (1M)},
    xtick={0, 0.5, 1.0},
    xticklabels={0, 0.5, 1.0},
    ytick={0, 20, 40, 60, 80, 100, 120},
    yticklabels={0, 20, 40, 60, 80, 100, 120},
]
\addplot graphics [
ymin=-3.6, ymax=123.6,
xmin=-0.06, xmax=1.06,
]{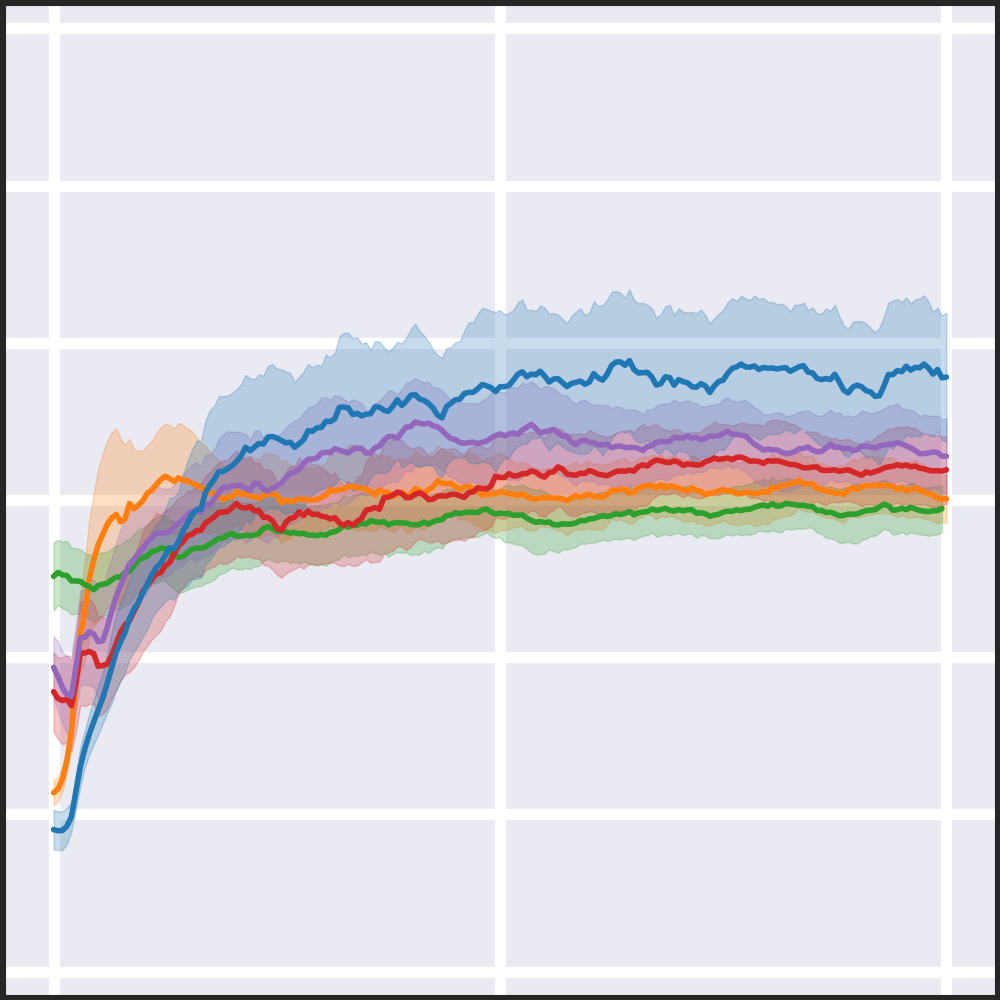};
\end{axis}
\end{tikzpicture}
\hfill
\begin{tikzpicture}[trim axis right]
\begin{axis}[
    width=0.175\textwidth,
    x tick label style={yshift=2pt},
    y tick label style={xshift=2pt},
    title={\shortstack{\vphantom{p}Hopper\\Medium-Replay\vphantom{p}}},
    xlabel={Time steps (1M)},
    xtick={0, 0.5, 1.0},
    xticklabels={0, 0.5, 1.0},
    ytick={0, 20, 40, 60, 80, 100, 120},
    yticklabels={0, 20, 40, 60, 80, 100, 120},
]
\addplot graphics [
ymin=-3.6, ymax=123.6,
xmin=-0.06, xmax=1.06,
]{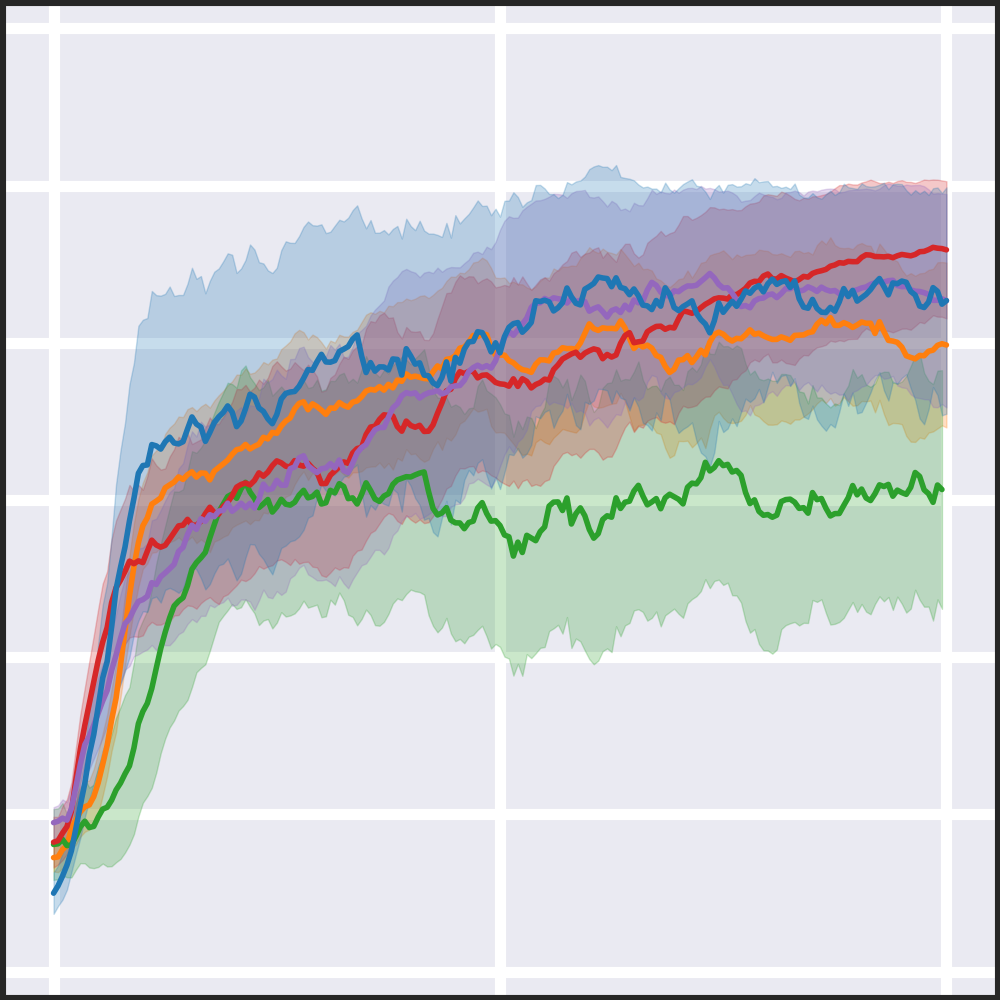};
\end{axis}
\end{tikzpicture}
\hfill
\begin{tikzpicture}[trim axis right]
\begin{axis}[
    width=0.175\textwidth,
    x tick label style={yshift=2pt},
    y tick label style={xshift=2pt},
    title={\shortstack{\vphantom{p}Hopper\\Medium-Expert\vphantom{p}}},
    xlabel={Time steps (1M)},
    xtick={0, 0.5, 1.0},
    xticklabels={0, 0.5, 1.0},
    ytick={0, 20, 40, 60, 80, 100, 120},
    yticklabels={0, 20, 40, 60, 80, 100, 120},
]
\addplot graphics [
ymin=-3.6, ymax=123.6,
xmin=-0.06, xmax=1.06,
]{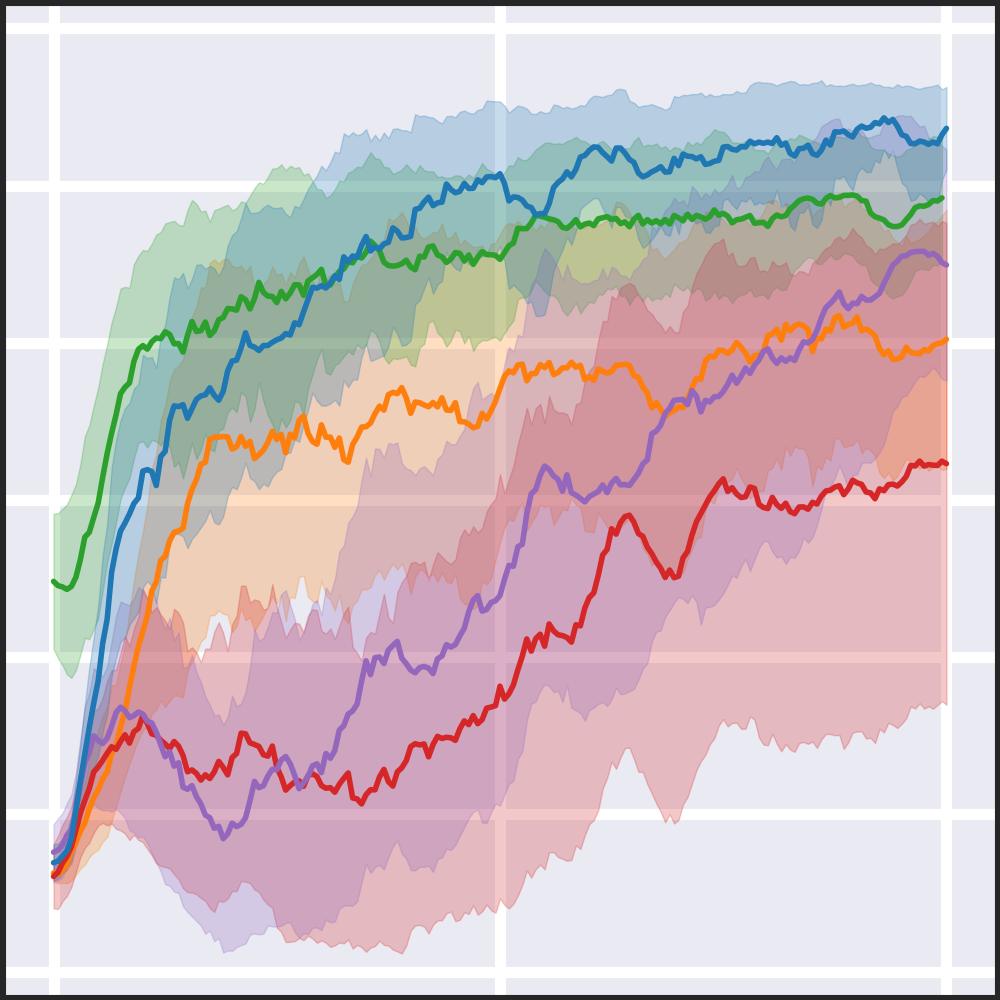};
\end{axis}
\end{tikzpicture}
\hspace{48pt}

\hspace{48pt}
\begin{tikzpicture}[trim axis right, trim axis left]
\begin{axis}[
    width=0.175\textwidth,
    x tick label style={yshift=2pt},
    y tick label style={xshift=2pt},
    ylabel style={yshift=-3pt},
    title={\shortstack{\vphantom{p}Walker2d\\Medium\vphantom{p}}},
    ylabel={Normalized Score},
    xlabel={Time steps (1M)},
    xtick={0, 0.5, 1.0},
    xticklabels={0, 0.5, 1.0},
    ytick={0, 20, 40, 60, 80, 100, 120},
    yticklabels={0, 20, 40, 60, 80, 100, 120},
]
\addplot graphics [
ymin=-3.6, ymax=123.6,
xmin=-0.06, xmax=1.06,
]{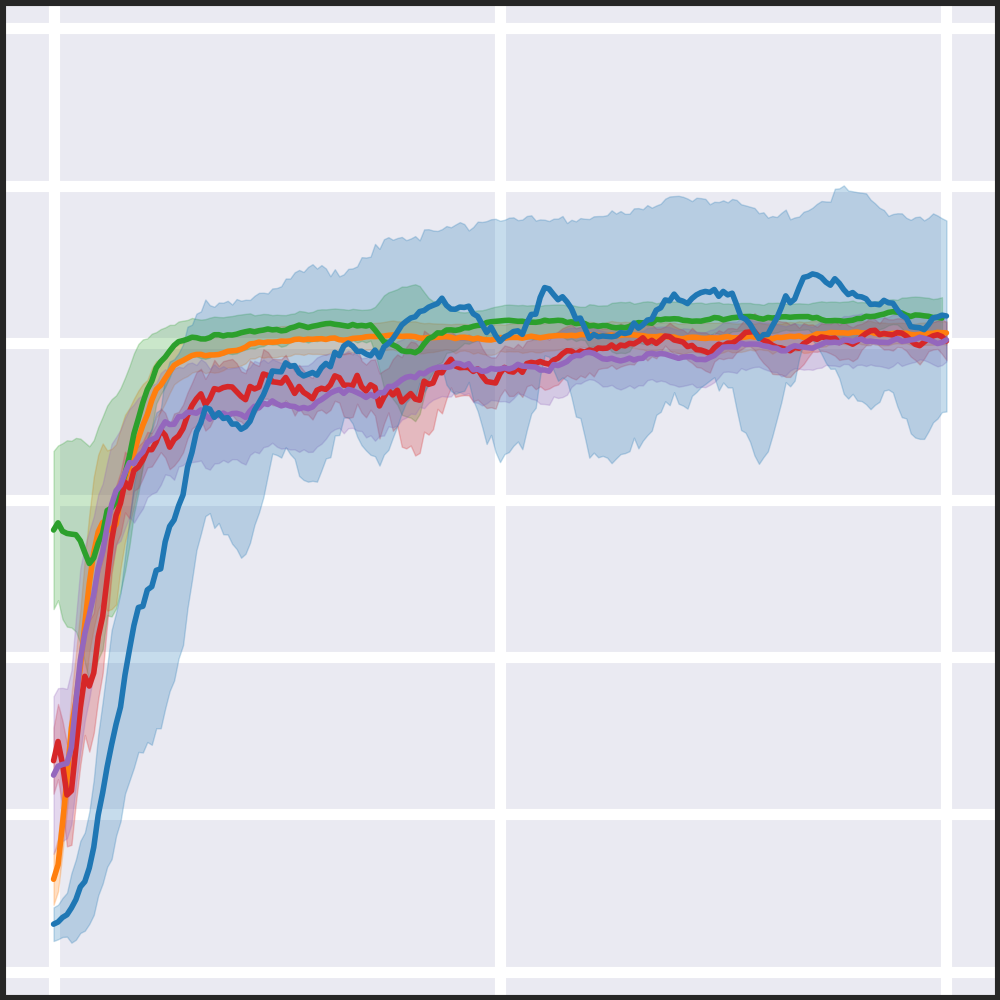};
\end{axis}
\end{tikzpicture}
\hfill
\begin{tikzpicture}[trim axis right]
\begin{axis}[
    width=0.175\textwidth,
    x tick label style={yshift=2pt},
    y tick label style={xshift=2pt},
    title={\shortstack{\vphantom{p}Walker2d\\Medium-Replay\vphantom{p}}},
    xlabel={Time steps (1M)},
    xtick={0, 0.5, 1.0},
    xticklabels={0, 0.5, 1.0},
    ytick={0, 20, 40, 60, 80, 100, 120},
    yticklabels={0, 20, 40, 60, 80, 100, 120},
]
\addplot graphics [
ymin=-3.6, ymax=123.6,
xmin=-0.06, xmax=1.06,
]{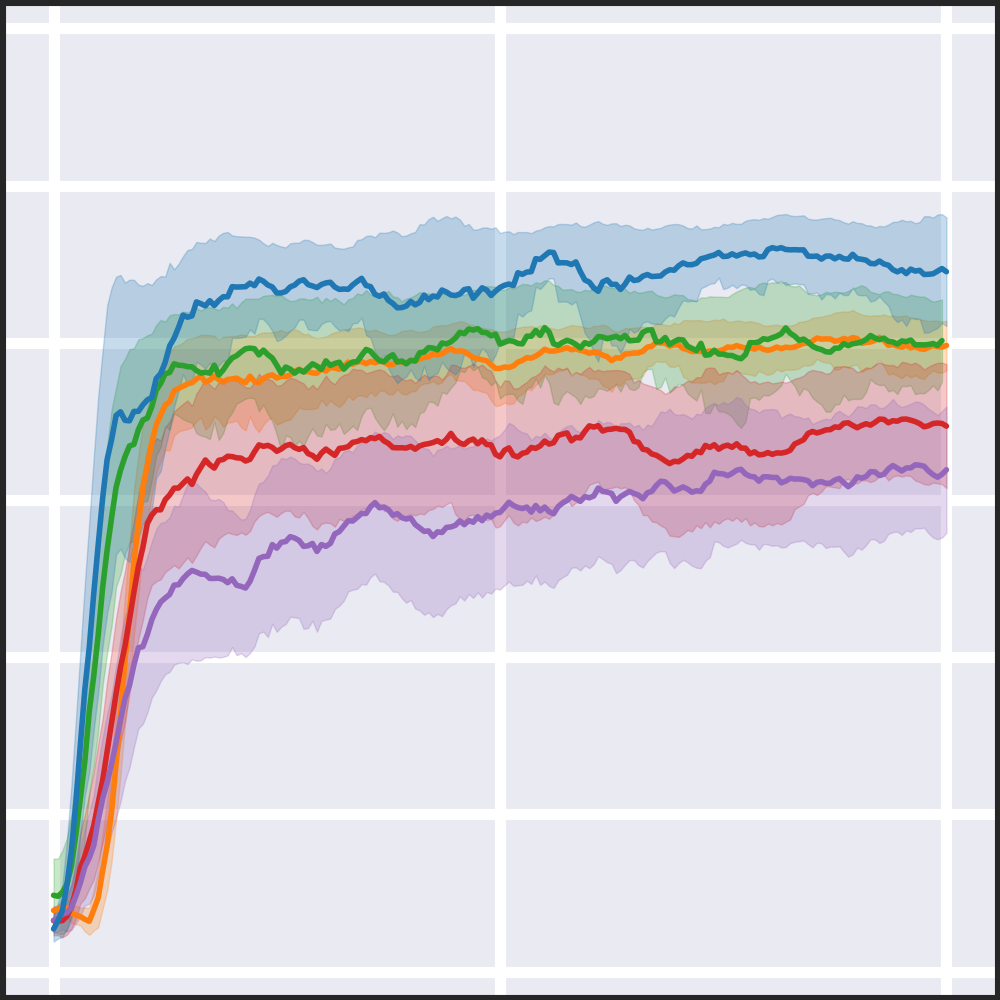};
\end{axis}
\end{tikzpicture}
\hfill
\begin{tikzpicture}[trim axis right]
\begin{axis}[
    width=0.175\textwidth,
    x tick label style={yshift=2pt},
    y tick label style={xshift=2pt},
    title={\shortstack{\vphantom{p}Walker2d\\Medium-Expert\vphantom{p}}},
    xlabel={Time steps (1M)},
    xtick={0, 0.5, 1.0},
    xticklabels={0, 0.5, 1.0},
    ytick={0, 20, 40, 60, 80, 100, 120},
    yticklabels={0, 20, 40, 60, 80, 100, 120},
]
\addplot graphics [
ymin=-3.6, ymax=123.6,
xmin=-0.06, xmax=1.06,
]{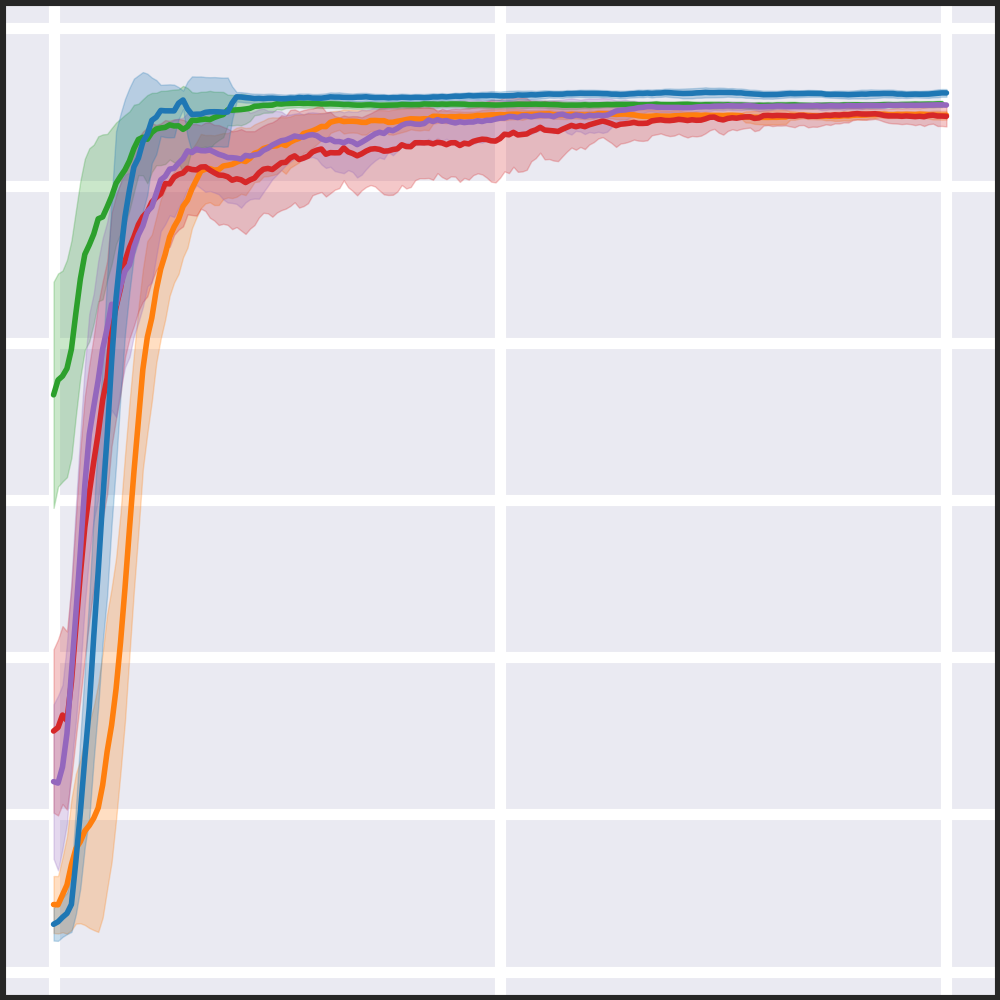};
\end{axis}
\end{tikzpicture}
\hspace{48pt}

\fcolorbox{gray}{gray!10}{
\small
\cblock{sb_blue}~TD7 \quad \cblock{sb_orange}~CQL \quad \cblock{sb_green}~TD3+BC \quad \cblock{sb_red}~IQL \quad \cblock{sb_purple} $\mathcal{X}$-QL
} 

\caption{Learning curves on the offline D4RL benchmark. Results are averaged over 10 seeds. The shaded area captures a 95\% confidence interval around the average performance.} 
\end{figure}

\clearpage

\section{Run time}

We benchmark the run time of TD7 and the baseline algorithms. Evaluation episodes were not included in this analysis. Each algorithm is trained using the same deep learning framework, PyTorch~\citep{paszke2019pytorch}. All experiments are run on a single Nvidia Titan X GPU and Intel Core i7-7700k CPU. Software is detailed in \autoref{appendix:sec:experimental}. %
We display run time adjusted learning curves in \autoref{appendix:fig:runtime_curves}. %

\begin{figure}[ht]
	\centering
	\hspace{14pt}
	\begin{tikzpicture}[trim axis right, trim axis left]
		\begin{axis}[
			width=0.175\textwidth,
			x tick label style={yshift=2pt},
			y tick label style={xshift=2pt},
			ylabel style={yshift=-3pt},
			title={\phantom{p}HalfCheetah\phantom{p}},
			ylabel={Total Reward (1k)},
			xlabel={Hours},
			xtick={0, 4, 8, 12, 16, 20},
			xticklabels={0, 4, 8, 12, 16, 20},
			ytick={0, 4, 8, 12, 16, 20},
			yticklabels={0, 4, 8, 12, 16, 20},
			]
			\addplot graphics [
			ymin=-0.6, ymax=20.6,
			xmin=-0.57, xmax=19.72,
			]{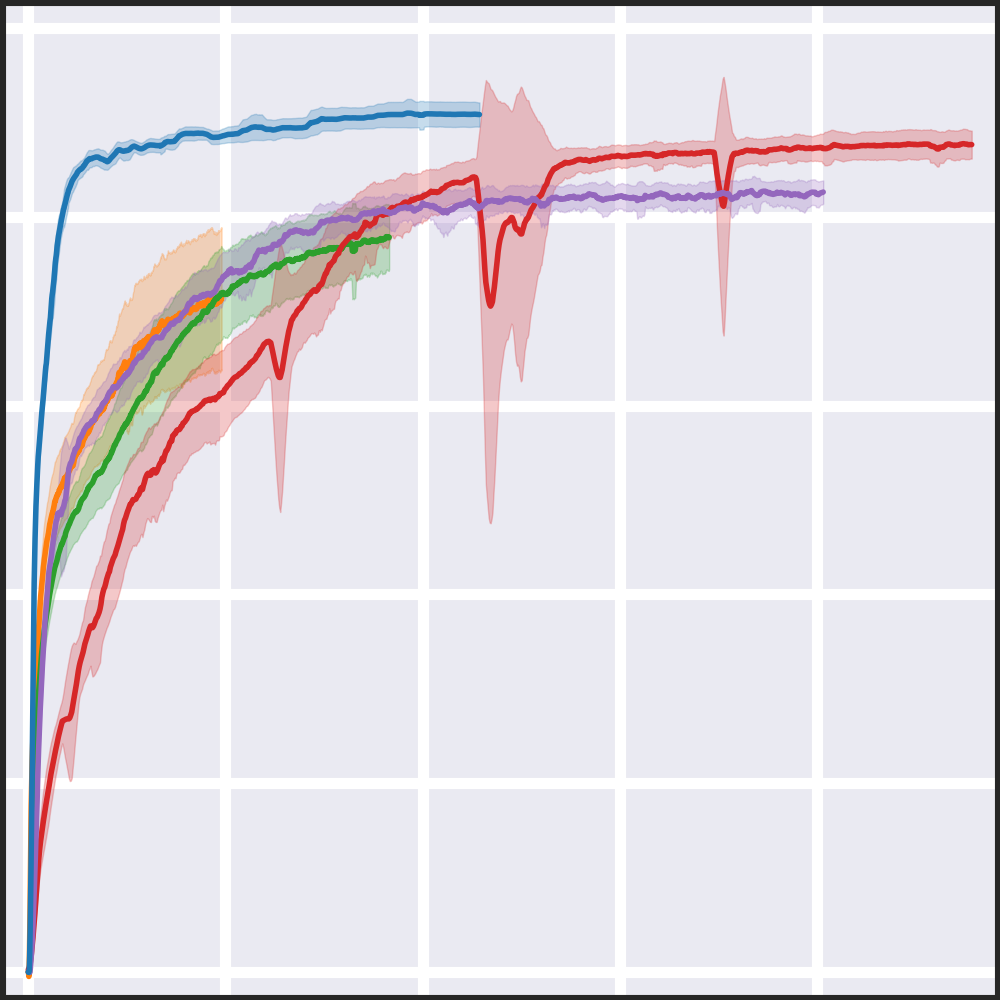};
		\end{axis}
	\end{tikzpicture}
	\hspace{-6pt}
	\begin{tikzpicture}[trim axis right]
		\begin{axis}[
			width=0.175\textwidth,
			x tick label style={yshift=2pt},
			y tick label style={xshift=2pt},
			title={Hopper},
			xlabel={Hours},
			xtick={0, 4, 8, 12, 16, 20},
			xticklabels={0, 4, 8, 12, 16, 20},
			ytick={0, 1, 2, 3, 4},
			yticklabels={0, 1, 2, 3, 4},
			]
			\addplot graphics [
			ymin=-0.135, ymax=4.635,
			xmin=-0.53, xmax=18.37,
			]{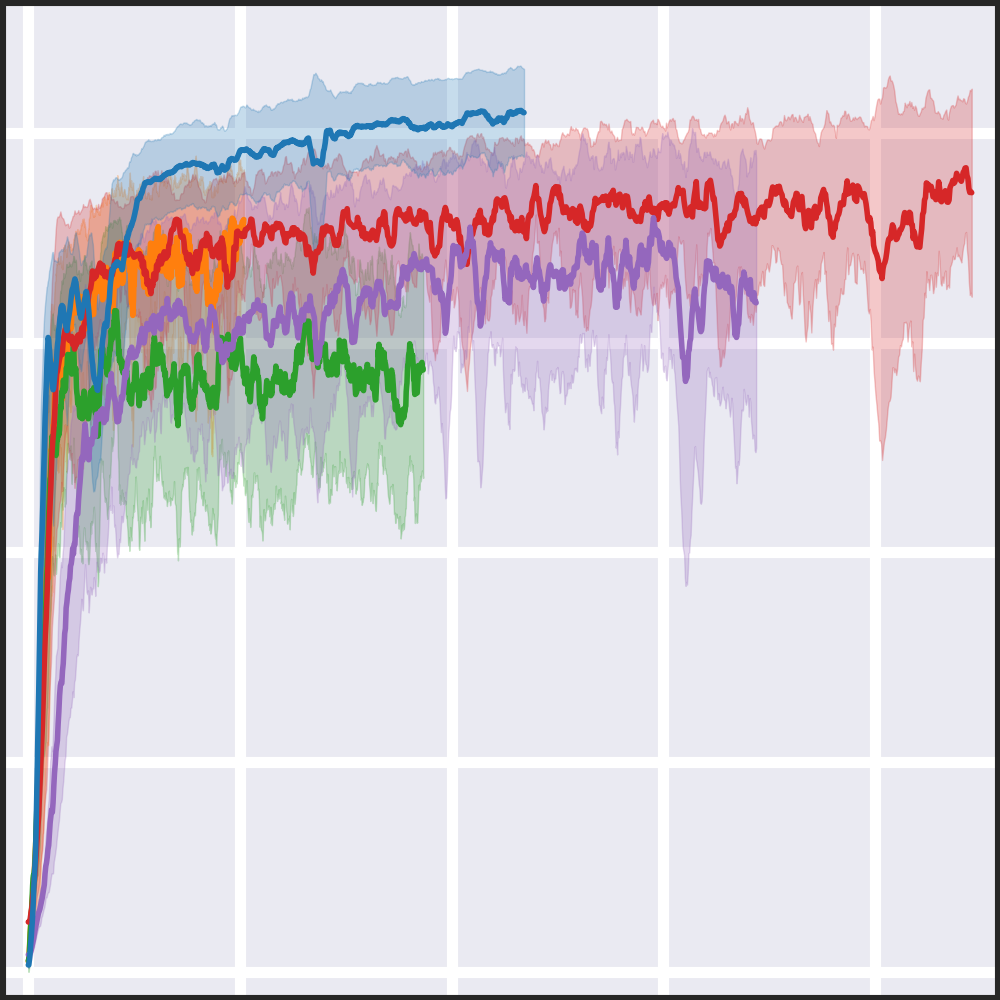};
		\end{axis}
	\end{tikzpicture}
	\hspace{-6pt}
	\begin{tikzpicture}[trim axis right]
		\begin{axis}[
			width=0.175\textwidth,
			x tick label style={yshift=2pt},
			y tick label style={xshift=2pt},    
			title={\phantom{p}Walker2d\phantom{p}},
			xlabel={Hours},
			xtick={0, 4, 8, 12, 16, 20},
			xticklabels={0, 4, 8, 12, 16, 20},
			ytick={0, 2, 4, 6, 8},
			yticklabels={0, 2, 4, 6, 8},
			]
			\addplot graphics [
			ymin=-0.24, ymax=8.24,
			xmin=-0.54, xmax=18.4,
			]{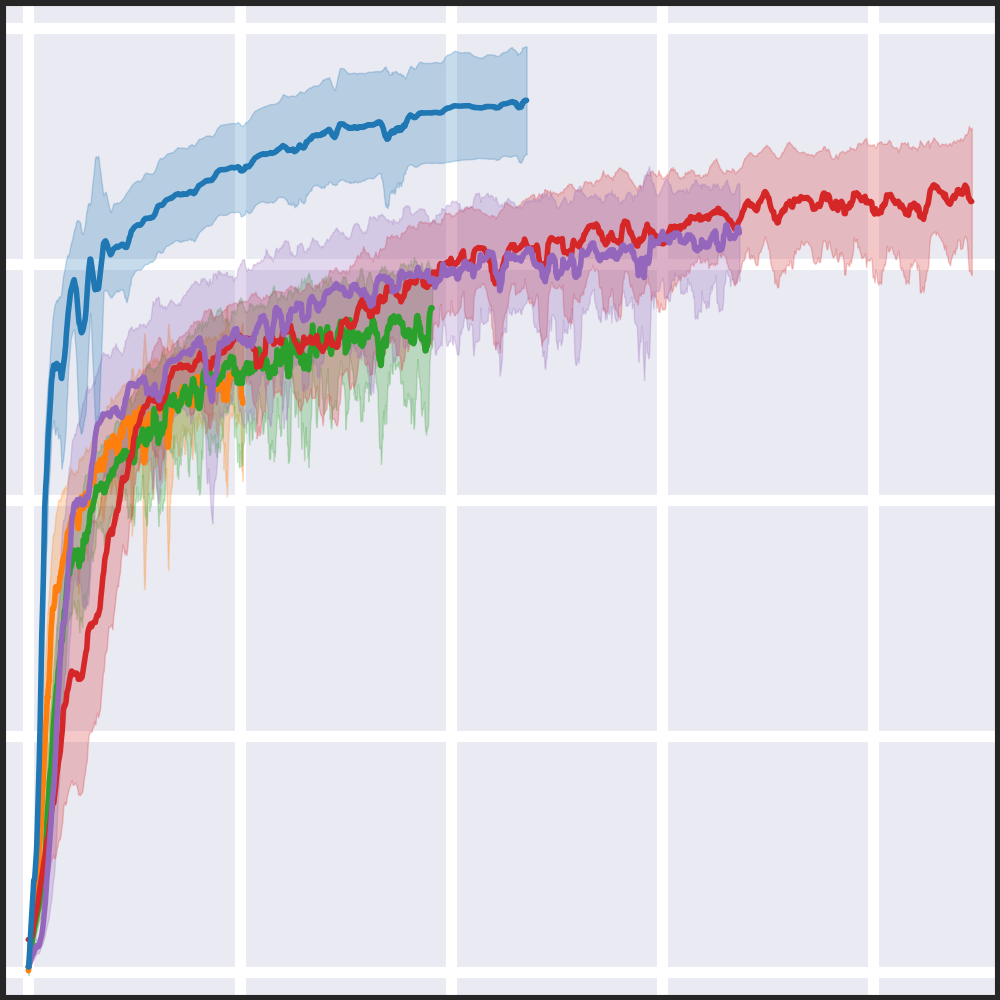};
		\end{axis}
	\end{tikzpicture}
	\hspace{-8pt}
	\begin{tikzpicture}[trim axis right]
		\begin{axis}[
			width=0.175\textwidth,
			x tick label style={yshift=2pt},
			y tick label style={xshift=2pt},
			title={\phantom{p}Ant\phantom{p}},
			xlabel={Hours},
			xtick={0, 4, 8, 12, 16, 20},
			xticklabels={0, 4, 8, 12, 16, 20},
			ytick={0, 3, 6, 9, 12},
			yticklabels={0, 3, 6, 9, 12},
			]
			\addplot graphics [
			ymin=-0.36, ymax=12.36,
			xmin=-0.56, xmax=19.19,
			]{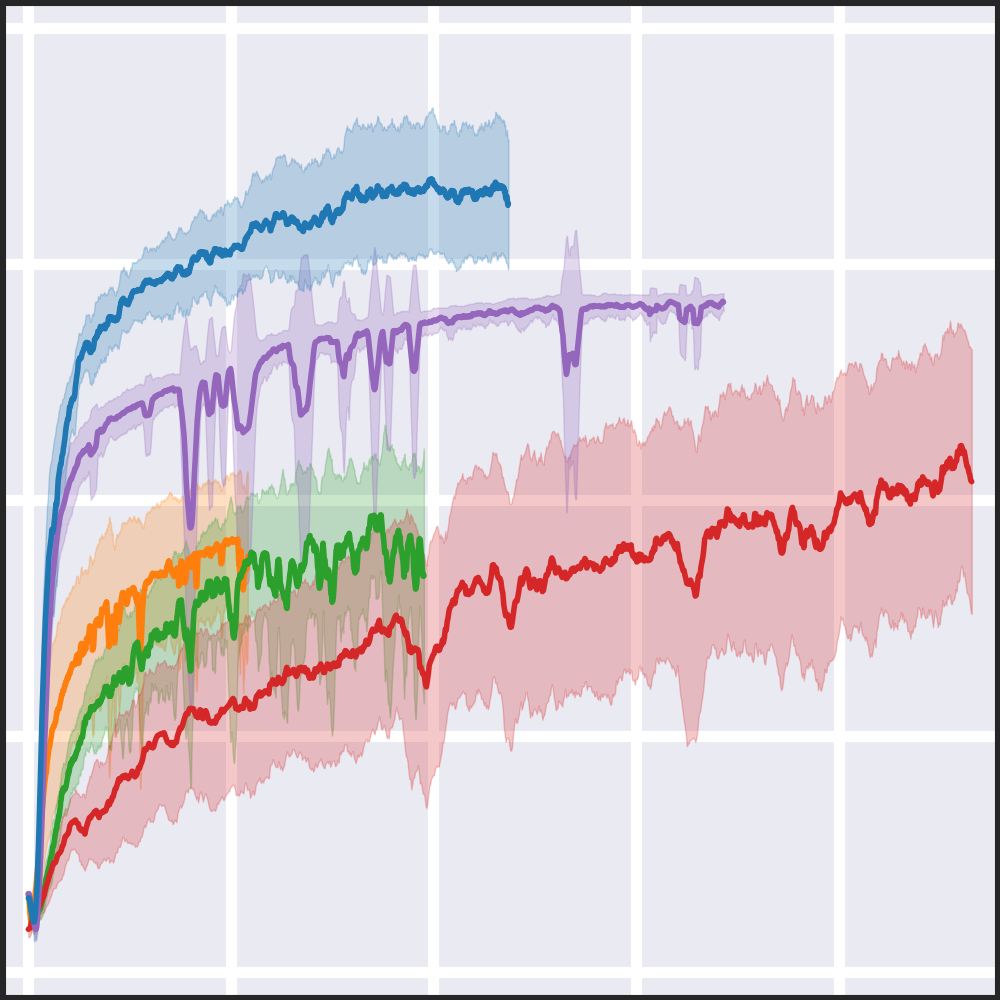};
		\end{axis}
	\end{tikzpicture}
	\hspace{-8pt}
	\begin{tikzpicture}[trim axis right]
		\begin{axis}[
			width=0.175\textwidth,
			x tick label style={yshift=2pt},
			y tick label style={xshift=2pt},
			title={\phantom{p}Humanoid\phantom{p}},
			xlabel={Hours},
			xtick={0, 4, 8, 12, 16, 20},
			xticklabels={0, 4, 8, 12, 16, 20},
			ytick={0, 2, 4, 6, 8, 10},
			yticklabels={0, 2, 4, 6, 8, 10},
			]
			\addplot graphics [
			ymin=-0.33, ymax=11.33,
			xmin=-0.59, xmax=20.21,
			]{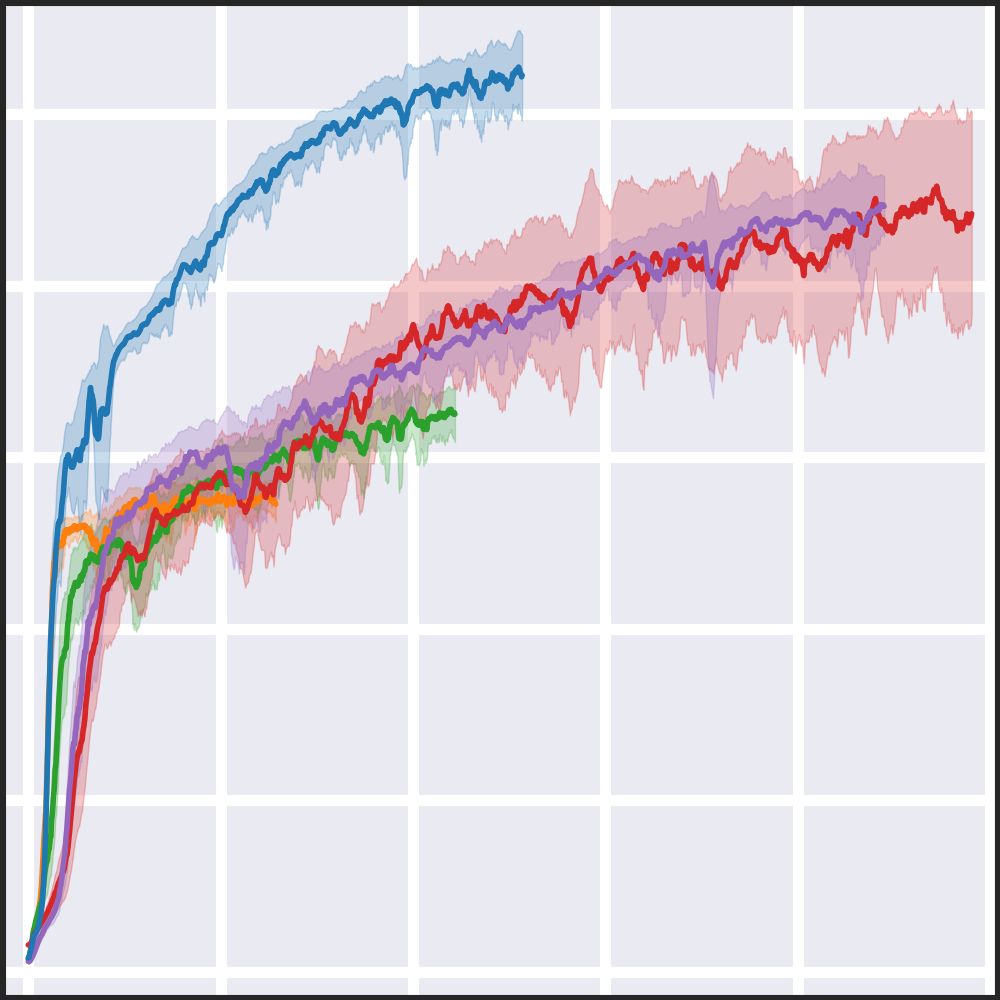};
		\end{axis}
	\end{tikzpicture}
	
	\fcolorbox{gray}{gray!10}{
		\small
		\cblock{sb_blue}~TD7 
		\quad \cblock{sb_orange}~TD3 \quad \cblock{sb_green}~SAC \quad \cblock{sb_red}~TQC \quad \cblock{sb_purple}~TD3+OFE
	} 
	
	\caption{\textbf{Run time adjusted curves.} Learning curves on the MuJoCo benchmark over 5M time steps, where the x-axis is run time. Results are averaged over 10 seeds. The shaded area captures a 95\% confidence interval around the average performance.} \label{appendix:fig:runtime_curves}
\end{figure}

\clearpage

\section{Limitations}

\textbf{Depth vs.\ Breadth.} Although our empirical study is extensive, the scope of our experimentation emphasizes depth over breadth. There remains many other interesting benchmarks for continuous control, such as examining a setting with image-based observations. 

\textbf{Baselines.} We benchmark TD7 against the highest performing agents in both the online and offline setting. However, due to the fact that most representation learning methods do not cover low level states, we only benchmark against a single other method focused on representation learning. While it would be surprising if a method, originally designed for a different setting, outperformed the strongest existing baselines, exploring additional methods could provide new insight into representation learning for low level states.

\textbf{Computational cost.} While TD7 has a lower computational cost than other competing methods, the run time over TD3 (the base algorithm) is more than double. 

\textbf{Theoretical results.} We perform extensive empirical analysis to uncover which factors of dynamics-based representation learning are important for performance. However, we do not address the theoretical side of representation learning. There are many important questions regarding why dynamics-based representation learning is effective, or what makes a good representation.

\fi

\end{document}